\RequirePackage{fix-cm}
\documentclass[twocolumn]{svjour3}
\smartqed 

\usepackage{mathptmx}
\usepackage{amsmath}
\usepackage{amssymb}
\usepackage{color}
\usepackage{cancel}
\usepackage{soul}
\usepackage{gensymb}
\usepackage{verbatim}
\usepackage{textcomp}

\usepackage{ulem}
\usepackage{graphicx}
\usepackage{caption}
\usepackage{subcaption}
\usepackage{float}

\usepackage[breaklinks=true,bookmarks=false]{hyperref}

\newcommand{\ie}{i.e.,}

\begin{document}
\title{Root Identification in Minirhizotron Imagery with
       Multiple Instance Learning}
       
\author{Guohao~Yu  \and
        Alina~Zare \and
        Hudanyun~Sheng \and
        Roser~Matamala \and
        Joel~Reyes-Cabrera \and
        Felix~B.~Fritschi \and
        Thomas~E.~Juenger}

\institute{ G. Yu \and A. Zare \and H. Sheng \at 
            Department of Electrical and Computer Engineering, University of Florida, Gainesville, FL USA 32611 \\
            \email{guohaoyu@ufl.edu} (corresponding author G. Yu) \\
            \email{azare@ece.ufl.edu} (corresponding author A. Zare) \\
           \and
             R. Matamala \at 
             Argonne National Laboratory, Lemont, IL USA 60439 \\
           \and
            J. Reyes-Cabrera \and F. B. Fritschi \at
            Division of Plant Sciences, University of Missouri, Columbia, MO USA 65211  \\
            \and
            T. E. Juenger \at
            Department of Integrative Biology, University of Texas at Austin, Austin, TX USA 78712 
           }

\date{Received:17 March 2019 }

\maketitle

\begin{abstract}
 In this paper, multiple instance learning (MIL) algorithms to automatically perform root detection and segmentation in minirhizotron imagery using only image-level labels are proposed. Root and soil characteristics vary from location to location, thus, supervised machine learning approaches that are trained with local data provide the best ability to identify and segment roots in minirhizotron imagery.  However, labeling roots for training data (or otherwise) is an extremely tedious and time-consuming task. This paper aims to address this problem by labeling data at the image level (rather than the individual root or root pixel level) and train algorithms to perform individual root pixel level segmentation using MIL strategies. Three MIL methods (multiple instance adaptive cosine coherence estimator, multiple instance support vector machine, multiple instance learning with randomized trees) were applied to root detection and compared to non-MIL approches. The results show that MIL methods improve root segmentation in challenging minirhizotron imagery and reduce the labeling burden. In our results, multiple instance support vector machine outperformed other methods. The multiple instance adaptive cosine coherence estimator algorithm was a close second with an added advantage that it learned an interpretable root signature which identified the traits used to distinguish roots from soil and did not require parameter selection. 

\end{abstract}

\begin{acknowledgements}
This work was supported by the U.S. Department of Energy, Office of Science, Office of Biological and Environmental Research award number DE-SC0014156 and by the Advanced Research Projects Agency - Energy award number DE-AR0000820.
\end{acknowledgements}

\section{Introduction}
\label{intro}
The lack of good sensors, instruments and techniques for field soil measurements is limiting decisive and rapid advances in soil, ecosystem and agronomic science. Currently, only a few techniques are available for visualization of roots in field conditions, and those that exist are labor intensive and/or limited in what they can see and measure. Because the root system of plants is involved in the absorption of nutrients and water, understanding their traits, dynamics, and behavior is important to a broad number of disciplines, including soil and plant science, agronomists, hydrologist, earth system modelers and others. For example, the study of root systems has become vital as we look for factors that could contribute to the improvement of crop yields to guarantee the food supply. Furthermore, it has been shown that the distribution of roots at depth and their response to fertilizer application substantially affected yield of rice plants \cite{liu2018root}, and changes in root system architecture, that resulted in improvements in water capture in soil, were directly associated to biomass accumulation and historical yield trends in maize \cite{hammer2009can}. Similarly, as we look for a sustainable way to manage ecosystem services, there is a critical need to study the root system, as roots are major contributors to the buildup of organic matter in soil, and the role they can play in greenhouse gas mitigation through enhancing the sequestration of carbon in soil.  

\begin{figure}[H] 
\begin{center}
\begin{subfigure}[t]{0.23\textwidth}
   \includegraphics[width=1\linewidth]{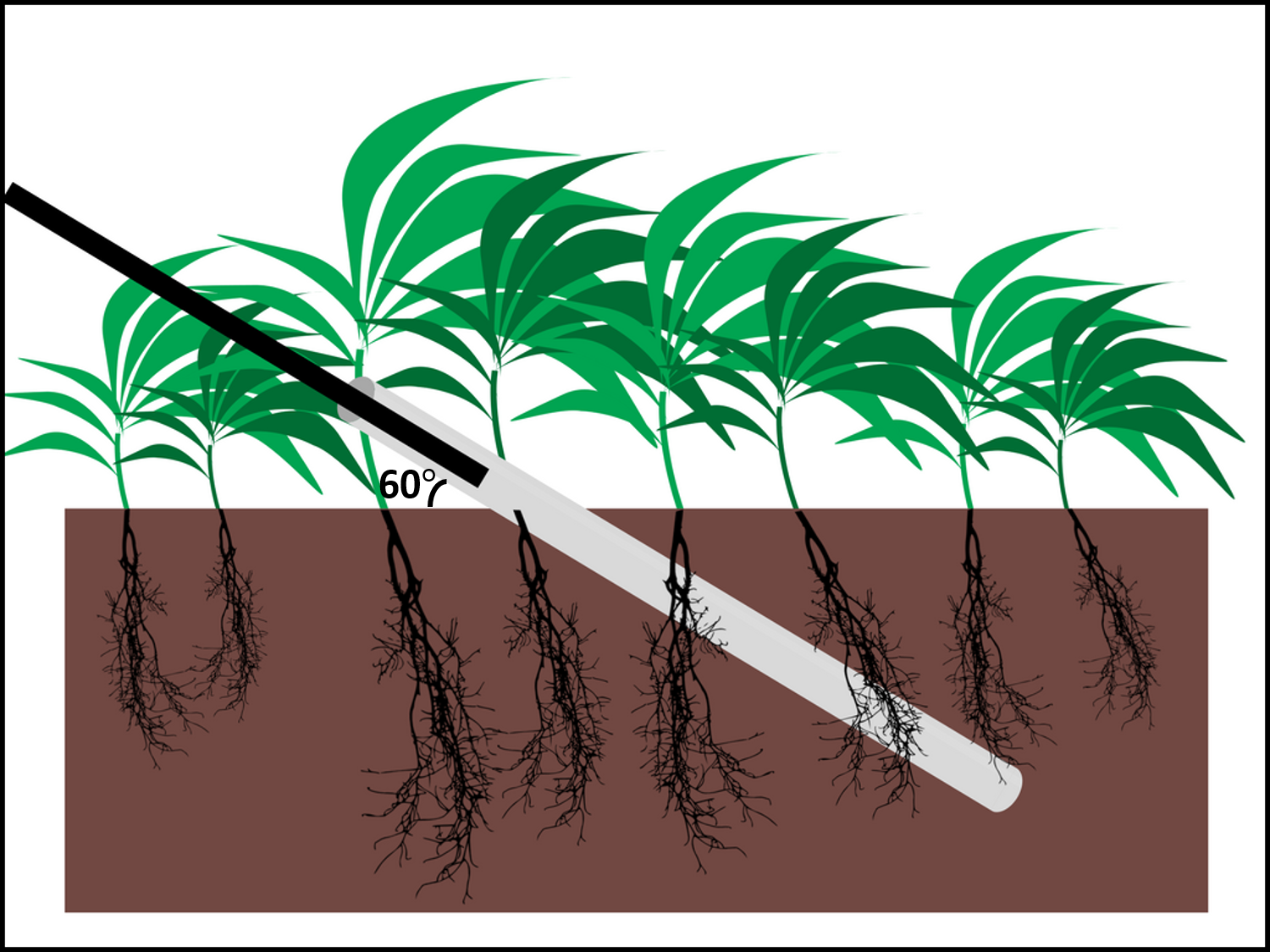}
   \caption{Minirhizotron imaging system illustration}
   \label{subfig:1_A}
\end{subfigure}
\begin{subfigure}[t]{0.23\textwidth}
   \includegraphics[width=1\linewidth]{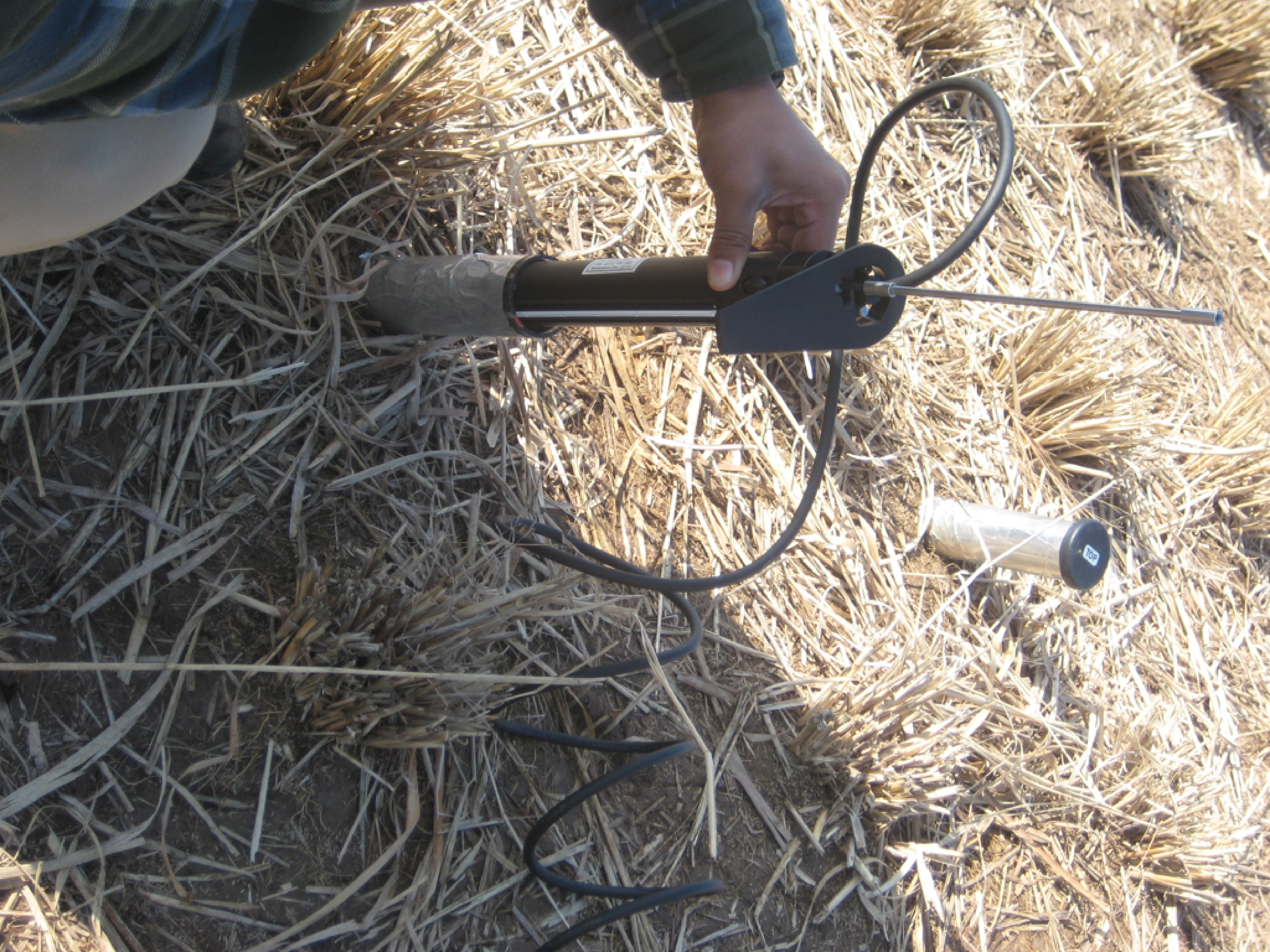}
   \caption{Minirhizotron imaging system in field}
   \label{subfig:1_B}
\end{subfigure}

   \caption{The illustration of a minirhizotron imaging system}
\label{fig:1}
\end{center}
\end{figure}

Common sampling methods used to examine root system dynamics are destructive and provide limited ability to draw inferences on the plant response to stresses experienced during the growing season. For example, soil coring, excavation, trenches, or ingrowth cores extract a unique dataset in time that destroys the integrity of the root system and thus limits repeat measurements from the same plant without the introduction of confounding effects \cite{taylor1991some}. In contrast to these destructive methods, once inserted into the soil, transparent minirhizotron access tubes allow non-destructive assessment of roots over extended periods of time without repeatedly altering critical soil conditions or root processes \cite{bates1937device,rewald2013minirhizotron,waddington1971observation,johnson2001advancing}. Roots that grow adjacent to the tube are imaged by inserting a camera (with a light source) into the tube and acquiring images along the length of the tube, providing information about roots present at different soil depths as illustrated in Fig. \ref{subfig:1_A} and \ref{subfig:1_B}. Since minirhizotron tubes allow for the collection of root images over time, this enables monitoring of complex root growth and turnover dynamics \cite{johnson2001advancing}. After collection, minirhizotron images are used to measure and characterize root traits, a process that currently is time consuming and tedious since standard analysis approaches involve manually outlining and labeling roots using commercially available software such as WinRhizo (Regent Instruments, Canada) or RootSnap (CID BioScience, Camas, WA, USA).

Machine learning methods have been studied to label roots automatically in mini-rhizotron imagery. One unsupervised learning approach proposed by Shojaedini and Heidari  \cite{shojaedini2013new} first initializes a segmentation by partitioning an image using the threshold that maximizes the second-order cross entropy of the gray-level transition probabilities between the points with intensities above the threshold and those below the threshold \cite{zeng2006detecting}. After this initial segmentation, a level set method is used to iteratively improve the identified root boundaries.

In terms of supervised machine learning, Zeng, et. al. \cite{zeng2006detecting} proposed an approach that first extracted a collection of features using linearly-shaped spatial filters. Then the feature map associated with each filter was partitioned into foreground and background classes using local entropy thresholding (LET). Features from each foreground object were then extracted and used within an Adaptive Boosting (AdaBoost) classifier to classify foreground objects as either root or non-root. Zeng, et. al. \cite{zeng2010rapid} also proposed a rapid root segmentation method that relies on seed point selection. Candidate seed points (identified using a local maxima operation) are classified as root or non-root using a linear classifier. These seed points are then ``grown'' into roots by identifying a root centerline. More recently, several deep learning methods achieved state-of-the-art results in root segmentation in mini-rhizotron imagery \cite{xu2019overcoming,wang2019segroot,yasrab2019rootnav,smith2020segmentation}. A number of different neural network architectures have been studied. The U-Net was used as the backbone architecture by Xu, et. al. and Smith, et. al.\cite{xu2019overcoming,smith2020segmentation}, Wang, et. al. \cite{wang2019segroot} relied on the SegNet architecture and hourglass networks \cite{newell2016stacked} were used by \cite{yasrab2019rootnav}. Both Xu, et. al. and Yasrab, et. al. investigated transfer learning approaches to overcome small labeled root image dataset \cite{xu2019overcoming} and \cite{yasrab2019rootnav}. Yet, these approaches tend to be either sensitive to variation in soil conditions, lighting and root color or require a very large training set. However, labeling individual root pixels to generate a large data set for a supervised learning algorithm is extremely time consuming, tedious, and prone to error. 

Although minirhizotrons significantly advance our ability to study plant root systems, data extraction from the acquired images commonly limits the extent of their use as the majority of software available to process images collected in the minirhizotrons requires human differentiation of roots and soil \cite{johnson2001advancing}. As described above, efforts to automate the labeling and segmentation of roots in minirhizotrons imagery have been undertaken \cite{zeng2006detecting,shojaedini2013new,zeng2010rapid,rahmanzadeh2016novel,xu2019overcoming,wang2019segroot,yasrab2019rootnav,smith2020segmentation}.  Yet, the best performing of these approaches require large amounts of training data and it is incredibly labor-intensive, time consuming, error-prone and tedious to label mini-rhizotron imagery at the pixel level. In this paper, we propose an efficient approach using MIL methods to label and segment root from minirhizotron data. 

MIL algorithms \cite{zare2018discriminative,dietterich1997solving,bolton2011random,chen2006miles,maron1998framework,zhang2002dd,zare2010pattern,jiao2015functions,shrivastava2014dictionary,shrivastava2015generalized} only require data to be labeled at the \textit{bag} level. Bags are a multi-set of instances and each bag is labeled as either “positive” or “negative”. A bag is labeled as a positive bag if at least one of the instances in the bag is an instance of the positive target class.  If none of the instances in a bag is of the target class, the bag is labeled as a negative bag. The advantage of this framework is that the bags can be constructed in such a way to ease the labeling burden.   In our case, the target class corresponds to root.  Thus, a positive bag could be an image containing roots and a negative bag is one not containing roots.  It is much easier to identify imagery that either do or do not contain roots as opposed to tracing out each individual root segment.   MIL algorithms have been used for molecule classification \cite{dietterich1997solving,bolton2011random,chen2006miles,maron1998framework,zhang2002dd,leistner2010miforests,andrews2003support}, image indexing \cite{shrivastava2014dictionary,shrivastava2015generalized,leistner2010miforests}. Among all the MIL algorithms, a few can estimate a discriminative target signature \cite{zare2018discriminative,maron1998framework,zhang2002dd,zare2010pattern,jiao2015functions,shrivastava2014dictionary,shrivastava2015generalized}. The estimated discriminative target signature can be examined to obtain insight into what characterizes the target class and can be easily interpreted to understand what characteristics are being used to detect and distinguish target from the background.

Our root segmentation approach used training data with only image-level labels instead of full pixel-level annotation. Such an approach significantly reduced the effort associated to label training data for the application of supervised learning algorithms used in the interpretation (or processing) of minirhizotron root images. We propose a downsampling strategy based on histograms of feature computed on superpixels that effectively increased the ratio of target instances to background instances in positive bag. We compare the ability of three MIL methods to detect roots of switchgrass plants over a range of soil textures and colors in minirhizotron images and demonstrated that application of these methods could substantially increase the speed of minirhizotron root segmentation.

\section{Methodology}

The proposed method consists of three steps: (1) image pre-processing and feature extraction; (2) root detection using an MIL algorithm (multiple instance adaptive cosine coherence estimator, multiple instance support vector machine, and multiple instance learning with randomized trees); and (3) post-processing to help reduce false detections. 

\subsection{Pre-processing and Feature Extraction}
\label{sec:preproc}

\begin{figure}[H] 
\begin{center}
~~
\begin{subfigure}[t]{0.15\textwidth}
   \includegraphics[width=1\linewidth]{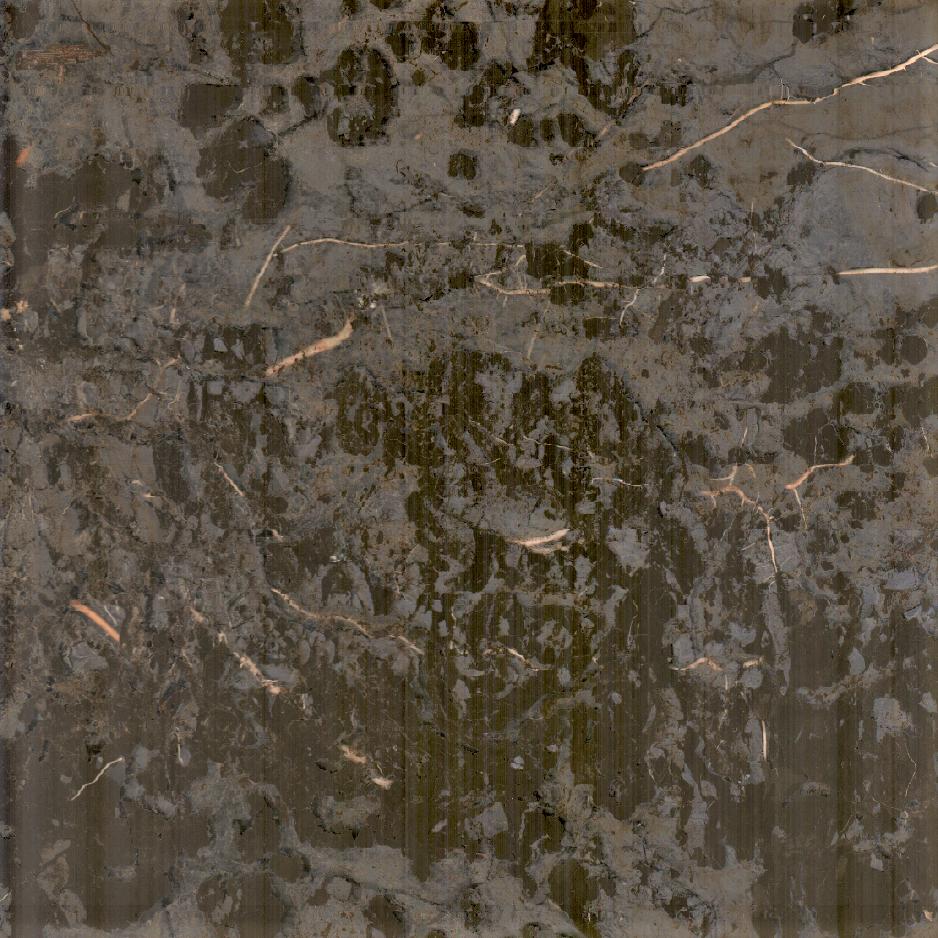}
   \caption{Original image}
   \label{subfig:2_A}
\end{subfigure}
\begin{subfigure}[t]{0.15\textwidth}
   \includegraphics[width=1\linewidth]{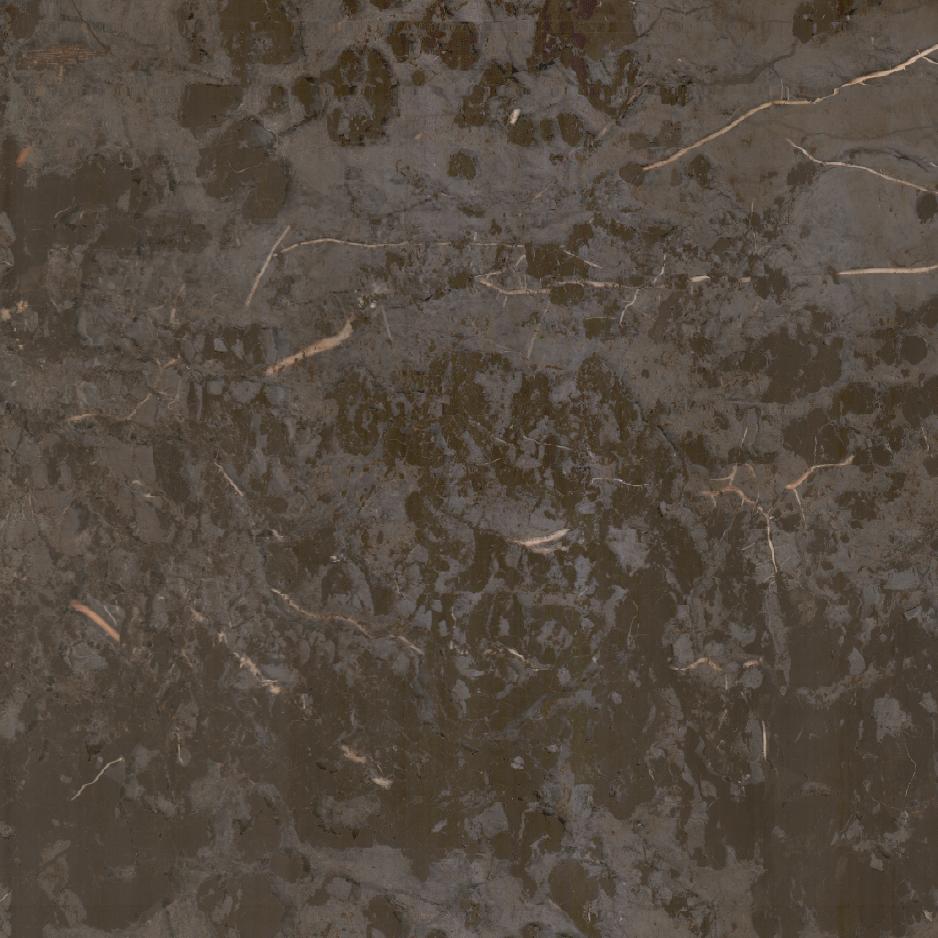}
   \caption{Pre-processed Image}
   \label{subfig:2_B}
\end{subfigure}
\begin{subfigure}[t]{0.15\textwidth}
   \includegraphics[width=1\linewidth]{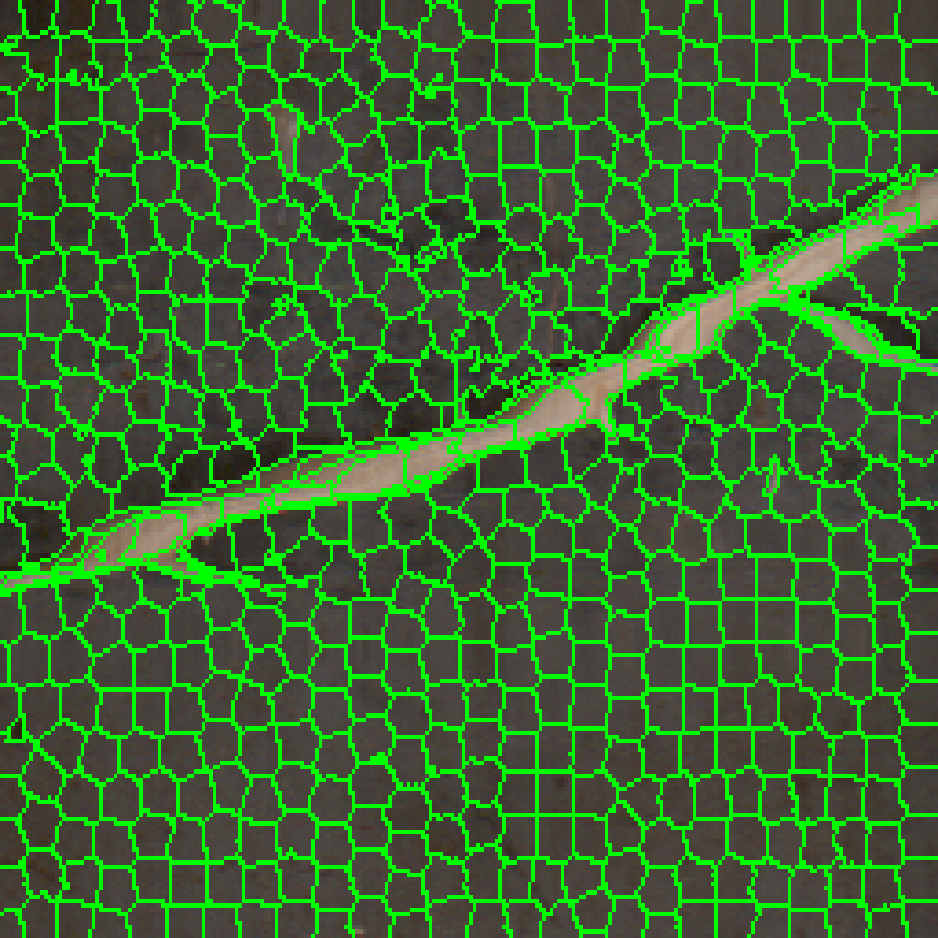}
   \caption{Superpixel segmentation of a sub-image}
   \label{subfig:2_C}
\end{subfigure}

   \caption{An example of the image pre-processing sequence: (\protect\subref{subfig:2_A}) shows the original minirhizotron image, (\protect\subref{subfig:2_B}) shows the pre-processed image after removing vertical striping noise, (\protect\subref{subfig:2_C}) is an example cropped sub-image illustrating the superpixel-segmentation}
\label{fig:2}
\end{center}
\end{figure}

The minirhizotron imagery collected for this study has vertical striping artifacts which is a common occurrence that happens when the numerous sensors in the scan head are slightly differently calibrated resulting in background stripes, or slightly darker or lighter hue bands as shown in Fig. \ref{subfig:2_A}. Pre-processing of the imagery to remove this striping noise \cite{chang2016remote} consists of subtracting the column mean from each column in the image and adding the global image mean, $\bar{I}$, back to the image as shown in Eq. \eqref{eq:five}: 
\begin{equation}\label{eq:five}
I_p(m,n,b) = I(m,n,b) - \frac{1}{m}\sum_{m} I(m,n,b) + \bar{I}(b)
\end{equation}

where $I(m,n,b)$ is the pixel in the $m^{th}$ row, $n^{th}$ column and $b^{th}$ band of image $I$, $\bar{I}(b)$ is the global mean of $I$ in band $b$, and $I_p$ is the resulting pre-processed image. An example of the pre-processed image is shown in Fig. \ref{subfig:2_B}.

After pre-processing, the images are oversegmented into \emph{superpixels} using the Simple Linear Iterative Clustering (SLIC) algorithm \cite{achanta2010slic}. Superpixels are groups of pixels which are spatially-contiguous and perceptually similar in color as shown in Fig. \ref{subfig:2_C}. This step helps to reduce the overall computational load since processing is done on the reduced superpixel level as opposed to the extremely large number of original pixels. 

After superpixel segmentation, the mean, variance, and entropy of each band in the RGB and LAB color spaces ([mean-R, mean-G, mean-B, mean-L, mean-a, mean-b, var-R, var-G, var-B, var-L, var-a, var-b, H-R, H-G, H-B, H-L, H-a, H-b]) are computed to make an 18-dimensional feature vector for each superpixel.  After feature extraction, the superpixel feature vectors of each image are scaled by their order of magnitude so that they are normalized for equal weight across the features, 
\begin{equation}\label{eq:six}
{x}(i) = \frac{F_i(I_p(m,n))}{{s}(i)}
\end{equation}
where $F_i$ is the $i^{th}$ feature generation function, ${s}(i)$ is the scaling value for the $i^{th}$ feature, and ${x}(i)$ is the scaled result for the $i^{th}$ feature. These scaled feature vectors are the features used within the MIL algorithms to differentiate between root superpixels and non-root superpixels.

\begin{figure}[t] 
\begin{center}
\begin{subfigure}{0.15\textwidth}
   \includegraphics[width=1\linewidth]{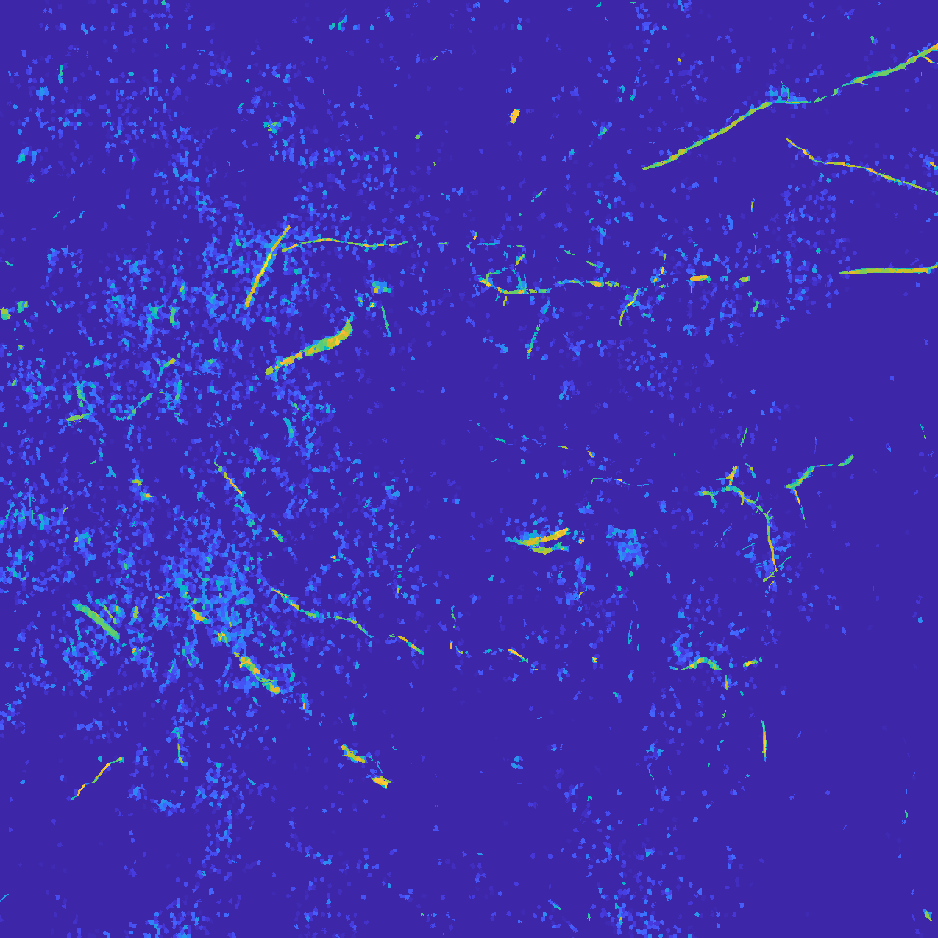}
   \caption{MI-ACE}
   \label{subfig:3_A}
\end{subfigure}
\begin{subfigure}{0.15\textwidth}
   \includegraphics[width=1\linewidth]{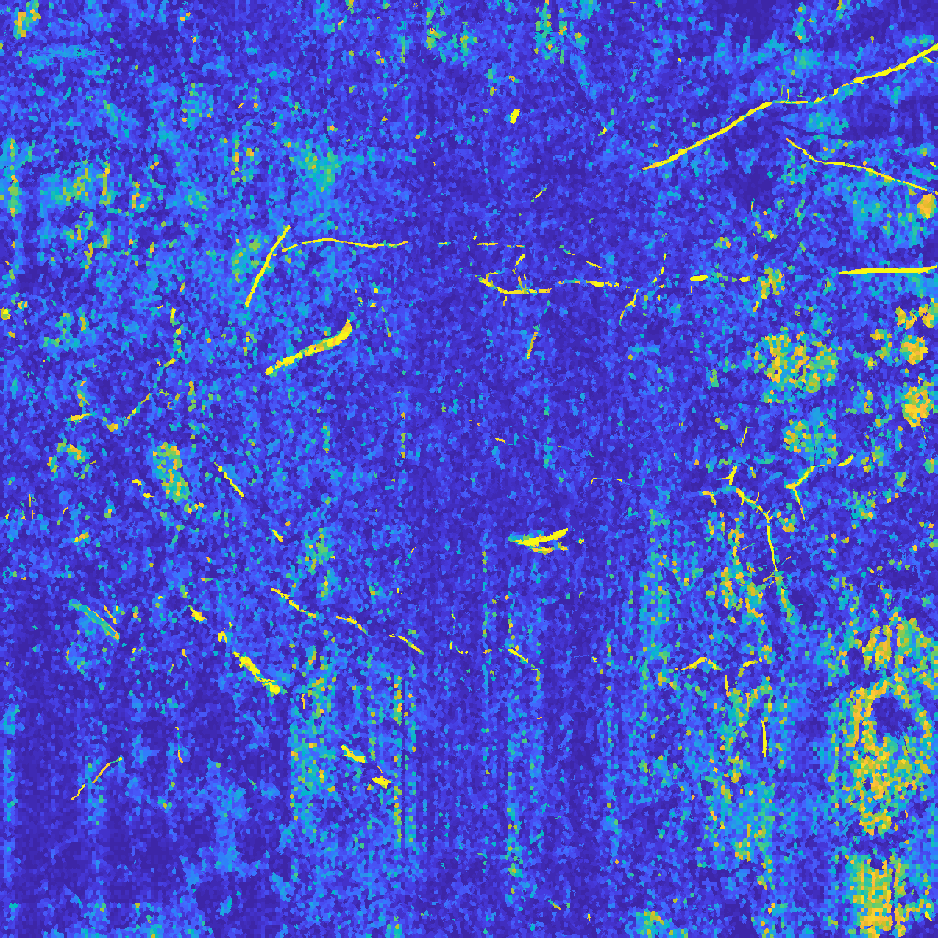}
   \caption{MIForests}
   \label{subfig:3_B}
\end{subfigure}
\begin{subfigure}{0.15\textwidth}
   \includegraphics[width=1\linewidth]{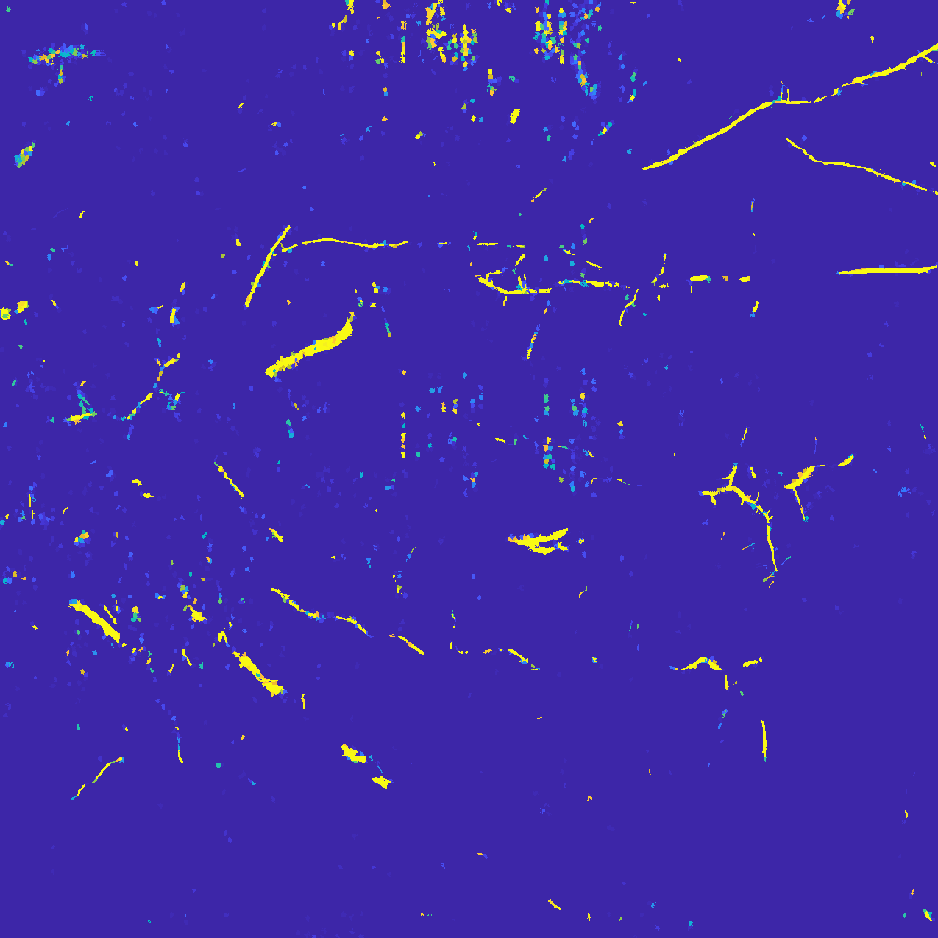}
   \caption{miSVM}
   \label{subfig:3_C}
\end{subfigure}
~~
\begin{subfigure}{0.15\textwidth}
   \includegraphics[width=1\linewidth]{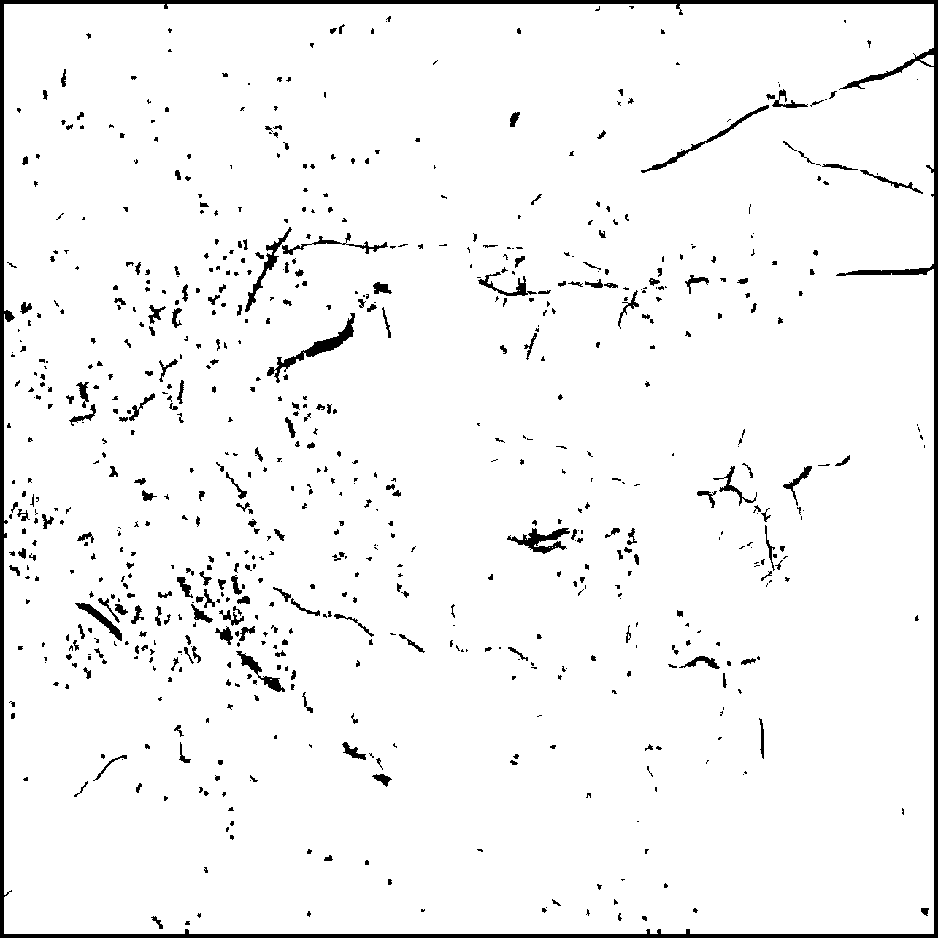}
   \caption{MI-ACE binary}
   \label{subfig:3_A}
\end{subfigure}
\begin{subfigure}{0.15\textwidth}
   \includegraphics[width=1\linewidth]{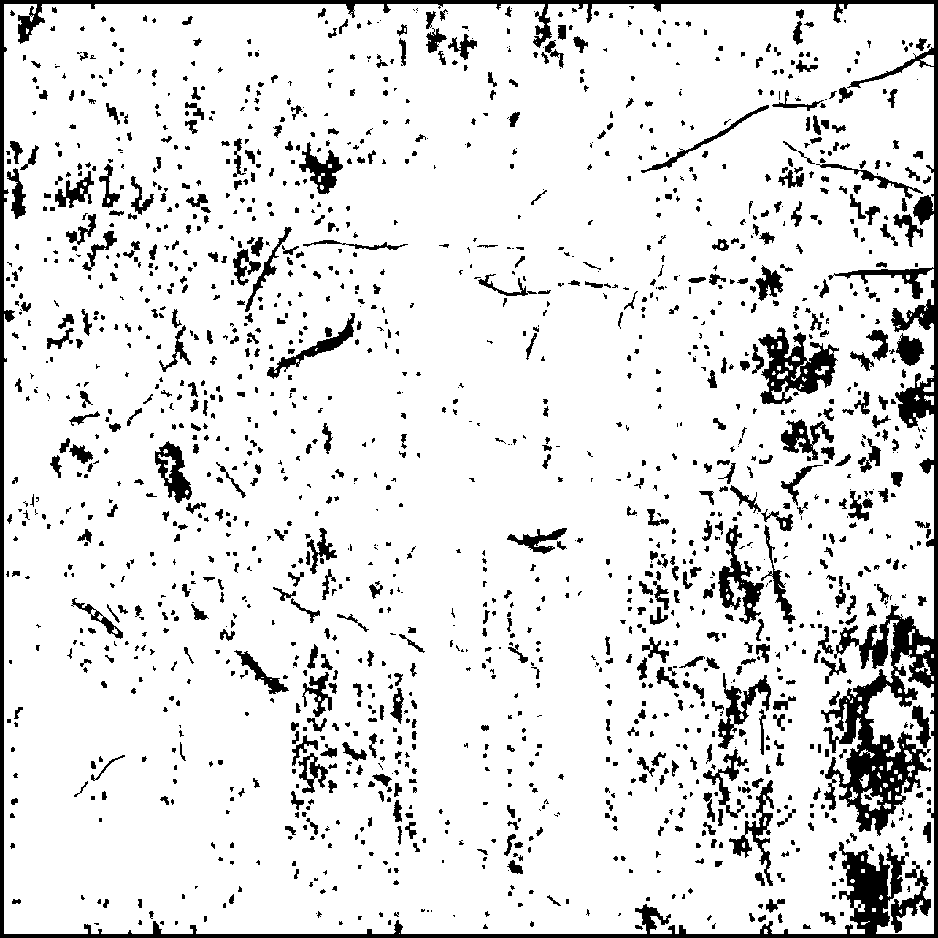}
   \caption{MIForests binary}
   \label{subfig:3_D}
\end{subfigure}
\begin{subfigure}{0.15\textwidth}
   \includegraphics[width=1\linewidth]{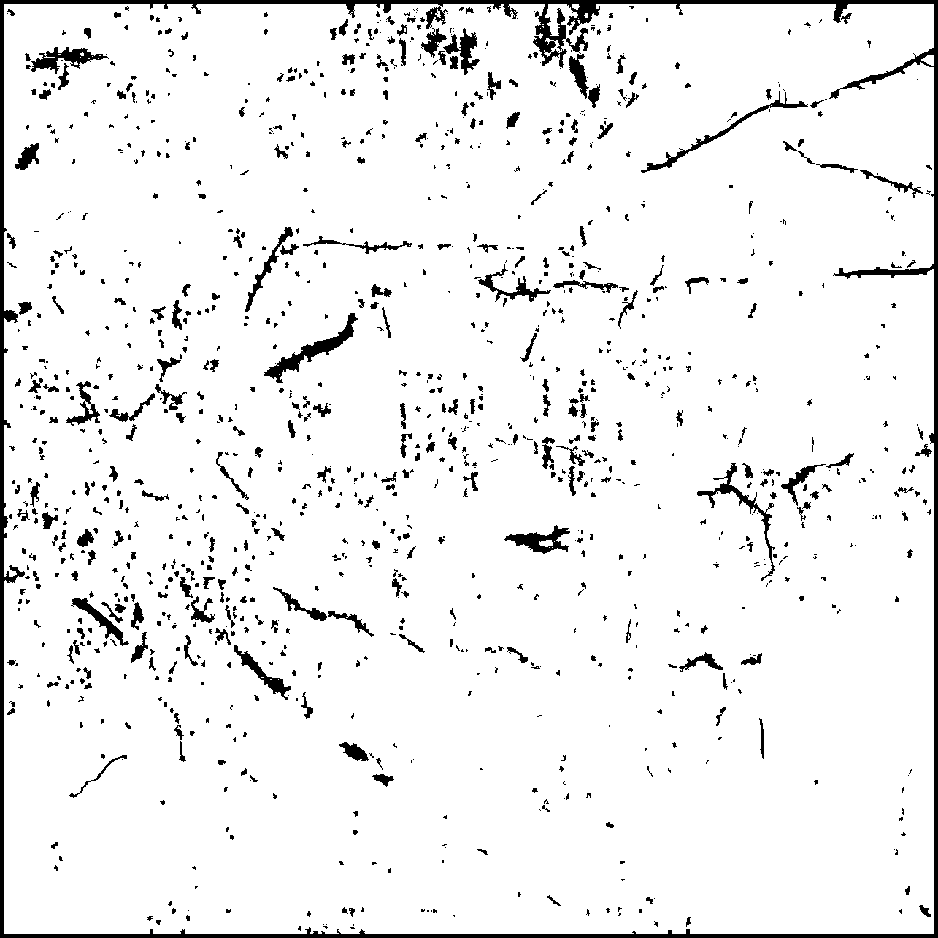}
   \caption{miSVM binary}
   \label{subfig:3_G}
\end{subfigure}
~~
\begin{subfigure}{0.15\textwidth}
   \includegraphics[width=1\linewidth]{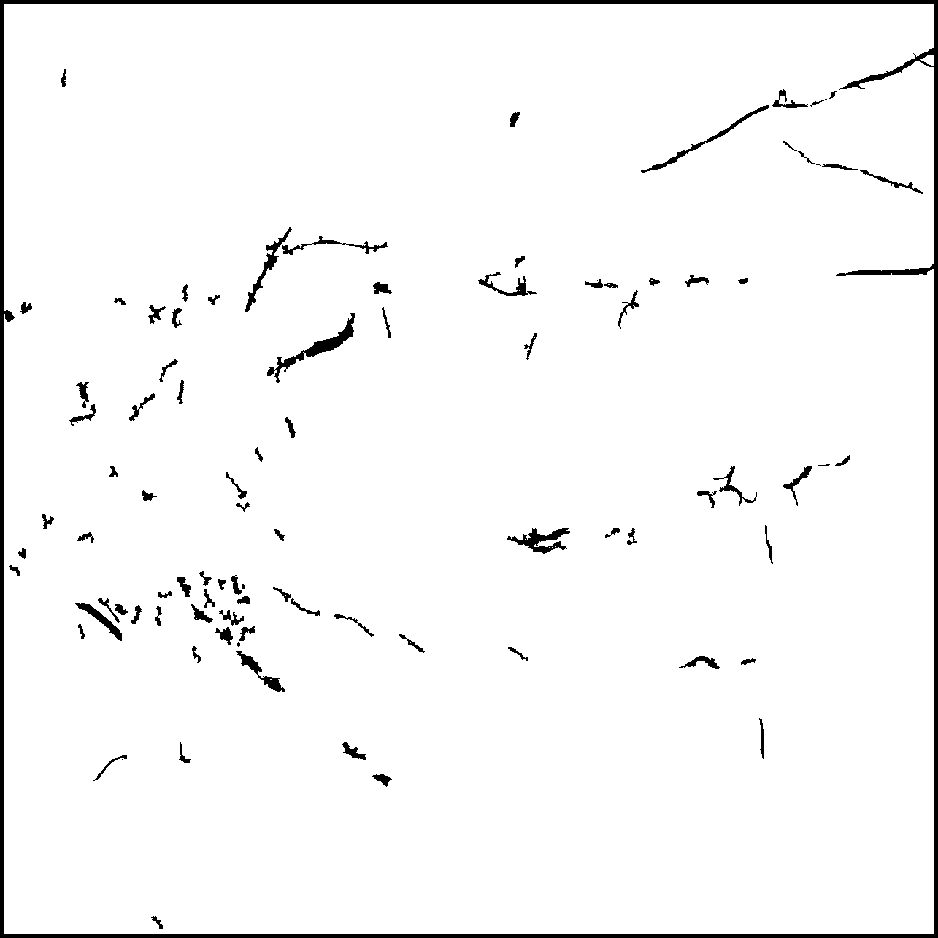}
   \caption{MI-ACE size}
   \label{subfig:3_B}
\end{subfigure}
\begin{subfigure}{0.15\textwidth}
   \includegraphics[width=1\linewidth]{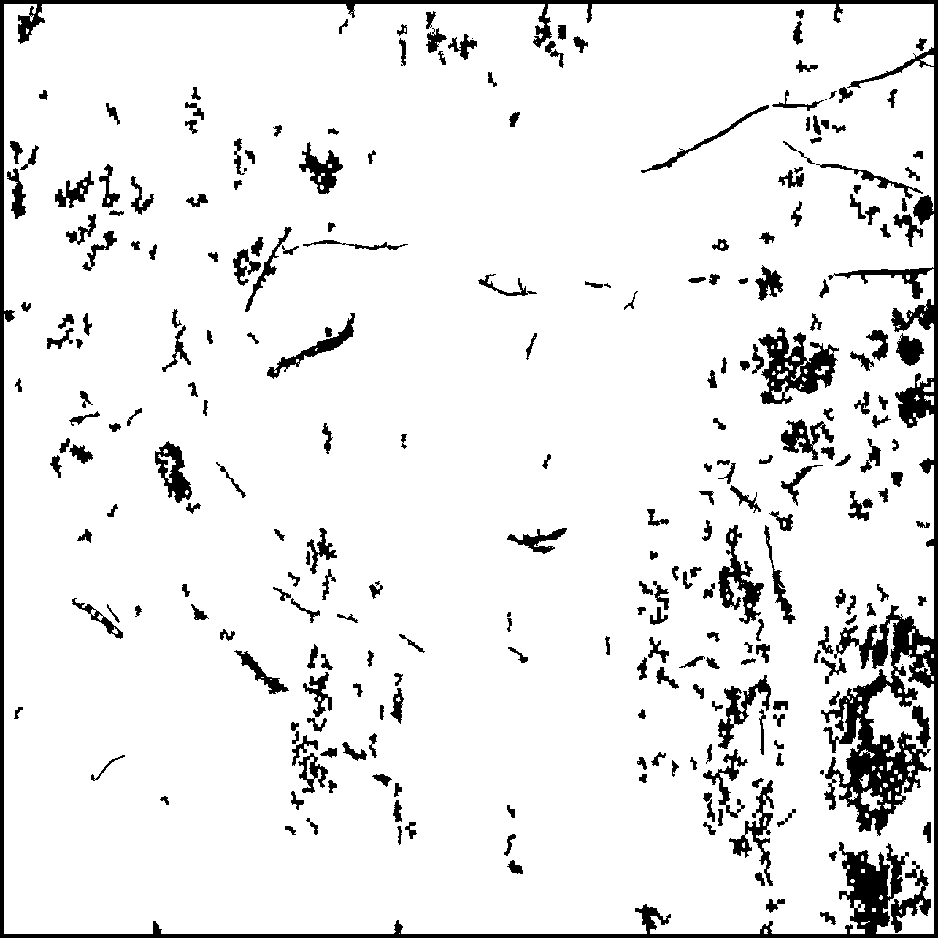}
   \caption{MIForests size}
   \label{subfig:3_E}
\end{subfigure}
\begin{subfigure}{0.15\textwidth}
   \includegraphics[width=1\linewidth]{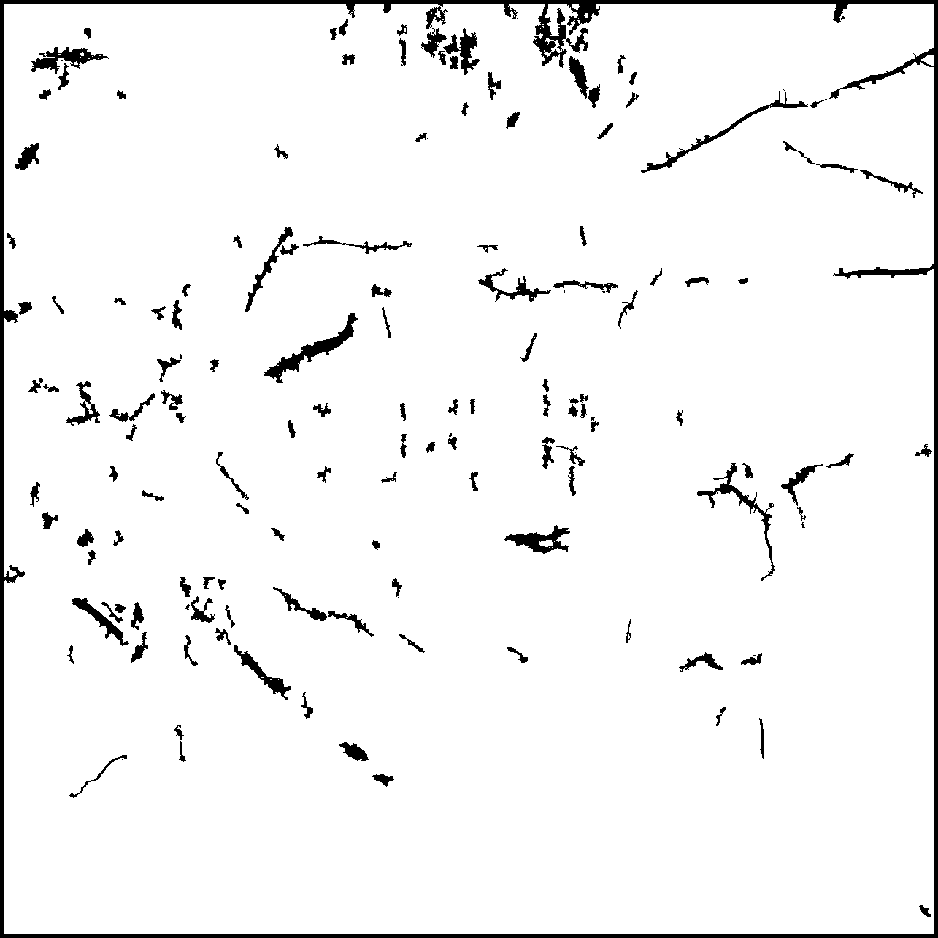}
   \caption{miSVM size}
   \label{subfig:3_H}
\end{subfigure}
~~
\begin{subfigure}{0.15\textwidth}
   \includegraphics[width=1\linewidth]{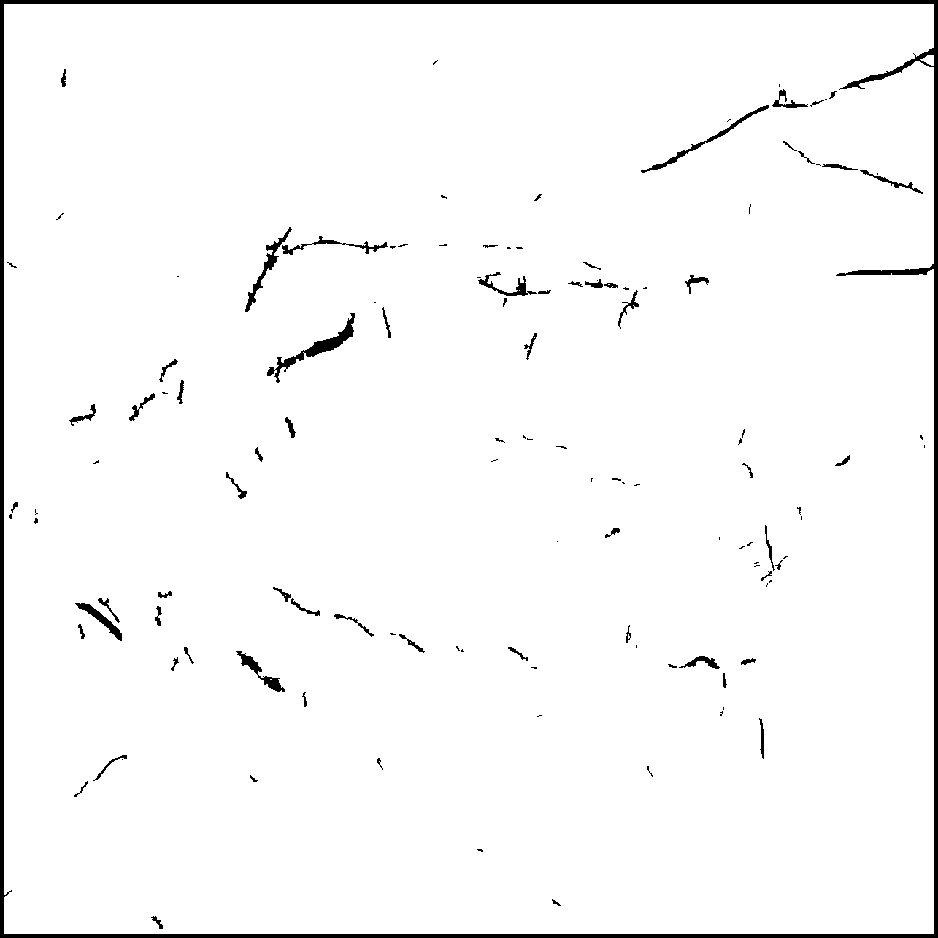}
   \caption{MI-ACE ecc}
   \label{subfig:3_C}
\end{subfigure}
\begin{subfigure}{0.15\textwidth}
   \includegraphics[width=1\linewidth]{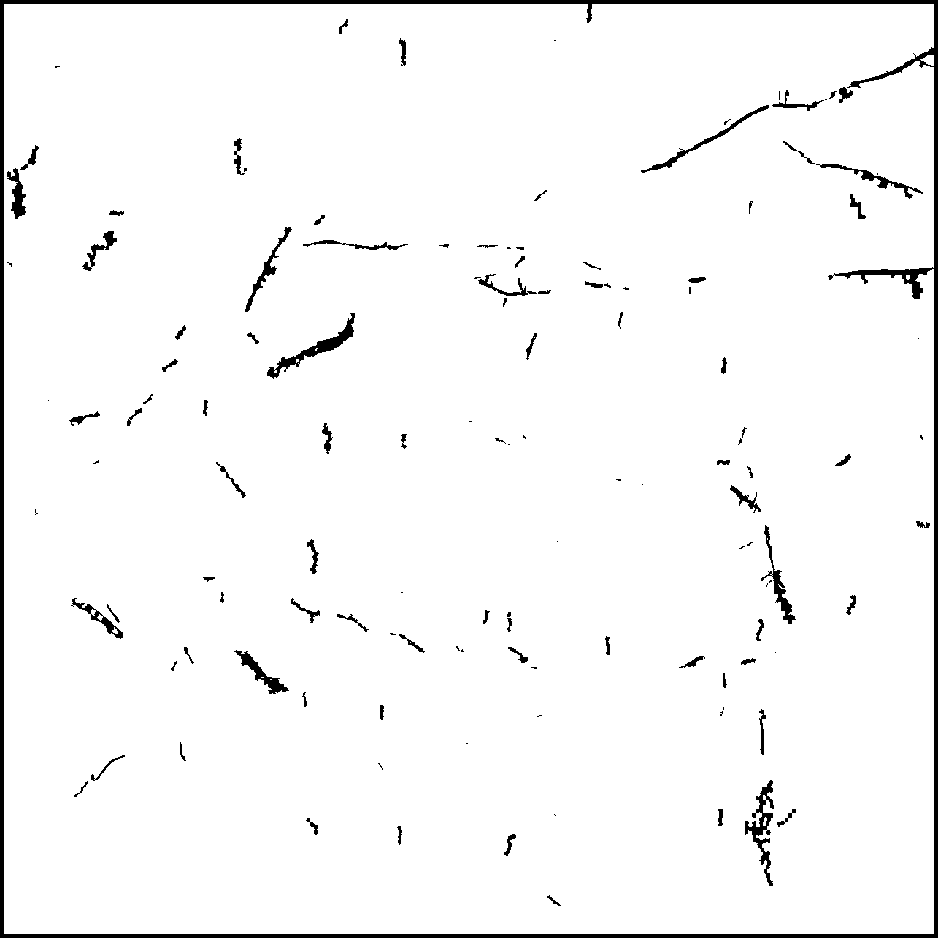}
   \caption{MIForests ecc}
   \label{subfig:3_F}
\end{subfigure}
\begin{subfigure}{0.15\textwidth}
   \includegraphics[width=1\linewidth]{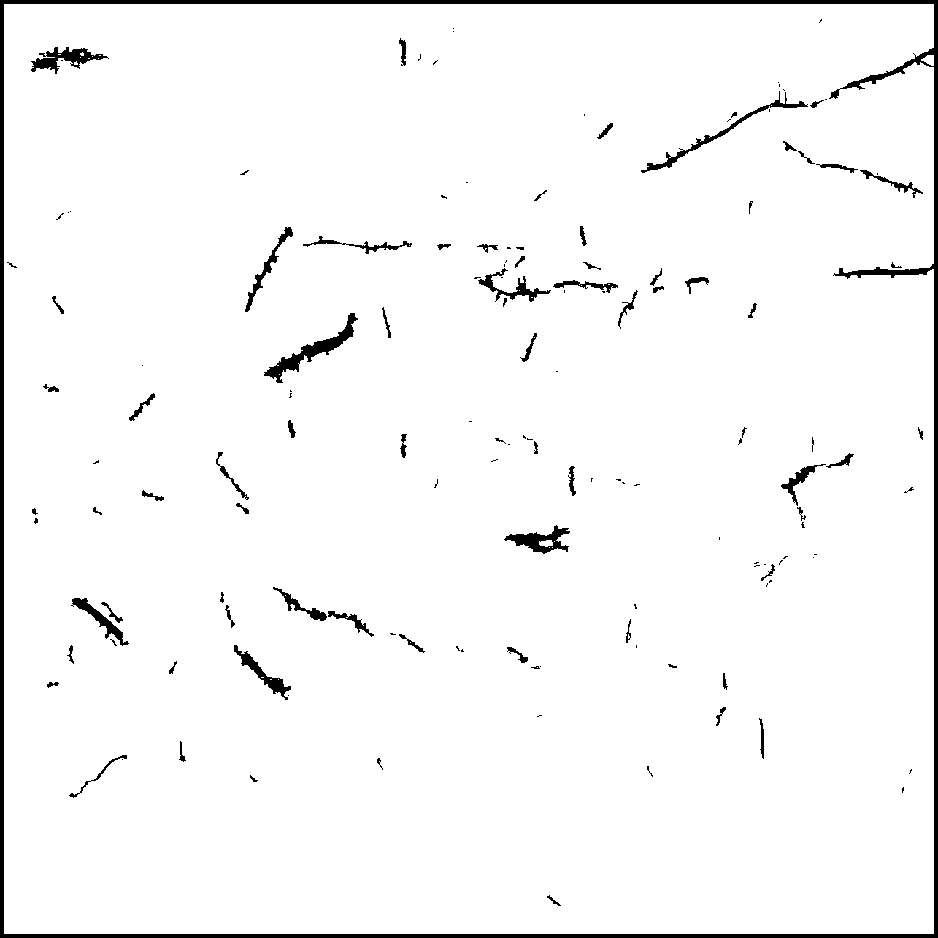}
   \caption{miSVM ecc}
   \label{subfig:3_I}
\end{subfigure}

   \caption{{MIL algorithm results corresponding to the image shown in Fig. \protect\ref{subfig:2_B}.} (\protect\subref{subfig:3_A})-(\protect\subref{subfig:3_C}) show the MI-ACE, MIForests, miSVM confidence maps, respectively.  The confidence map is the pixel-level output from the MIL algorithms indicating the confidence the algorithms assign to each pixel for belonging to the root class.  The color map for these confidences range from dark blue (very low confidence) to bright yellow (very high confidence).  (\protect\subref{subfig:3_A})-(\protect\subref{subfig:3_G}) are binarized results after thresholding MI-ACE, MIForests, and miSVM confidences maps, respectively. (\protect\subref{subfig:3_B})-(\protect\subref{subfig:3_H}) are the results after filtering the threshold results based on connected component size where components with too few pixels are removed. (\protect\subref{subfig:3_C})-(\protect\subref{subfig:3_I}) are the results after filtering based on connected component eccentricity where components that do not have sufficiently large enough eccentricity are removed}
\label{fig:3}
\end{center}
\end{figure}

\subsection{ Multiple Instance Learning for Root Segmentation }
In our MIL approach, each superpixel is considered an \textit{instance} (or data point) and each minirhizotron image corresponds to a bag. An image containing roots is a positive bag. Otherwise, it is a negative bag. To describe this more precisely, let $\mathbf{B} = \left\{\mathbf{B}_{1},\dotsc,\mathbf{B}_{K}\right\} $ be the set $K$ bags with label $L = \left\{L_{1},\dotsc,L_{K}\right\}$. $L_{i} \in \left\{0,1\right\}$, where 0 represents a negative bag and 1 represents a positive bag. Each bag, $\mathbf{B}_i$ contains $N_i$ superpixels, $\mathbf{B}_{i} = \left[\mathbf{x}_{i}^{1},\mathbf{x}_{i}^{2},\dotsc,\mathbf{x}_{i}^{N_{i}}\right].$ Each feature vector, $\mathbf{x}_{i}^{j} \in \mathbb{R}^{18}$ is computed on a superpixel as described in Section \ref{sec:preproc}.  A bag,  $\mathbf{B}_{i}$, is labeled as positive, $L_{i} = 1$, if there exists at least one feature vector in the bag corresponding to a root superpixel.

Three MIL algorithms are investigated for root detection and segmentation, the multiple instance adaptive cosine coherence estimator (MI-ACE) \cite{zare2018discriminative}, the multiple instance support vector machine (miSVM) \cite{andrews2003support}, and the multiple instance learning with randomized trees (MIForests) \cite{leistner2010miforests} algorithms.  
MI-ACE is a supervised multiple instance learning algorithm that estimates a discriminative target signature (in this case, root signatures) $\mathbf{s}$ during training and, then, uses this discriminative signature within the Adaptive Cosine Estimator detector \cite{broadwater2007hybrid,kraut1999cfar,kraut2001adaptive} to detect root in unlabeled test imagery. The advantage of MI-ACE (and other MIL concept learning methods \cite{maron1998framework,zhang2002dd,zare2010pattern,jiao2015functions,shrivastava2014dictionary,shrivastava2015generalized}) is that the estimated discriminative target signature can be examined to obtain insight into what characterizes the target (\ie root) class and can be easily interpreted to understand what root characteristics are being used to detect the roots and distinguish them from the background (\ie soil).  The miSVM algorithm iteratively trains a support vector machine \cite{cortes1995support} classifier while enforcing the multiple instance learning constraints that each positive bag contains a target and each negative  bag does not contain any target. MIForests uses a deterministic annealing approach to optimize an objective function that enforces the MIL constraints on the data while training a random forest classifier.  MIForests initially trains a random forest classifier using all data instances and each instance is labeled the same as its bag label. Then, iteratively refines the instance labels while optimizing the MIForests objective function. 

We trained the three MIL algorithms with pre-processed minirhizotron images with image-level labels. The trained models were then used to predict the roots in other images and to determine the degree of confidence of those predictions as illustrated in Fig. \ref{fig:3} (\protect\subref{subfig:3_A})-(\protect\subref{subfig:3_C}) (see Experimental Setup section for details).

\subsection{Post-processing}
\label{sec:postproc}
After computing the confidence map, some post-processing is applied to reduce the number of false detections. First, the confidence map is binarized by thresholding as shown in Fig. \ref{fig:3} (\protect\subref{subfig:3_A})-(\protect\subref{subfig:3_G}). After thresholding, root shape and size characteristics are used to filter out likely false detections.  The size and eccentricity are computed on each connected component made by pixels identified as root. Connected components with a small size and a small eccentricity are removed from the binarized image. The results after post-processed by size are shown in Fig. \ref{fig:3} (\protect\subref{subfig:3_B})-(\protect\subref{subfig:3_H}). The results after post-processing by eccentricity are shown in Fig. \ref{fig:3} (\protect\subref{subfig:3_C})-(\protect\subref{subfig:3_I}).

\begin{figure}[h] 
\begin{center}
\begin{subfigure}[b]{0.23\textwidth}
   \includegraphics[width=1\linewidth]{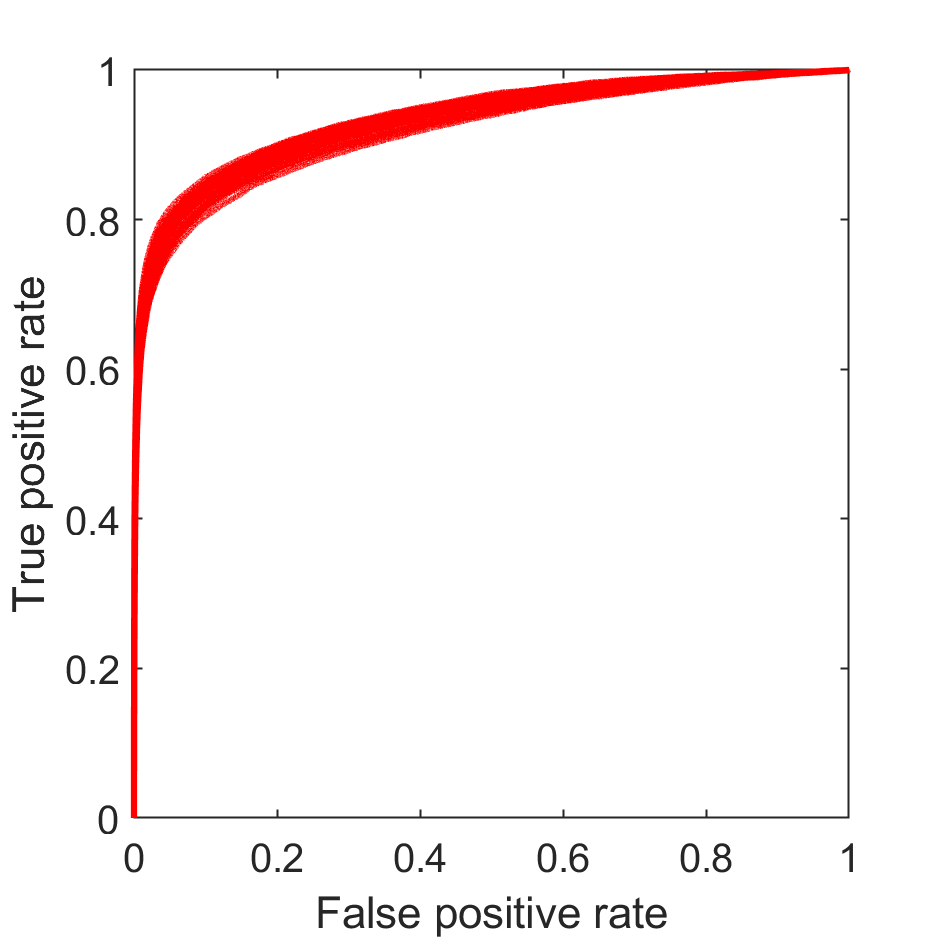}
   \caption{MI-ACE}
   \label{subfig:4_A}
\end{subfigure}
\begin{subfigure}[b]{0.23\textwidth}
   \includegraphics[width=1\linewidth]{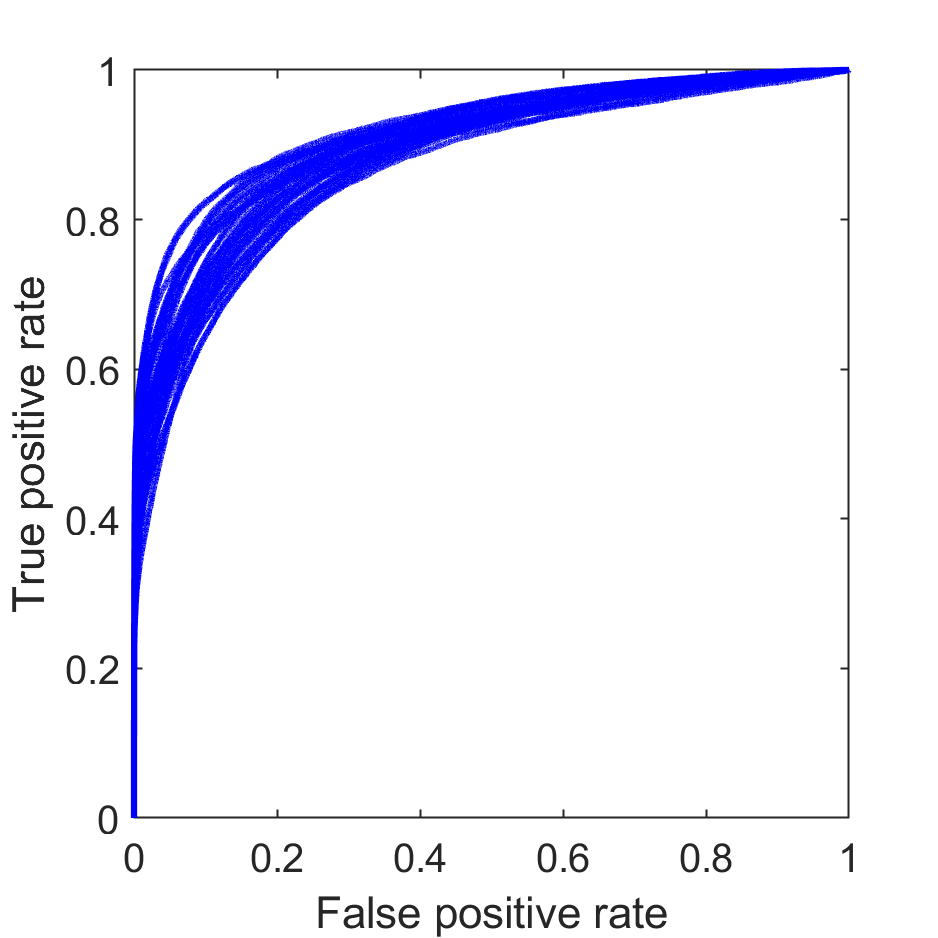}
   \caption{MIForests}
   \label{subfig:4_B}
\end{subfigure}
~~
\begin{subfigure}[b]{0.23\textwidth}
   \includegraphics[width=1\linewidth]{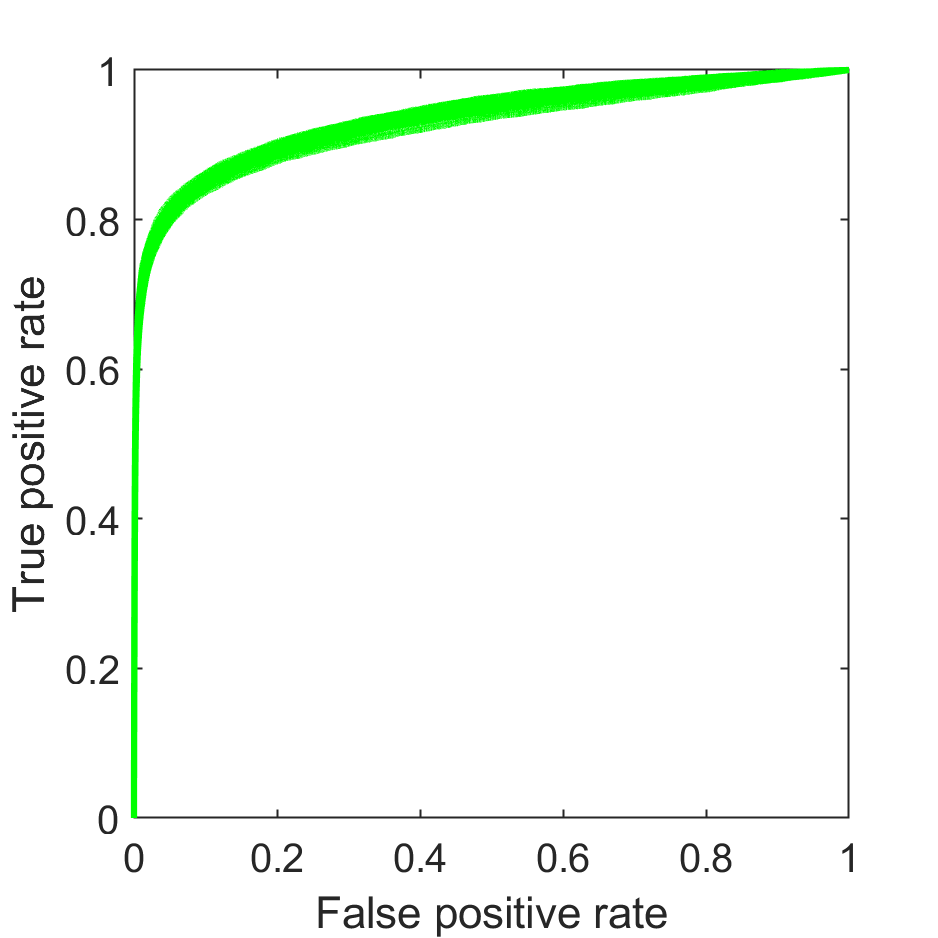}
   \caption{miSVM}
   \label{subfig:4_C}
\end{subfigure}
\begin{subfigure}[b]{0.23\textwidth}
   \includegraphics[width=1\linewidth]{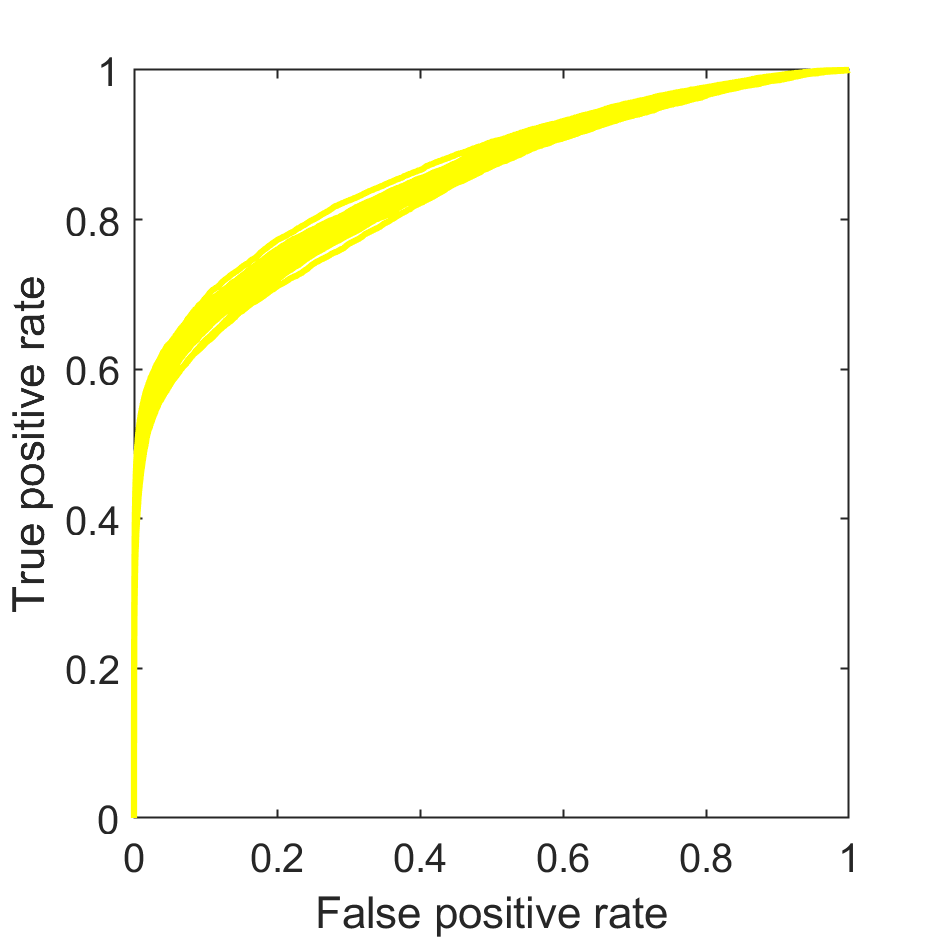}
   \caption{RF}
   \label{subfig:4_D}
\end{subfigure}
~~
\begin{subfigure}[b]{0.23\textwidth}
   \includegraphics[width=1\linewidth]{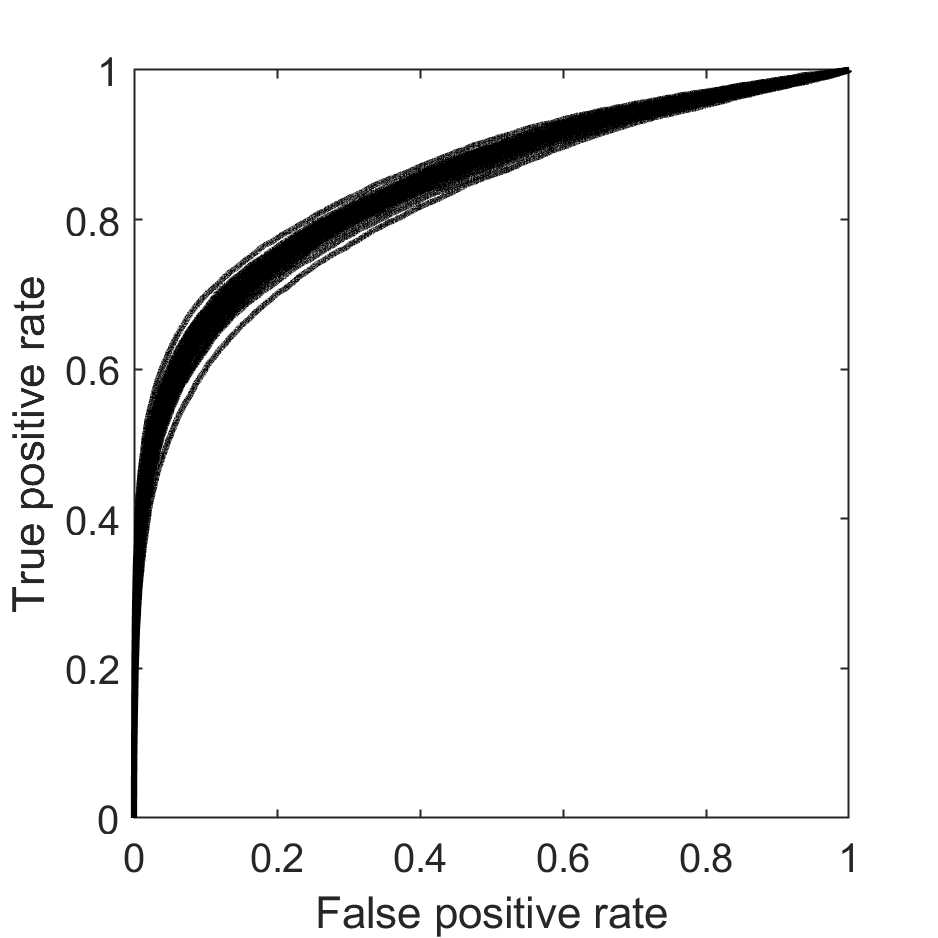}
   \caption{SVM}
   \label{subfig:4_E}
\end{subfigure}
\begin{subfigure}[b]{0.23\textwidth}
   \includegraphics[width=1\linewidth]{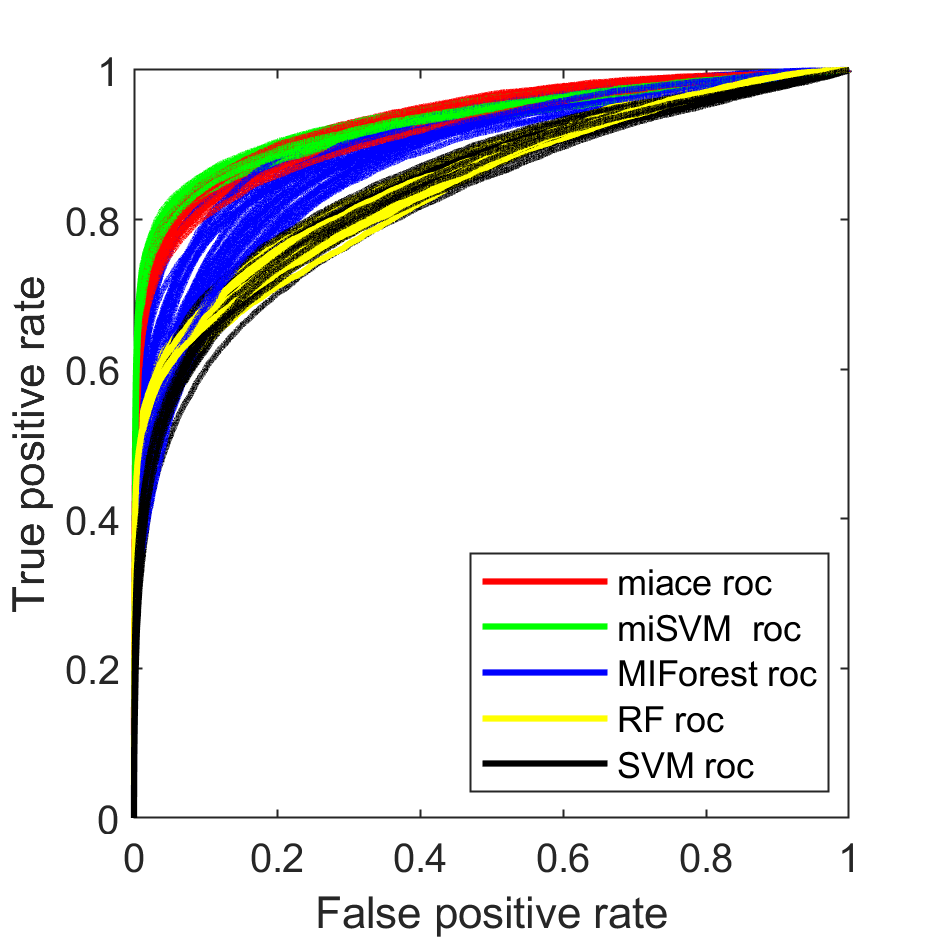}
   \caption{Comparison}
   \label{subfig:4_F}
\end{subfigure}

   \caption{Root detection results with different methods across 30 runs without post-processing. (\protect\subref{subfig:4_A}) ROC curves of MI-ACE. (\protect\subref{subfig:4_B}) ROC curves of MIForests. (\protect\subref{subfig:4_C}) ROC curves of miSVM.  (\protect\subref{subfig:4_D}) ROC curves of RF. (\protect\subref{subfig:4_E}) ROC curves of SVM. (\protect\subref{subfig:4_F}) Comparison of ROC curves across different methods}
\label{fig:4}
\end{center}
\end{figure}

\section{Experiments}
\subsection{Data Description}
Switchgrass (\textit{Panicum virgatum} L.) is recognized as the most promising perennial grass for bioenergy production in the U.S. Nonetheless, despite being the subject of abundant and multidisciplinary research, there is a lack of understanding of the contributions of the switchgrass root system to the adaptability to different environments, biomass productivity, and substantial ecosystem services \cite{schmer2008net}. The extensive root system of switchgrass is increasingly recognized to contribute to long-term carbon sequestration, high resource use efficiency, and high biomass productivity \cite{comas2013root,de2013variation}. However, despite the essential biological function, information about switchgrass root system dynamics in space and time is limited, at least in part because of the challenges associated with access to the soil-root interface without disturbing the soil environment, which can impair subsequent measurements of the same plant \cite{berhongaray2013fine}. 

Switchgrass root images were collected from minirhizotron access tubes that were installed in a 2-y old switchgrass field at the U.S. Department of Energy National Environmental Research Park at Fermilab in Batavia, IL, USA. Minirhizotron access tubes measuring 1.82 m in length were installed at 60° off the horizontal axis using an angled, guided hydraulic soil core sampler. The tubes were inserted to reach a maximum vertical depth of approximately 1.2 m, and foam caps were installed on the end protruding from the soil to insulate the tubes, block light, and protect them from UV damage. Images were collected by inserting a CI-602 in-situ root imager (CID BioScience, Camas, WA, USA) in to the minirhizotron tubes. This scanner was set to collect 360-degree images at 300 dpi in 28 cm depth increments.  

\subsection{Experimental Setup}
The pixel level segmentation was estimated with MIL algorithms (MI-ACE, miSVM, MIForests) using training imagery with only image level labels. The training data for this study contained 34 positive bags and 34 negative bags. Each bag contains instances from an image, and the bag label is the same as the image label. 

To help reduce computational complexity, a subset of the instances in each bag were selected and used within the MIL algorithms. These instances were selected in such a way to ensure a wide distribution of root and non-root instances with varying color properties.  Specifically, a 200-bin histogram of the green channel values was constructed using the feature vectors across all superpixels.  Then, one superpixel corresponding to each non-empty histrogram bin was selected at random (uniform random selection) and kept for processing.  If all bins are non-empty, this results in 200 instances per bag. Three multiple instance learning models: MI-ACE, miSVM, and MIForests were trained multiple times. Each time, the instances in each bag were randomly selected again using the approach described above and the same training data was used for all three models. 

For each method, parameters were tuned to maximize performance. Performance was measured using the F-score \cite{powers2011evaluation} on the validation data set. 
For the miSVM algorithm, a radial basis function (RBF) kernel was used and the parameters were set to be $C = 10$ and $\gamma = 1$.  These parameters were determined by varying the RBF kernel width, $\gamma$, from $2^{-15}$ to $2^{3}$ by a factor of $4$. Similarly, the parameter $C$ was varied from $2^{-5}$ to $2^{15}$ by a factor of $4$. The parameter settings with the best performance over this range were selected as the parameters to be used for comparison. For MIForests, the parameters were set to be $t = 100$ and $x_d = 4$. These parameters were selected by varying the number of trees $t$ from $2^{4}$ to $2^{13}$ by a factor of $2$ and the dimension of the subset of input variables $x_d$ was varied from $2$ to $16$ by a factor of $2$. All other parameters of MIForests are set to be the default value. MI-ACE did not have parameters that required tuning. After training and validation, the models were tested by a set of images with pixel level labels.

\subsection{Root Detection Results}
\noindent\textbf {MIL methods}: 
MIL methods (MI-ACE, MIForests, and miSVM) were run 30 times and the root detection results were evaluated by Receiver Operating Characteristic (ROC) curves, shown in Fig. \ref{fig:4} (\protect\subref{subfig:4_A})-(\protect\subref{subfig:4_C}). The mean and variance of the true positive rate (TPR) of different methods at false positive rate (FPR) ranging from 0.01 to 0.06 are presented in Table \ref{table:rootDet}. It was found that miSVM outperformed the other approaches with a higher average TPR at the same FPR while MI-ACE was a close second. The performance of miSVM was also more consistent than that of the others.  Namely, miSVM had the least variance in performance.  In contrast, the MIForests variance was one order of magnitude larger than the others, indicating that the MIForests approach was more sensitive to samples selected in each bag than the others.

\noindent\textbf {Comparison with non-MIL methods}: 
The results of MI-ACE, MIForests, and miSVM were compared with non-MIL methods (support vector machine and random forests). The training data used to run the support vector machine (SVM) \cite{CC01a} and random forests (RF) \cite{breiman2001random} methods were the same as those instances in bags used to run miSVM and MIForest, and each instance is assigned the bag label. The parameters of SVM and RF were also set to be the same as miSVM and MIForests. Both SVM and RF were run 30 times. Fig. \ref{subfig:4_D} and \ref{subfig:4_E} are the ROC curves of RF and SVM which are compared with MI-ACE, MIForests and miSVM in Fig. \ref{subfig:4_F}. 

\begin{table*}[h]
\begin{center}
\caption{Mean and variance of TPR of comparison algorithms at FPR varying from 0.01 to 0.06}
\label{table:rootDet}
\begin{tabular}{|l|*{11}{c|}}
\hline
FPR & \multicolumn{2}{|c|}{MI-ACE TPR} & \multicolumn{2}{|c|}{MIForests TPR} & \multicolumn{2}{|c|}{miSVM TPR} & \multicolumn{2}{|c|}{SVM TPR} & \multicolumn{2}{|c|}{RF TPR} \\
\hline
    & mean &var & mean &var & mean &var & mean &var & mean &var\\
\hline\hline
0.01 & \underline{0.66} & \underline{0.38e-03} & 0.49 & 3.7e-03 & \textbf{0.70} & \textbf{0.21e-03} & 0.40 & 0.87e-03 & 0.51 & 0.47e-03 \\
0.02 & \underline{0.72} & \underline{0.33e-03} & 0.54 & 4.1e-03 & \textbf{0.75} & \textbf{0.12e-03} & 0.46 & 0.74e-03 & 0.55 & 0.33e-03 \\
0.03 & \underline{0.75} & \underline{0.32e-03} & 0.58 & 4.1e-03 & \textbf{0.78} & \textbf{0.10e-03} & 0.52 & 0.67e-03 & 0.58 & 0.26e-03 \\
0.04 & \underline{0.77} & \underline{0.29e-03} & 0.61 & 4.0e-03 & \textbf{0.80} & \textbf{0.08e-03} & 0.55 & 0.61e-03 & 0.60 & 0.23e-03 \\
0.05 & \underline{0.79} & \underline{0.25e-03} & 0.63 & 3.8e-03 & \textbf{0.81} & \textbf{0.07e-03} & 0.57 & 0.55e-03 & 0.61 & 0.20e-03 \\
0.06 & \underline{0.80} & \underline{0.23e-03} & 0.66 & 3.6e-03 & \textbf{0.82} & \textbf{0.06e-03} & 0.59 & 0.52e-03 & 0.63 & 0.22e-03 \\
\hline
\end{tabular}
\end{center}
\end{table*}

The ROC curves show that non-MIL methods can not compete with MIL methods when dealing with imprecisely labeled root data (i.e., labeled at the image level). The ROC curves of non-MIL methods were below the ROC curves of the MIL methods which means that at the same FPR value, the average TPR of non-MIL methods was less than the TPR value of MIL methods. For the case of FPR = 0.03 in Table \ref{table:rootDet}, the average TPR of miSVM was 0.78 which detected $26\%$ more roots than the SVM method and $20\%$ more roots than the RF method. The average TPR of MI-ACE was 0.75 which detected $23\%$ more roots than the SVM method and $17\%$ more roots than the RF method. However, the average TPR of MIForests was similar to the average TPR of RF because the variance of MIForests TPR was one order of magnitude greater than that of non-MIL methods, and some of the ROC curves of the MIForests method overlapped with the RF method ROC curves. Therefore, the average performance of MIForests was similar to the RF. 

\begin{figure}[hbt!] 
\begin{center}
\begin{subfigure}[b]{0.23\textwidth}
   \includegraphics[width=1\linewidth]{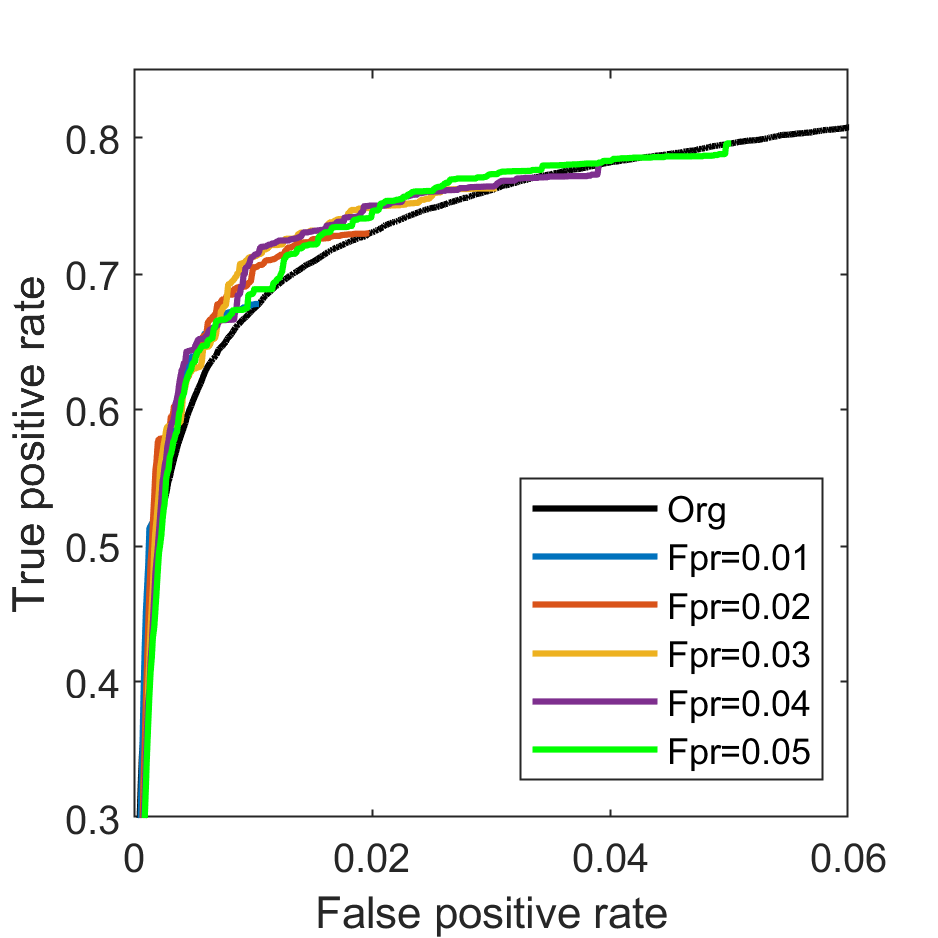}
   \caption{MI-ACE ecc}
   \label{subfig:5_A}
\end{subfigure}
\begin{subfigure}[b]{0.23\textwidth}
   \includegraphics[width=1\linewidth]{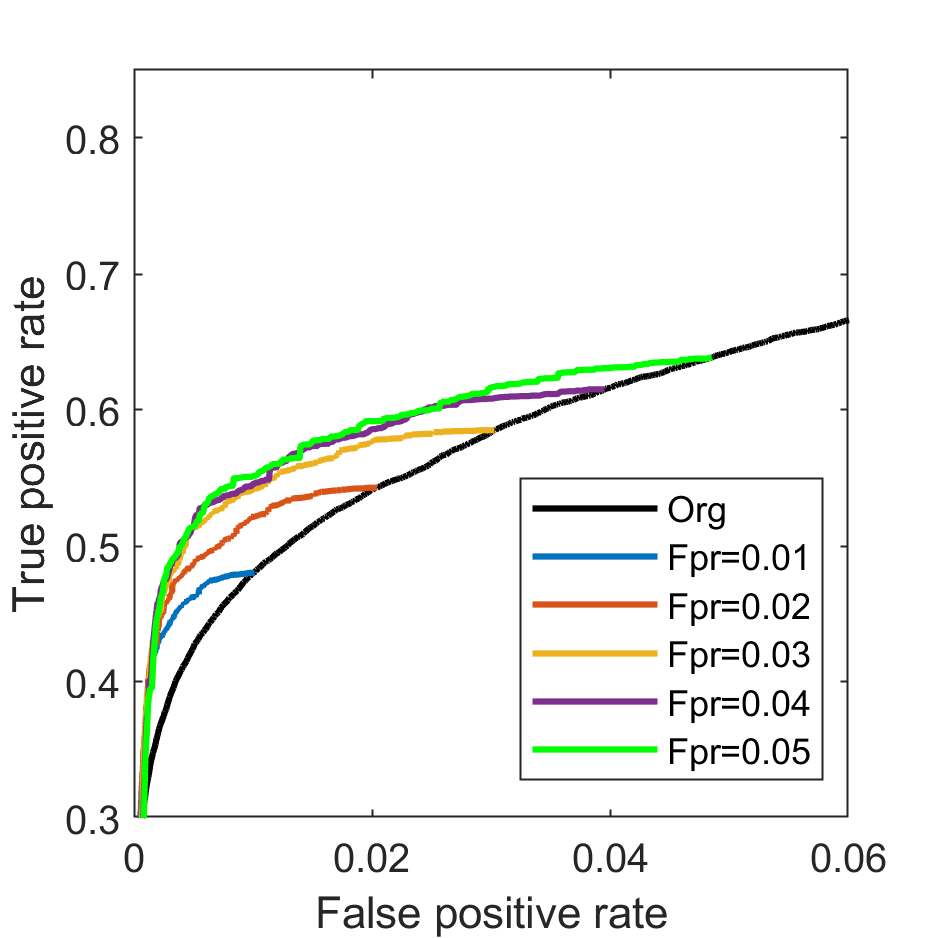}
   \caption{MIForests ecc}
   \label{subfig:5_B}
\end{subfigure}
~~
\begin{subfigure}[b]{0.23\textwidth}
   \includegraphics[width=1\linewidth]{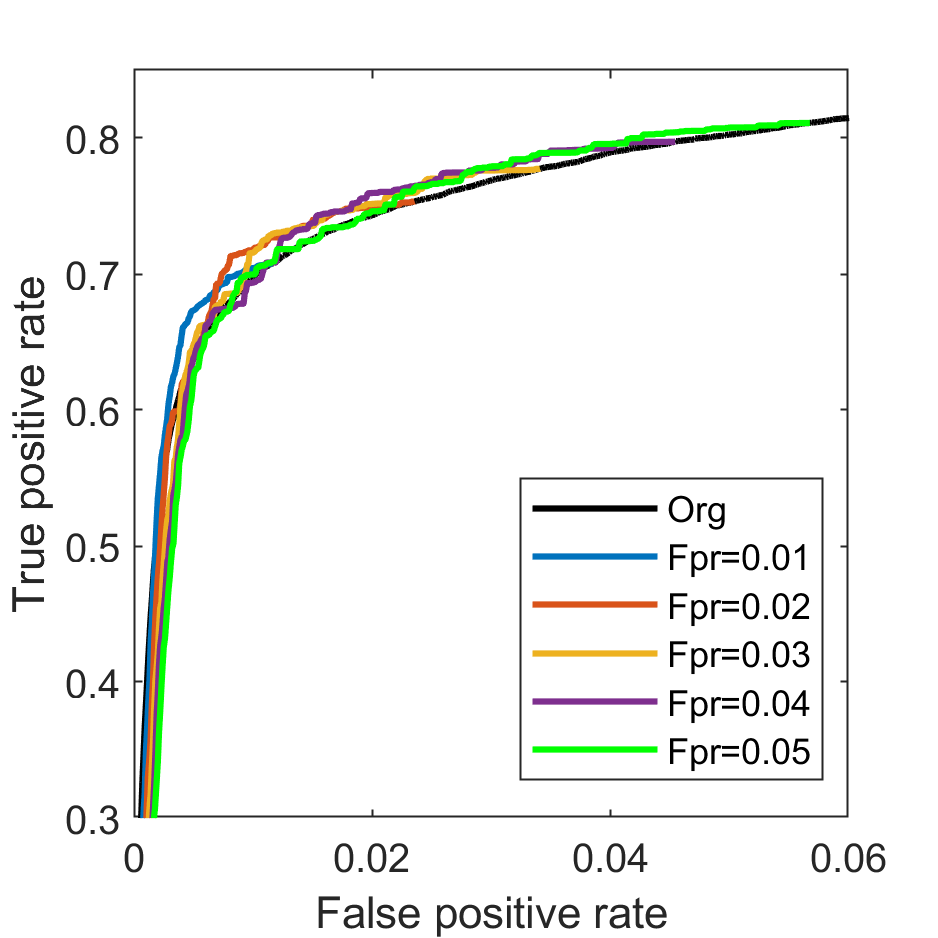}
   \caption{miSVM ecc}
   \label{subfig:5_C}
\end{subfigure}
\begin{subfigure}[b]{0.23\textwidth}
   \includegraphics[width=1\linewidth]{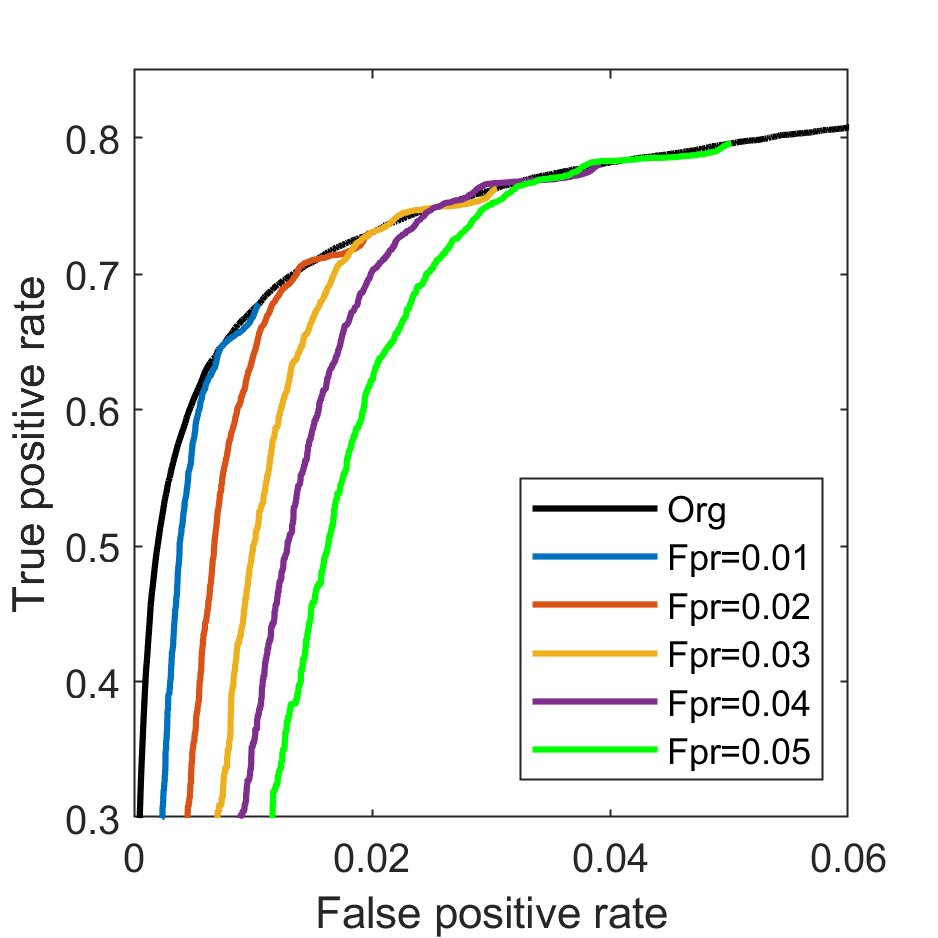}
   \caption{MI-ACE size}
   \label{subfig:5_D}
\end{subfigure}
~~
\begin{subfigure}[b]{0.23\textwidth}
   \includegraphics[width=1\linewidth]{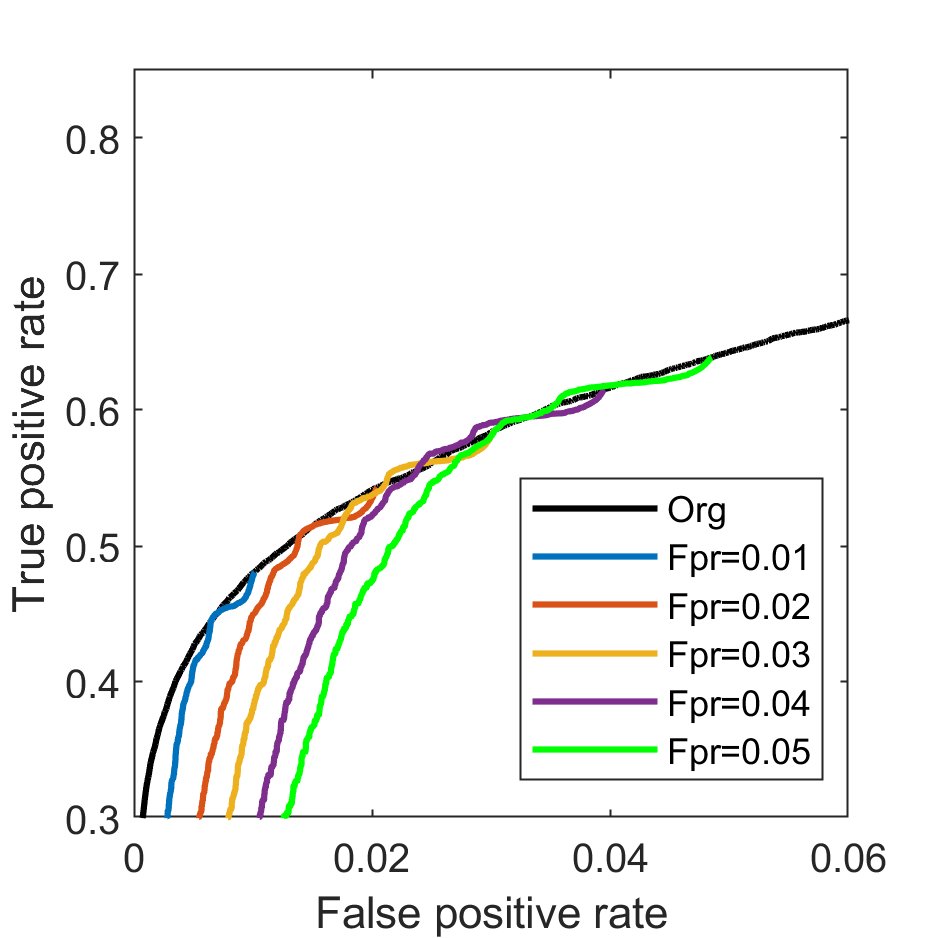}
   \caption{MIForests size}
   \label{subfig:5_E}
\end{subfigure}
\begin{subfigure}[b]{0.23\textwidth}
   \includegraphics[width=1\linewidth]{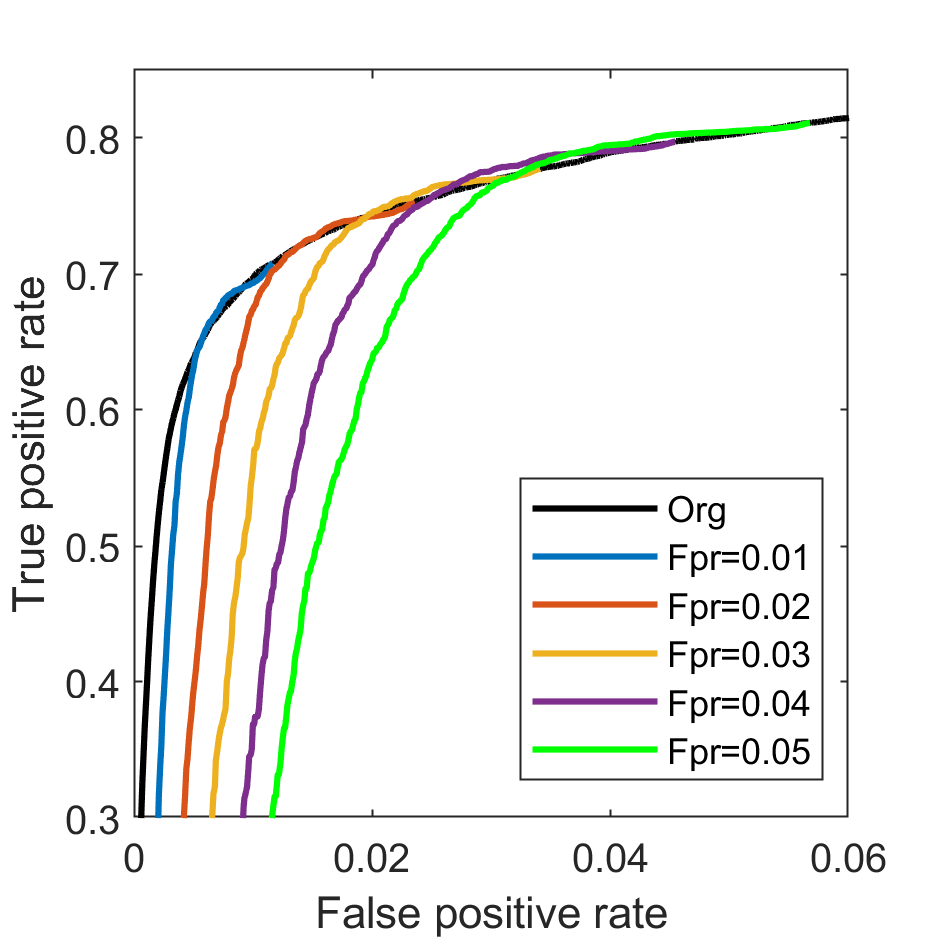}
   \caption{miSVM size}
   \label{subfig:5_F}
\end{subfigure}

   \caption{ROC curves after post-processing. Results are shown after thresholding confidence maps at FPR rates ranging from 0.01 to 0.05. After thresholding, results are filtered by removing connected components with either too few pixels in size or too small of an eccentricity score. The Org ROC curve is the ROC curve without post-processing. (\protect\subref{subfig:5_A})-(\protect\subref{subfig:5_C}) show ROC curves of post-process with eccentricity computed from connected components in the the binarized confidence maps of MI-ACE, MIForests, and miSVM. These ROC curves are generated by varying the eccentricty threshold.  (\protect\subref{subfig:5_D})-(\protect\subref{subfig:5_F}) show ROC curves of post-process with size computed from connected components in the binarized confidence maps of MI-ACE, MIForests, and miSVM. These ROC curves are generated by varying the size threshold}
\label{fig:5}
\end{center}
\end{figure}

\noindent\textbf {Post-processing results}:
Results were also evaluated after post-processing. The confidence maps from each image were thresholded at FPR values ranging from 0.01 to 0.05. These binarized maps were constructed by filtering with an eccentricity or size value lower than a fixed threshold to remove connected components. The threshold was varied to produce the ROC curves shown in Fig. \ref{fig:5}. In the post-processing step, it was important to decrease the FPR without decreasing TPR too much so that false roots were removed from binarized maps and root objects were not affected. The eccentricity can effectively decrease FPR and keep TPR changing slowly such that at the same FPR, the TPR on post-processing ROC curves using eccentricity was larger than the TPR on Org ROC curves as shown in Fig. \ref{fig:5} (\protect\subref{subfig:5_A})-(\protect\subref{subfig:5_C}). This indicated that the root objects had larger eccentricity than non-root objects. Size can also be used to remove false roots. The ROC curves using size to remove likely false roots show that root objects had larger size than non-root objects in Fig. \ref{fig:5} (\protect\subref{subfig:5_D})-(\protect\subref{subfig:5_F}). When the threshold on size was large, the post-processing ROC curves dropped steeply indicating that too many root objects were removed because the size of these root objects were smaller than the corresponding threshold. This experiment proved that eccentricity and size were useful attributes to separate roots from non-root objects.

\begin{figure}[h] 
\begin{center}
   \includegraphics[width=1\linewidth]{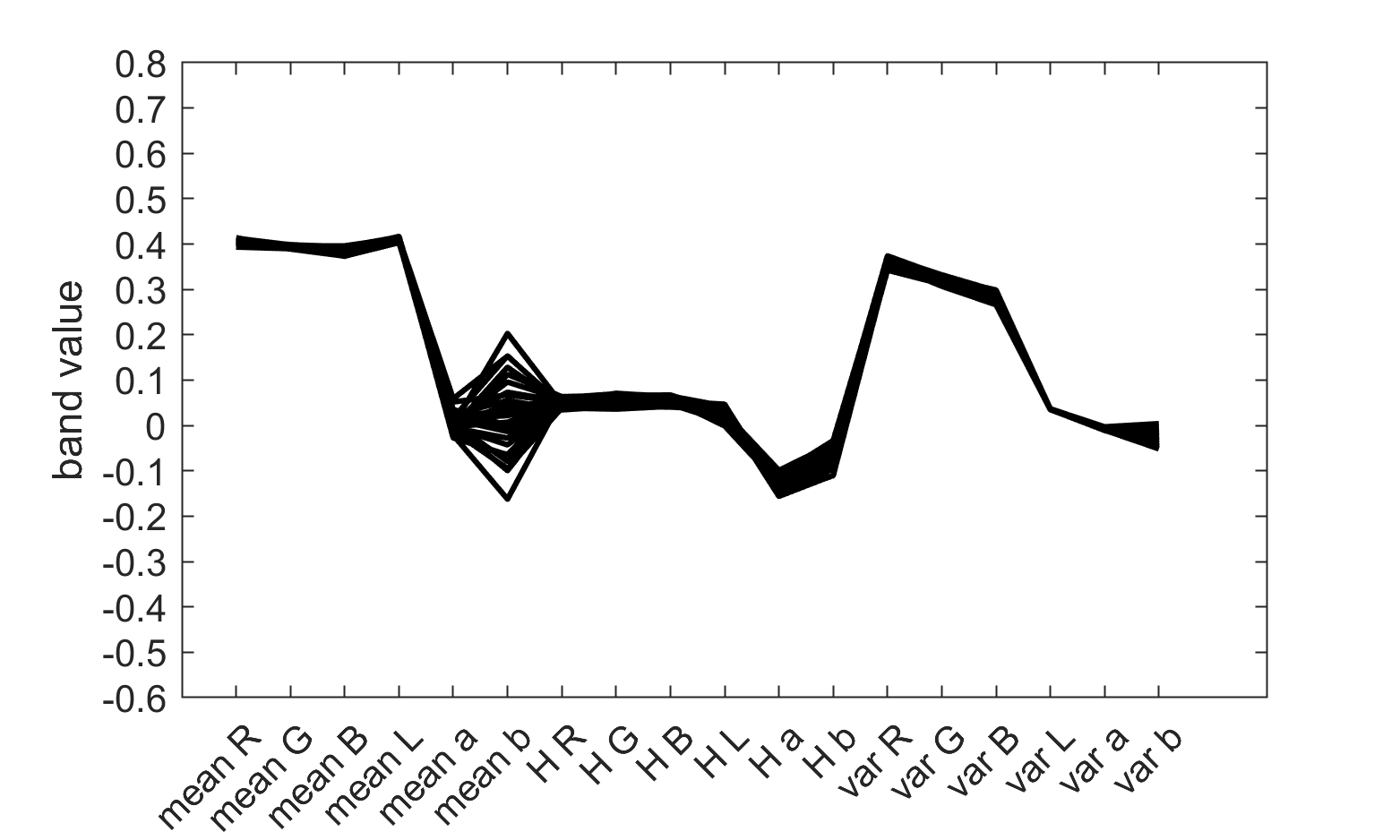}

   \caption{Root signatures. Root signatures generated by MI-ACE method with training data labeling one image as a bag.  The [mean-R, mean-G, mean-B, mean-L, mean-a, mean-b, H-R, H-G, H-B, H-L, H-a, H-b, var-R, var-G, var-B, var-L, var-a, var-b] features correspond to the 18-feature set extracted from each superpixel.  They  corresponding to the mean, entropy, and variance of samples in RGB and Lab band of each image, respectively}
\label{fig:6}
\end{center}
\end{figure}

\noindent\textbf {Evaluating training features via root signature}: 
An advantage of the MI-ACE algorithm (in addition the lack of parameters that need to be determined) was that a discriminative root signature was estimated.  This signature helped to identify the unique root attributes that distinguish root pixels from soil. Fig. \ref{fig:6} plots the estimated discriminative root target signatures. The signature bands of mean-R, mean-G, mean-B, mean-L, var-R, var-G, and var-B had positive values with small variances, indicating that these features were generally larger in value for roots than for soil and background. The signature bands of H-a, and H-b had negative values, indicating that roots generally had smaller values in these features than background. The larger the absolute value was, the more informative that feature was in distinguishing roots from soil. The signature bands of mean-a, H-R, H-G, H-B, H-L, var-L, var-a, and var-b were close to zeros which meant the difference between roots and soil at those feature bands were small. However, in some feature bands, the root signature values were widely spread. At a widely spread band, the signature value depended on the initial value and the differences between training results were large, for example the value of signature at mean-b in Fig. \ref{fig:6}.

\noindent\textbf {Evaluating sensitivity of MIL methods to features}: 
The features in these signatures in Fig. \ref{fig:6} can be divided into three groups by their corresponding signature value. The first group was the signature features with large average absolute signature value and small variance, the second group was the features that had small average absolute signature value with small variance, and the third group was features whose signature value had large variance. The comparison experiments with different selected features were done on the same training samples to examine the effects of features with different type of signature value. The experiments using all 18 features were compared with experiments using 9 and 17 features. The features which had large average absolute signature values with small variances were used in experiments with 9 features. Those feature were mean-R, mean-G, mean-B, mean-L, var-R, var-B, var-B, H-a, and H-b. In the experiment with 17 features, the mean-b feature with very large variance was removed from the feature set and all other features were kept.

\begin{table}[h]
\begin{center}
\caption{Mean and variance of TPR of MI-ACE at FPR ranging from 0.01 to 0.06. The number of features of training dataset were varied to be 9,17, or 18}
\label{table:MIACEVarfea}
\begin{tabular}{|l|*{7}{c|}}
\hline
FPR & \multicolumn{2}{|c|}{MI-ACE-9} & \multicolumn{2}{|c|}{MI-ACE-17} & \multicolumn{2}{|c|}{MI-ACE-18} \\
\hline
    & mean &var & mean &var & mean &var \\
\hline\hline
0.01 & \textbf{0.68} & \textbf{0.11e-4} & 0.66 & 0.54e-3 & 0.66 & 0.38e-3  \\
0.02 & \textbf{0.74} & \textbf{0.03e-4} & 0.72 & 0.34e-3 & 0.72 & 0.33e-3  \\
0.03 & \textbf{0.77} & \textbf{0.04e-4} & 0.76 & 0.26e-3 & 0.75 & 0.32e-3  \\
0.04 & \textbf{0.79} & \textbf{0.03e-4} & 0.78 & 0.21e-3 & 0.77 & 0.29e-3  \\
0.05 & \textbf{0.80} & \textbf{0.04e-4} & 0.80 & 0.19e-3 & 0.79 & 0.25e-3  \\
0.06 & \textbf{0.81} & \textbf{0.04e-4} & 0.81 & 0.16e-3 & 0.80 & 0.23e-3  \\
\hline
\end{tabular}
\end{center}
\end{table}

\begin{table}[h]
\begin{center}
\caption{mean and variance of TPR of MIForests at FPR ranging from 0.01 to 0.06. The number of features of training dataset varied to be 9,17, or 18}
\label{table:MIForestVarfea}
\begin{tabular}{|l|*{7}{c|}}
\hline
FPR & \multicolumn{2}{|c|}{MIForests-9} & \multicolumn{2}{|c|}{MIForests-17} & \multicolumn{2}{|c|}{MIForests-18} \\
\hline
    & mean &var & mean &var & mean &var \\
\hline\hline
0.01 & 0.36 & 1.3e-3 & \textbf{0.51} & \textbf{2.0e-3} & 0.49 & 3.7e-3  \\
0.02 & 0.40 & 1.5e-3 & \textbf{0.56} & \textbf{2.2e-3} & 0.54 & 4.1e-3  \\
0.03 & 0.43 & 1.5e-3 & \textbf{0.60} & \textbf{2.3e-3} & 0.58 & 4.1e-3  \\
0.04 & 0.46 & 1.6e-3 & \textbf{0.62} & \textbf{2.3e-3} & 0.61 & 4.0e-3  \\
0.05 & 0.48 & 1.7e-3 & \textbf{0.65} & \textbf{2.3e-3} & 0.63 & 3.8e-3  \\
0.06 & 0.50 & 1.7e-3 & \textbf{0.67} & \textbf{2.2e-3} & 0.66 & 3.6e-3  \\
\hline
\end{tabular}
\end{center}
\end{table}

\begin{table}[h]
\begin{center}
\caption{Mean and variance of TPR of miSVM at FPR ranging from 0.01 to 0.06. The number of features of training dataset varied to be 9,17, or 18}
\label{table:miSVMVarfea}
\begin{tabular}{|l|*{7}{c|}}
\hline
FPR & \multicolumn{2}{|c|}{miSVM-9} & \multicolumn{2}{|c|}{miSVM-17} & \multicolumn{2}{|c|}{miSVM-18} \\
\hline
    & mean &var & mean &var & mean &var \\
\hline\hline
0.01 & 0.67 & 0.86e-4 & 0.69 & 1.84e-04 & \textbf{0.70} & \textbf{2.10e-04}  \\
0.02 & \textbf{0.75} & \textbf{0.17e-4} & 0.75 & 1.19e-04 & 0.75 & 1.17e-04  \\
0.03 & \textbf{0.78} & \textbf{0.10e-4} & 0.78 & 0.88e-04 & 0.78 & 1.03e-04  \\
0.04 & \textbf{0.80} & \textbf{0.06e-4} & 0.80 & 0.70e-04 & 0.80 & 0.82e-04  \\
0.05 & \textbf{0.81} & \textbf{0.05e-4} & 0.81 & 0.55e-04 & 0.81 & 0.74e-04  \\
0.06 & \textbf{0.82} & \textbf{0.03e-4} & 0.82 & 0.48e-04 & 0.82 & 0.64e-04  \\
\hline
\end{tabular}
\end{center}
\end{table}

For MI-ACE and miSVM methods, the features with large average absolute value were important for detection results as seen in Table \ref{table:MIACEVarfea} and \ref{table:miSVMVarfea}. It was also noted that the variances of detection results of MI-ACE and miSVM were significant improved by removing features which had large signature variance and small average absolute signature value. As seen in Table \ref{table:MIACEVarfea} and \ref{table:miSVMVarfea}, the variance of TPR with 9 features was one order of magnitude smaller than the variance of TPR with 18 features. Therefore, large average absolute signature values dominated the detection results and the larger the value was, the more consistent the results were for the MI-ACE and miSVM methods.

In contrast, we found that the features with small variance were important for the MIForests method. The sensitivity analysis showed an improvement in the variance when the features with the largest variance were removed as shown in Table \ref{table:MIForestVarfea}. Although the variance of detection using only the features with the small variance and large mean further improved the variance, it also reduced the TPR mean. Thus, the use of all features with small variance regardless of their mean values was important for the MIForests method.

\begin{figure*}[t] 
\begin{center}
\begin{subfigure}{0.12\textwidth}
   \includegraphics[width=1\linewidth]{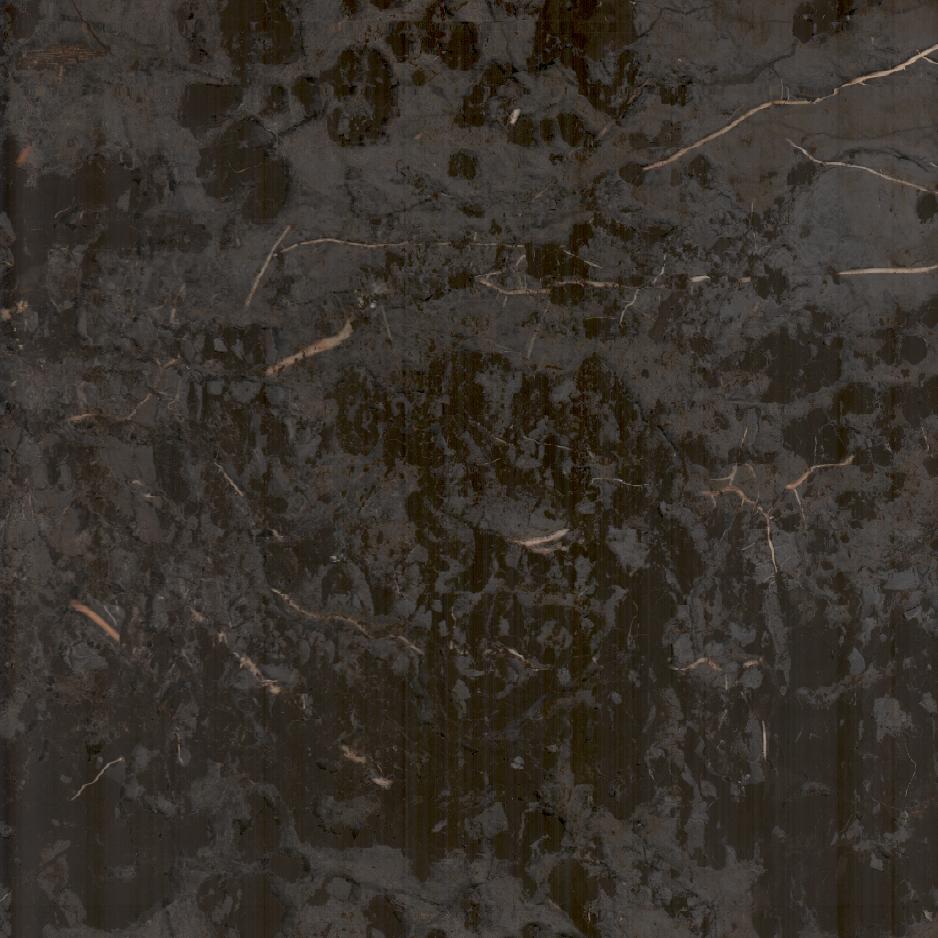}
\end{subfigure}
\begin{subfigure}{0.12\textwidth}
   \includegraphics[width=1\linewidth]{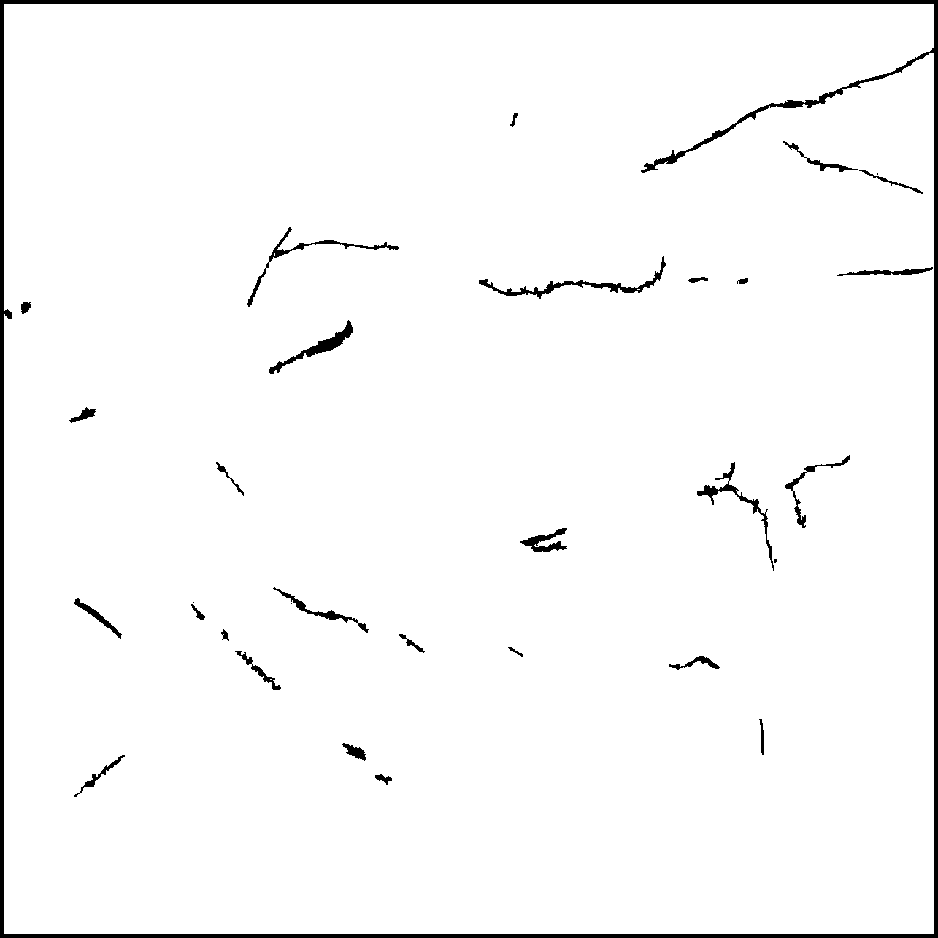}
\end{subfigure}
\begin{subfigure}{0.12\textwidth}
   \includegraphics[width=1\linewidth]{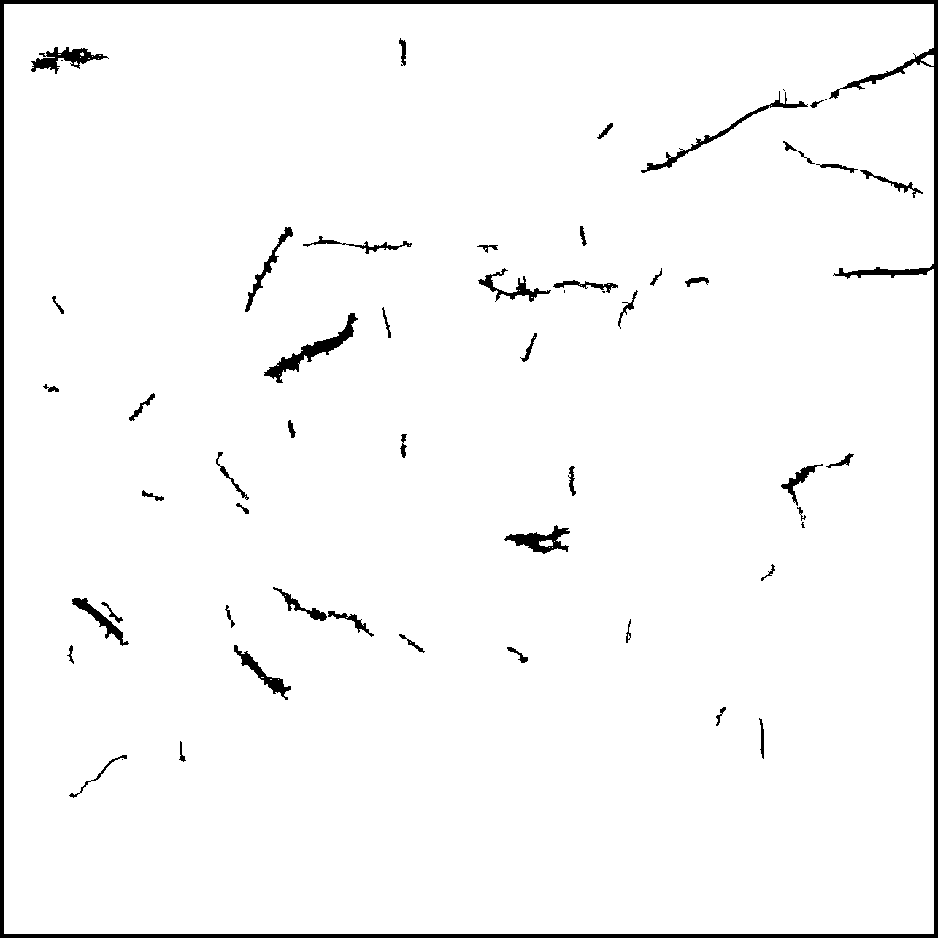}
\end{subfigure}
\begin{subfigure}{0.12\textwidth}
   \includegraphics[width=1\linewidth]{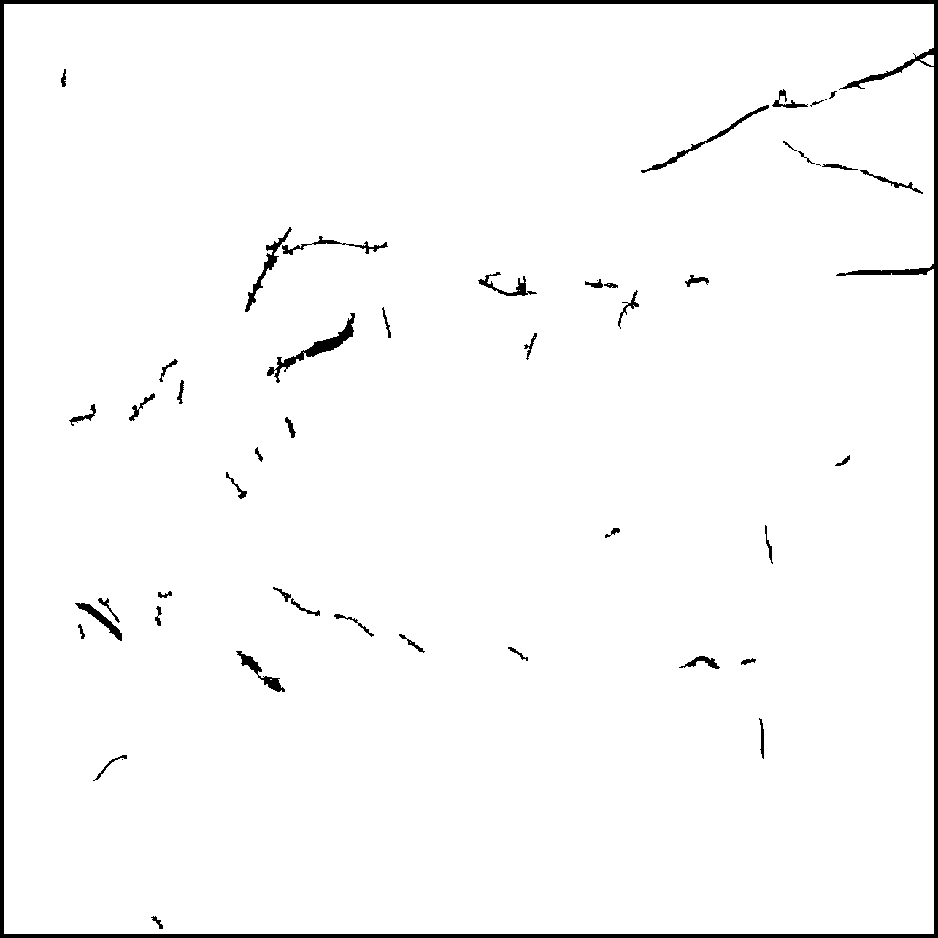}
\end{subfigure}
\begin{subfigure}{0.12\textwidth}
   \includegraphics[width=1\linewidth]{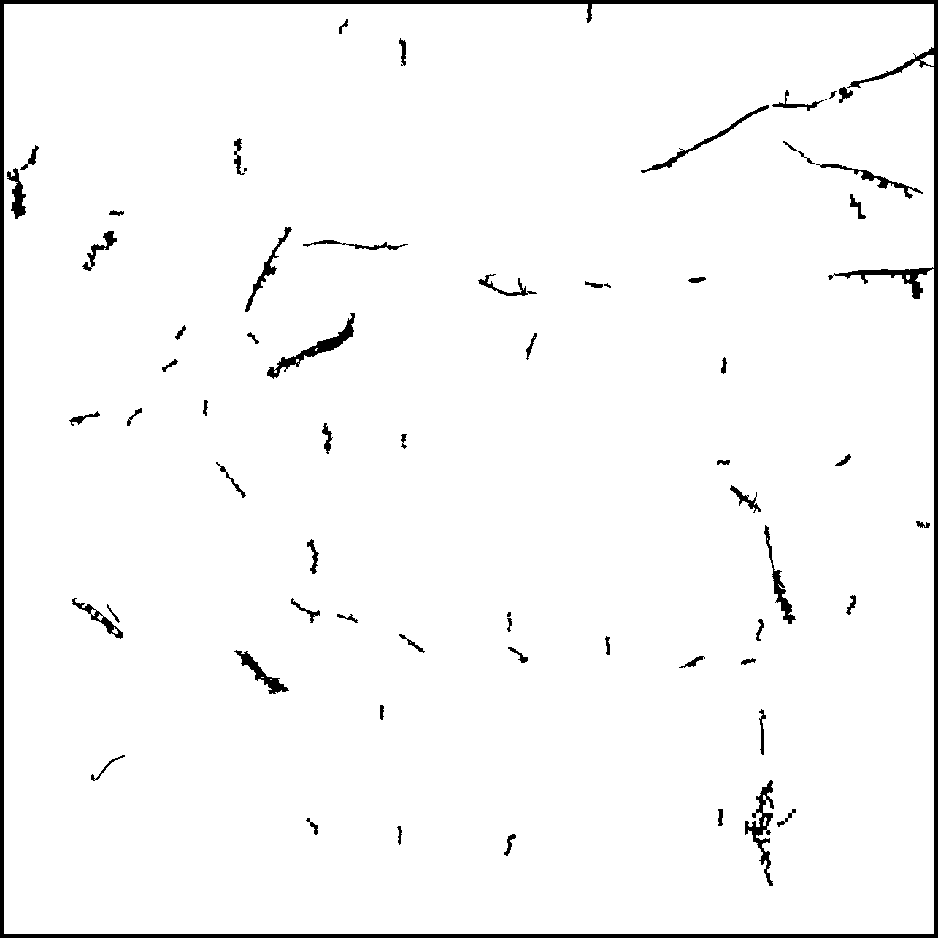}
\end{subfigure}
\begin{subfigure}{0.12\textwidth}
   \includegraphics[width=1\linewidth]{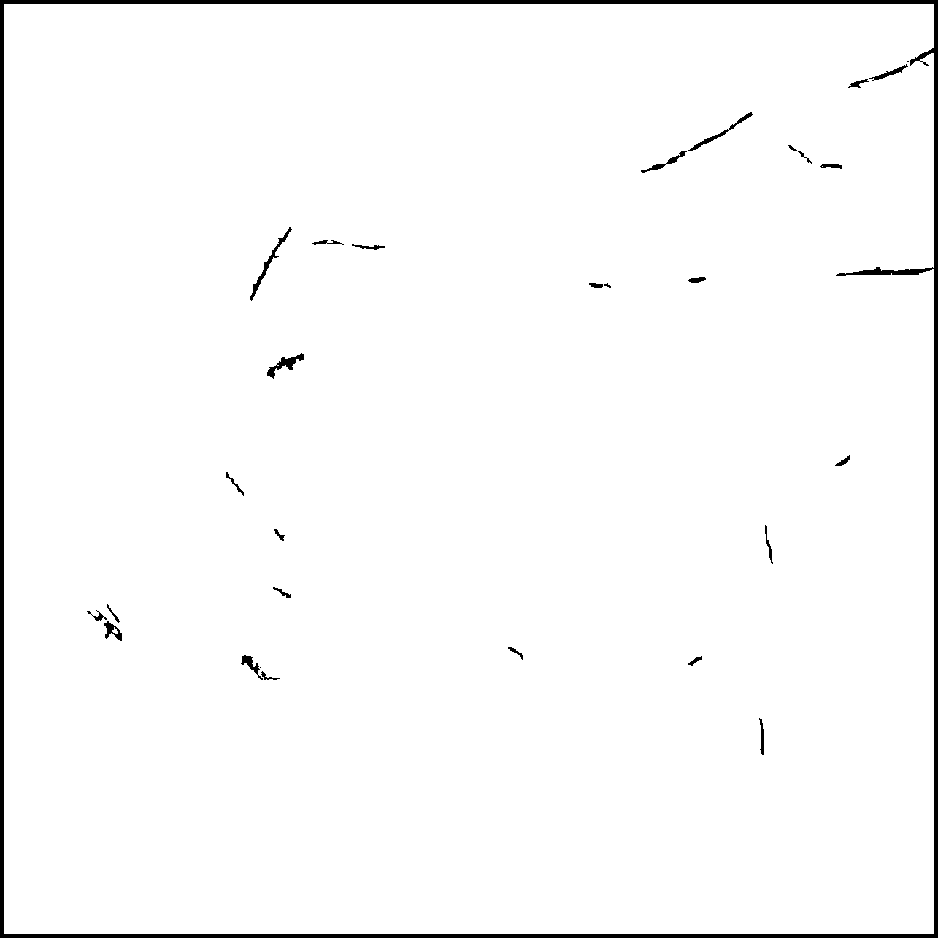}
\end{subfigure}
\begin{subfigure}{0.12\textwidth}
   \includegraphics[width=1\linewidth]{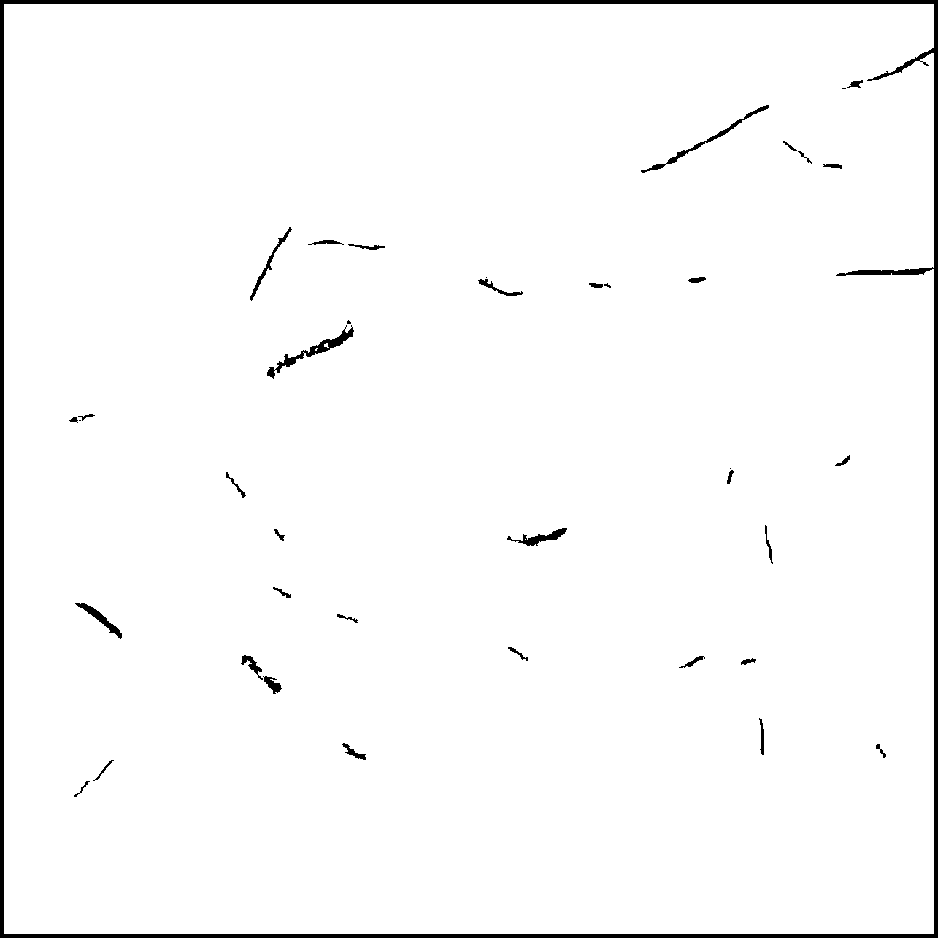}
\end{subfigure}
\\[\baselineskip]
\begin{subfigure}{0.12\textwidth}
   \includegraphics[width=1\linewidth]{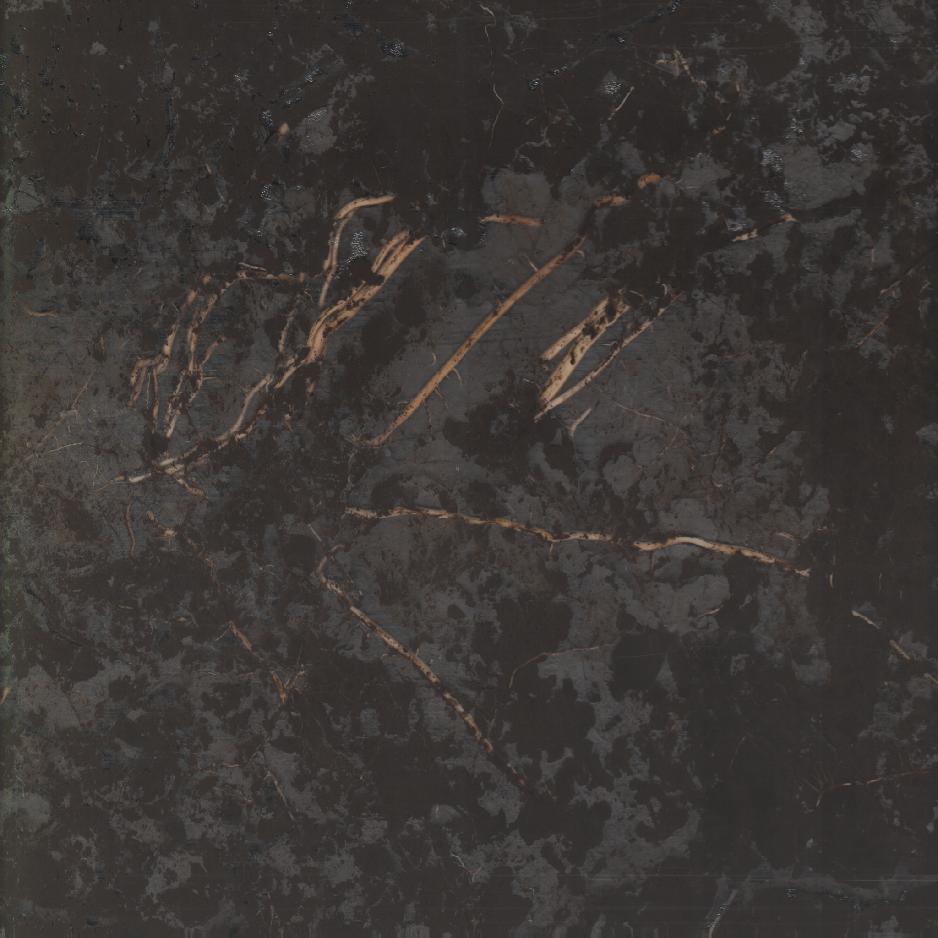}
\end{subfigure}
\begin{subfigure}{0.12\textwidth}
   \includegraphics[width=1\linewidth]{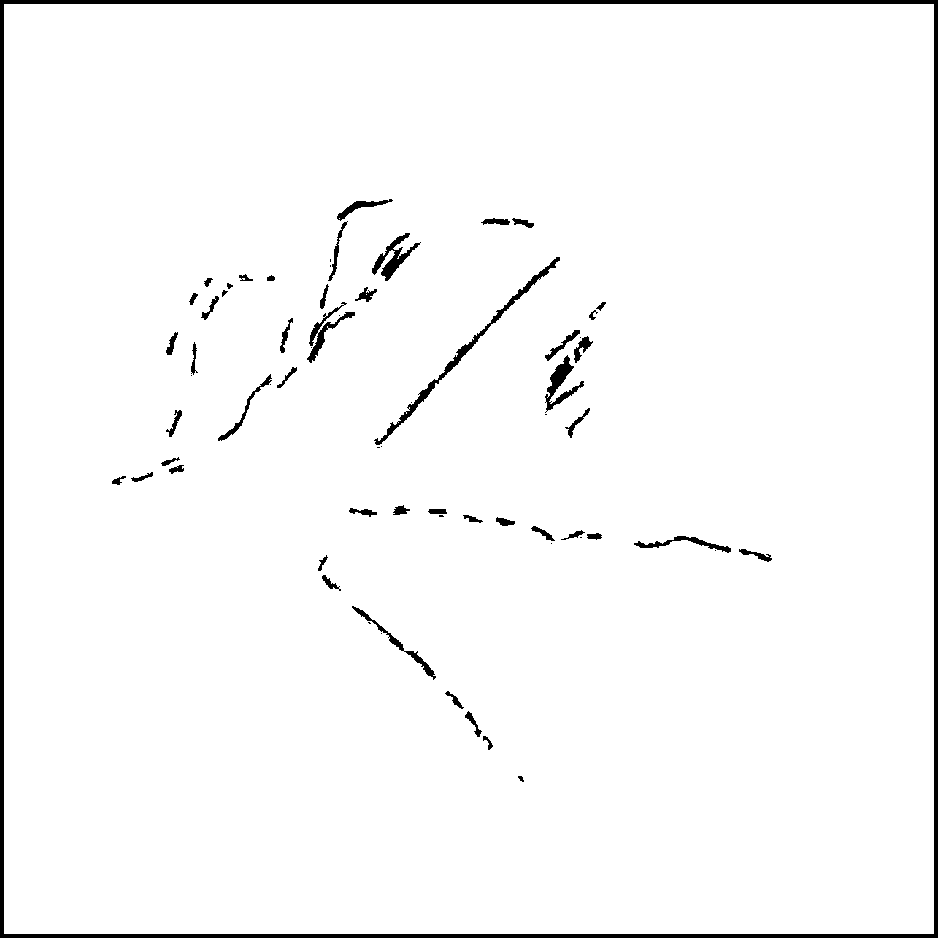}
\end{subfigure}
\begin{subfigure}{0.12\textwidth}
   \includegraphics[width=1\linewidth]{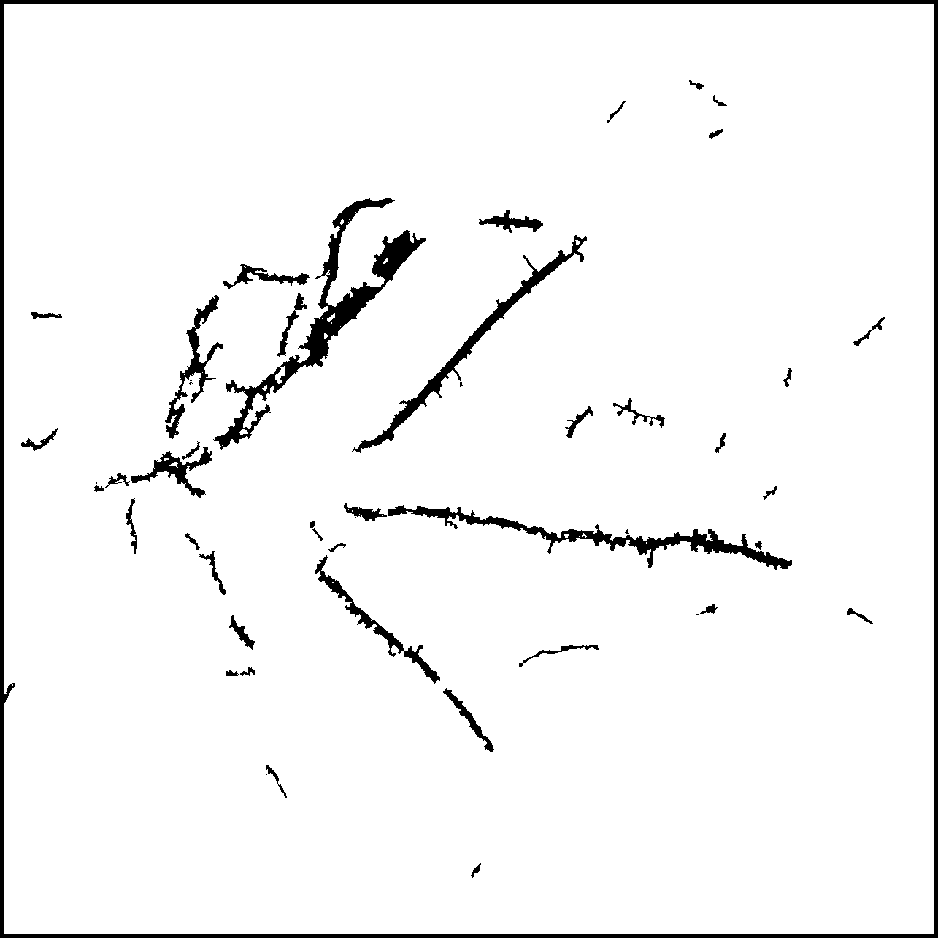}
\end{subfigure}
\begin{subfigure}{0.12\textwidth}
   \includegraphics[width=1\linewidth]{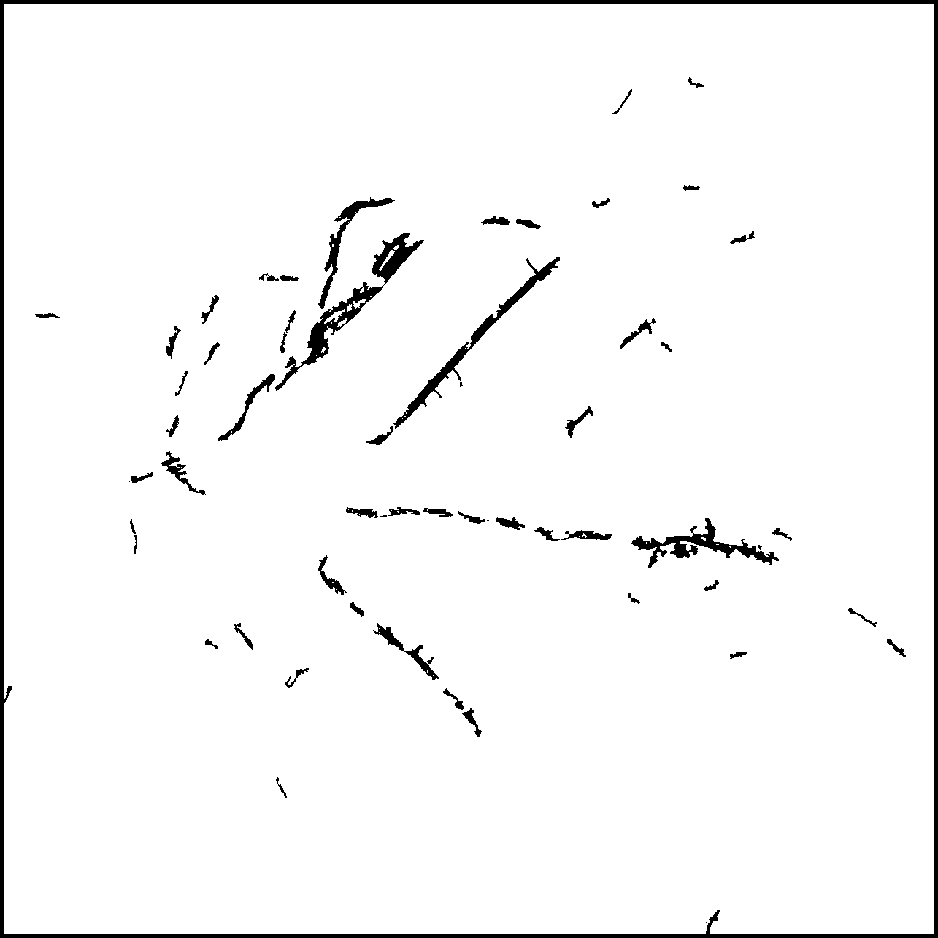}
\end{subfigure}
\begin{subfigure}{0.12\textwidth}
   \includegraphics[width=1\linewidth]{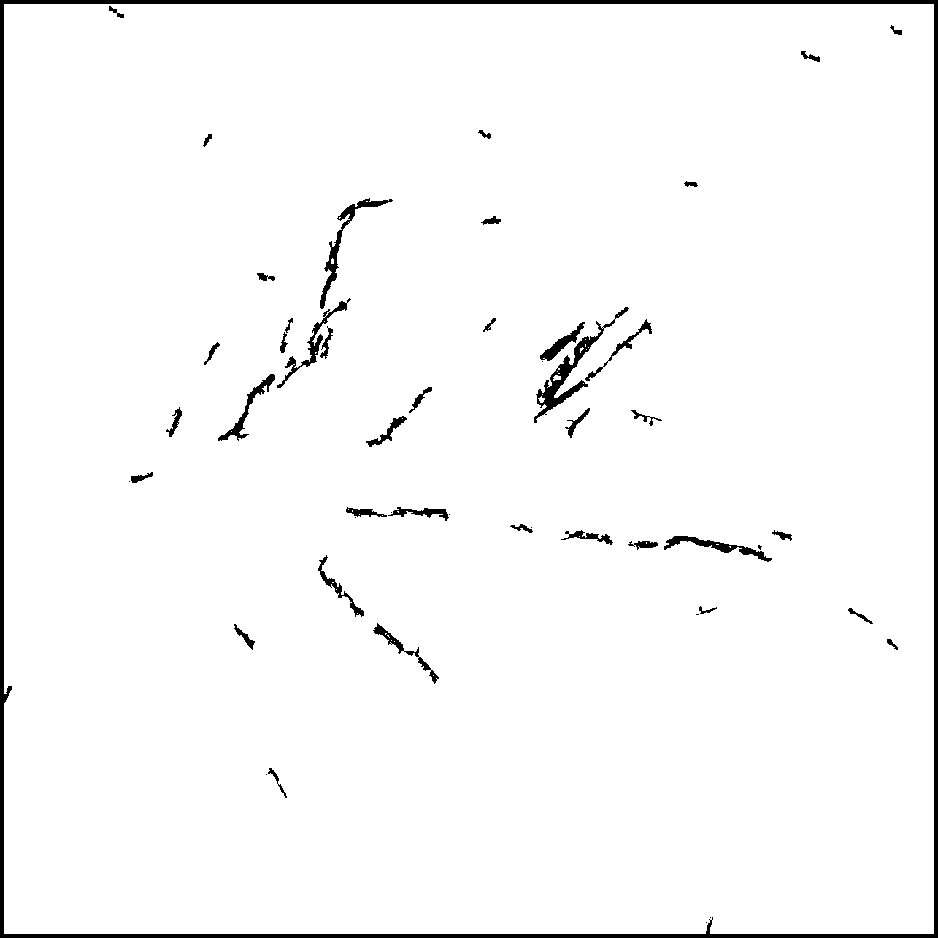}
\end{subfigure}
\begin{subfigure}{0.12\textwidth}
   \includegraphics[width=1\linewidth]{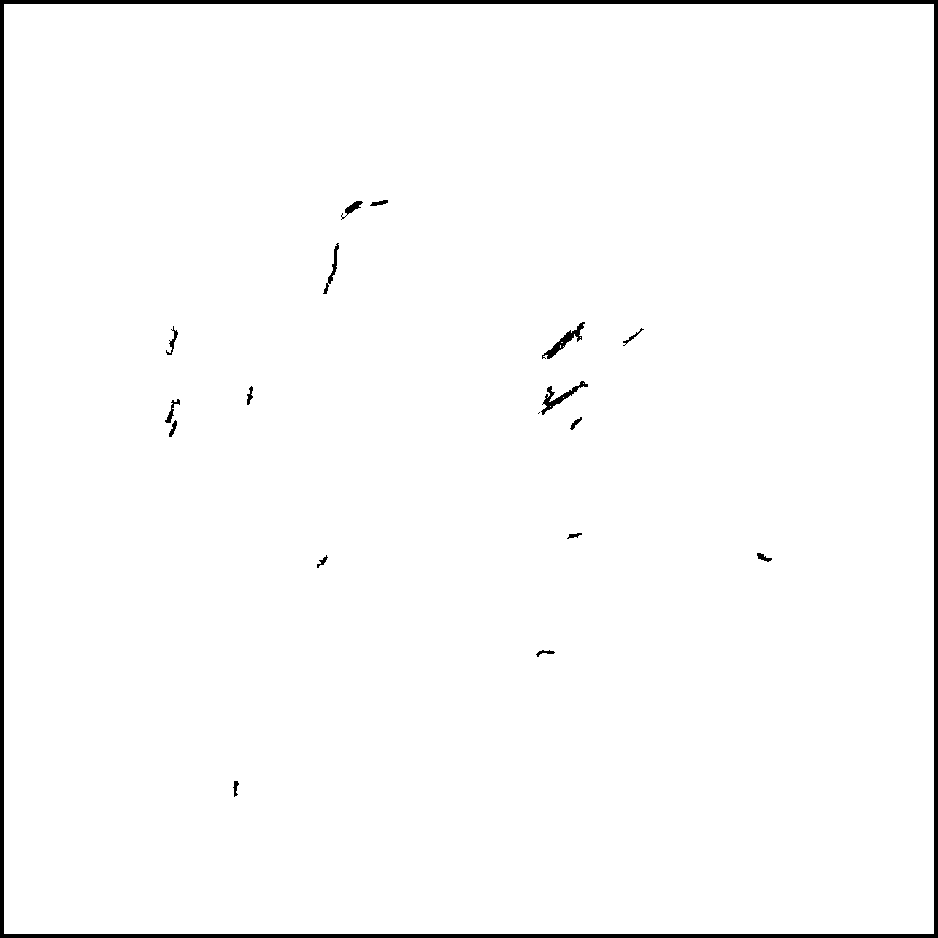}
\end{subfigure}
\begin{subfigure}{0.12\textwidth}
   \includegraphics[width=1\linewidth]{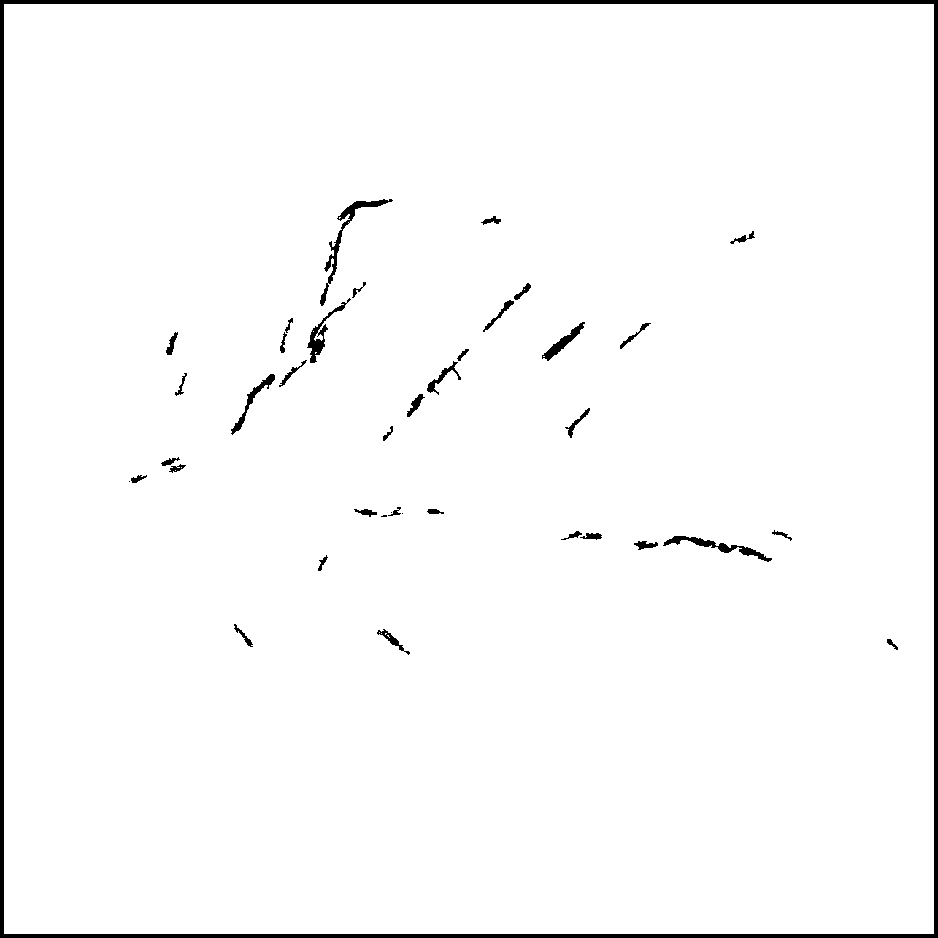}
\end{subfigure}
\\[\baselineskip]
\begin{subfigure}{0.12\textwidth}
   \includegraphics[width=1\linewidth]{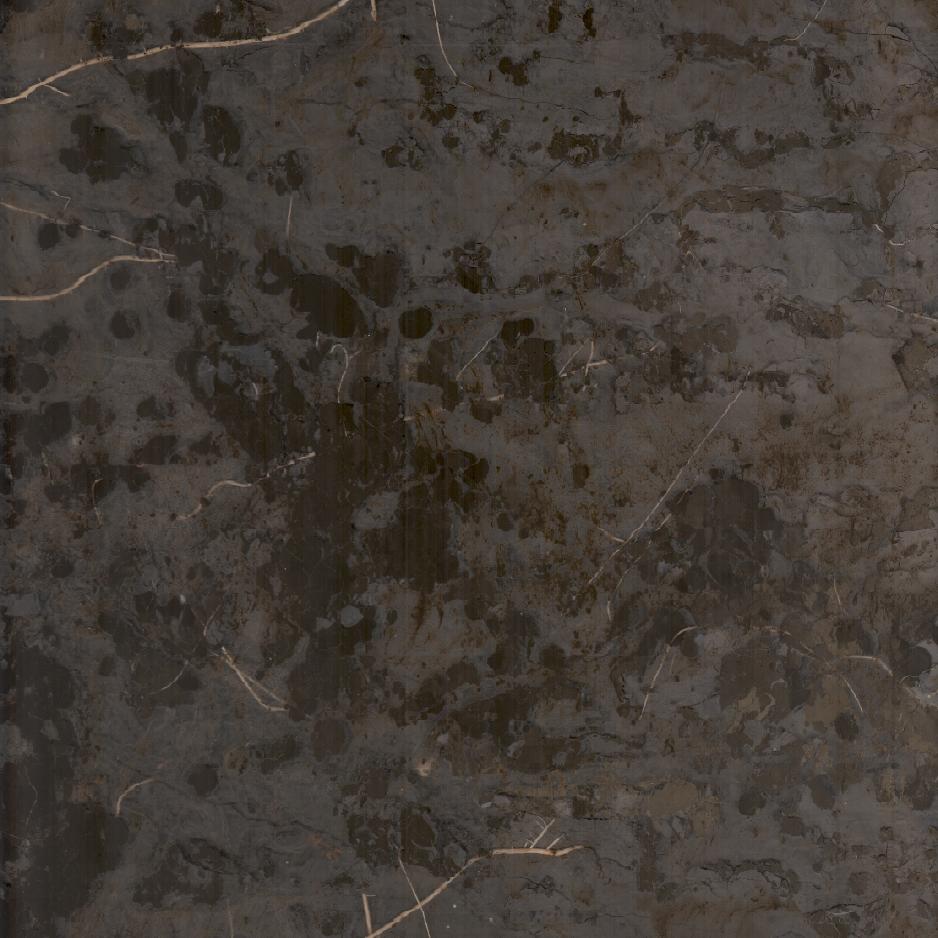}
\end{subfigure}
\begin{subfigure}{0.12\textwidth}
   \includegraphics[width=1\linewidth]{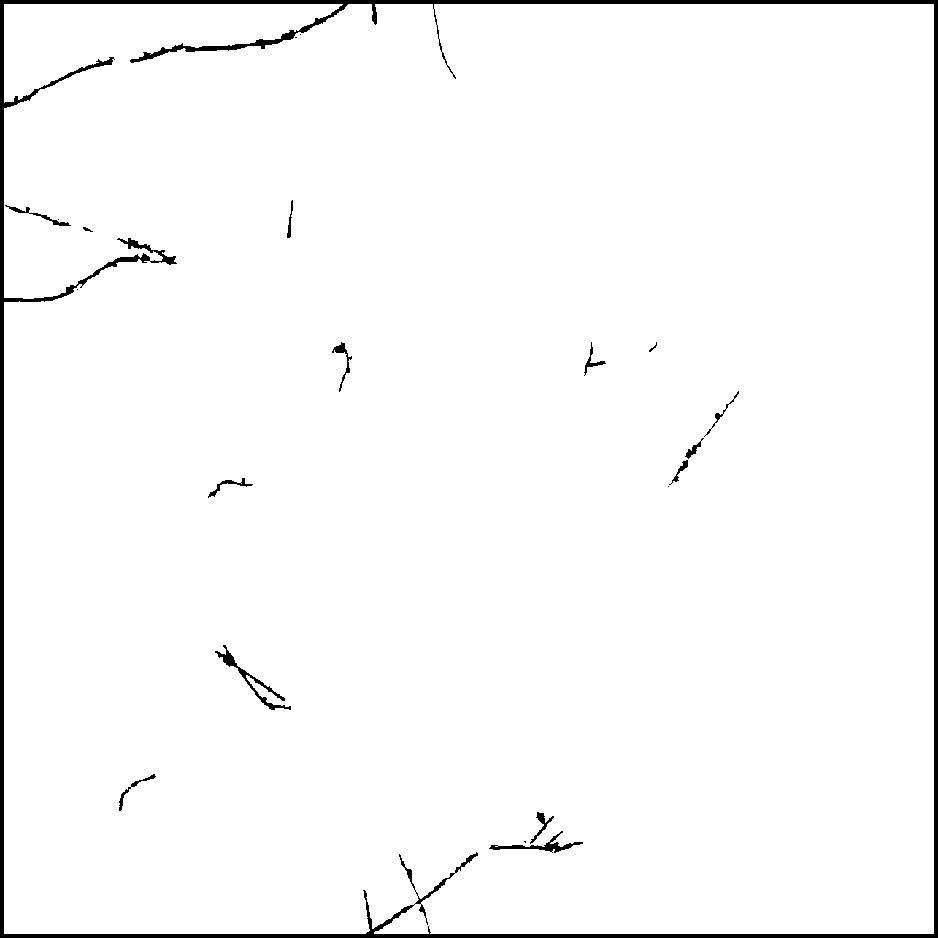}
\end{subfigure}
\begin{subfigure}{0.12\textwidth}
   \includegraphics[width=1\linewidth]{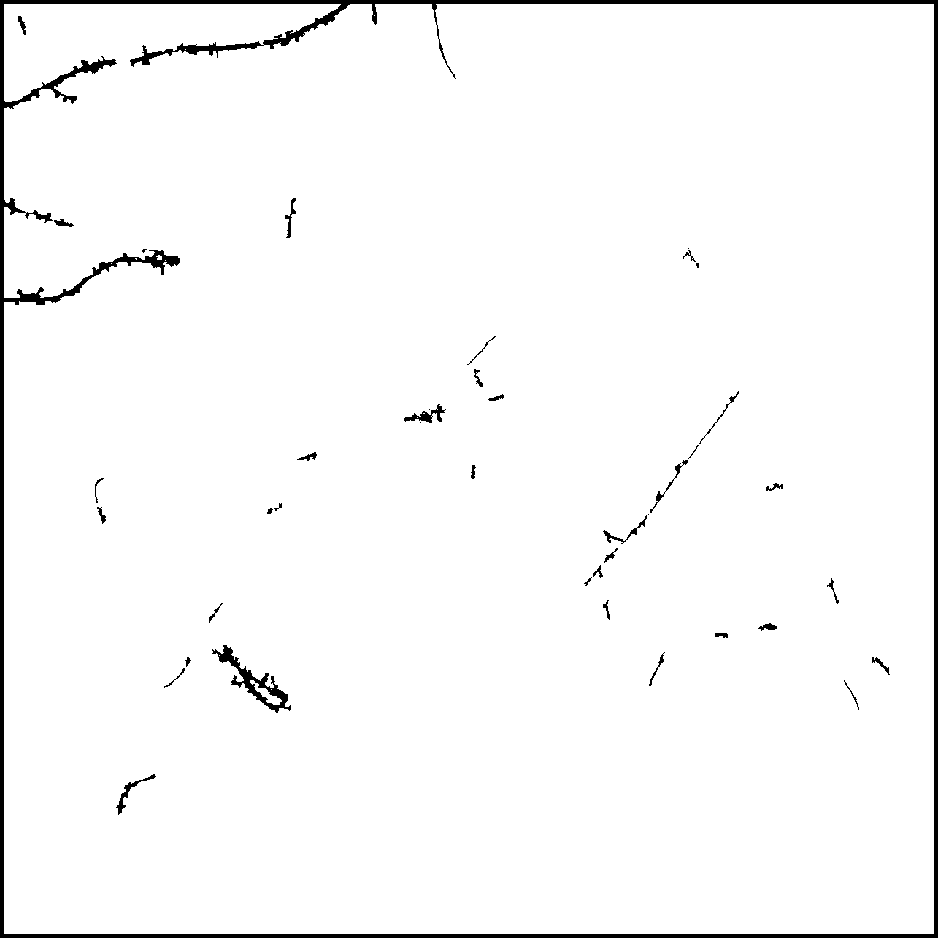}
\end{subfigure}
\begin{subfigure}{0.12\textwidth}
   \includegraphics[width=1\linewidth]{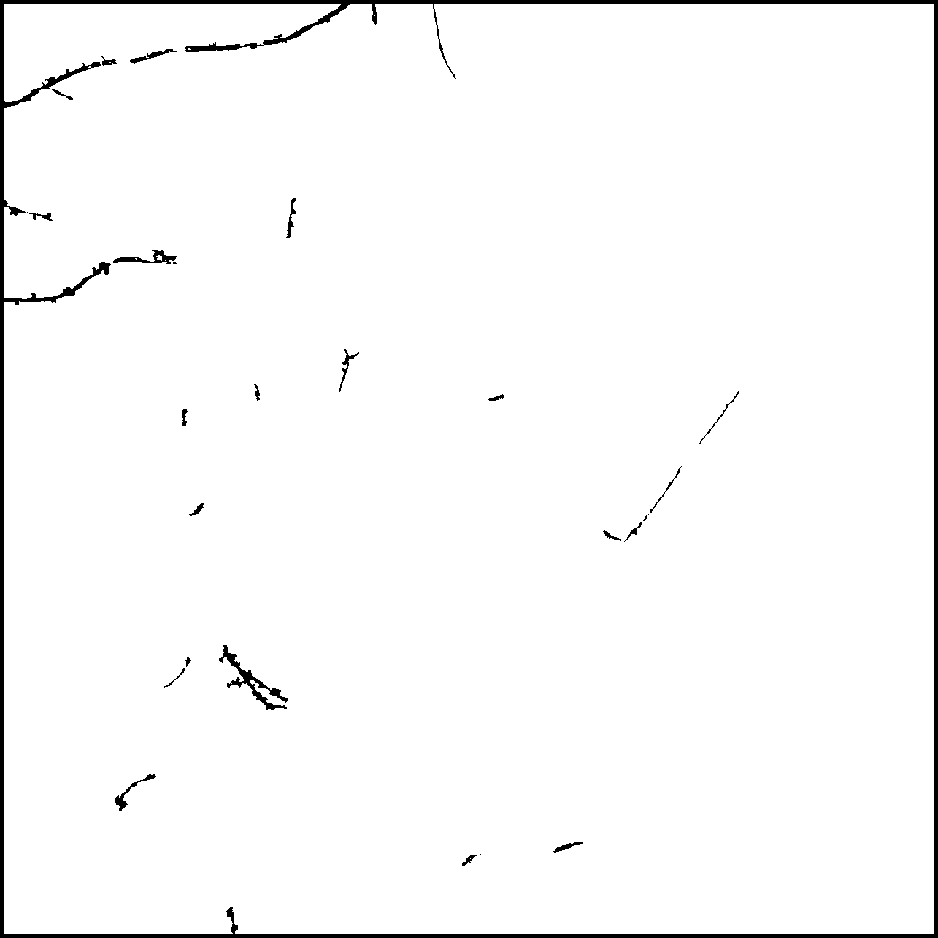}
\end{subfigure}
\begin{subfigure}{0.12\textwidth}
   \includegraphics[width=1\linewidth]{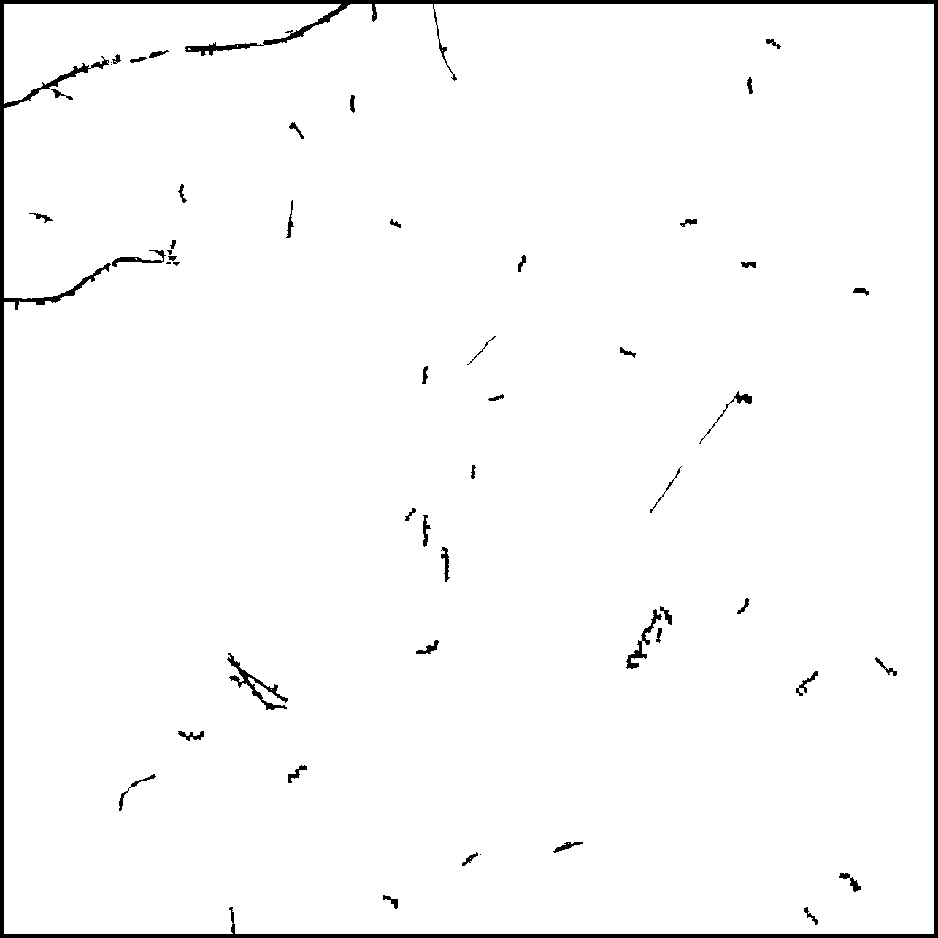}
\end{subfigure}
\begin{subfigure}{0.12\textwidth}
   \includegraphics[width=1\linewidth]{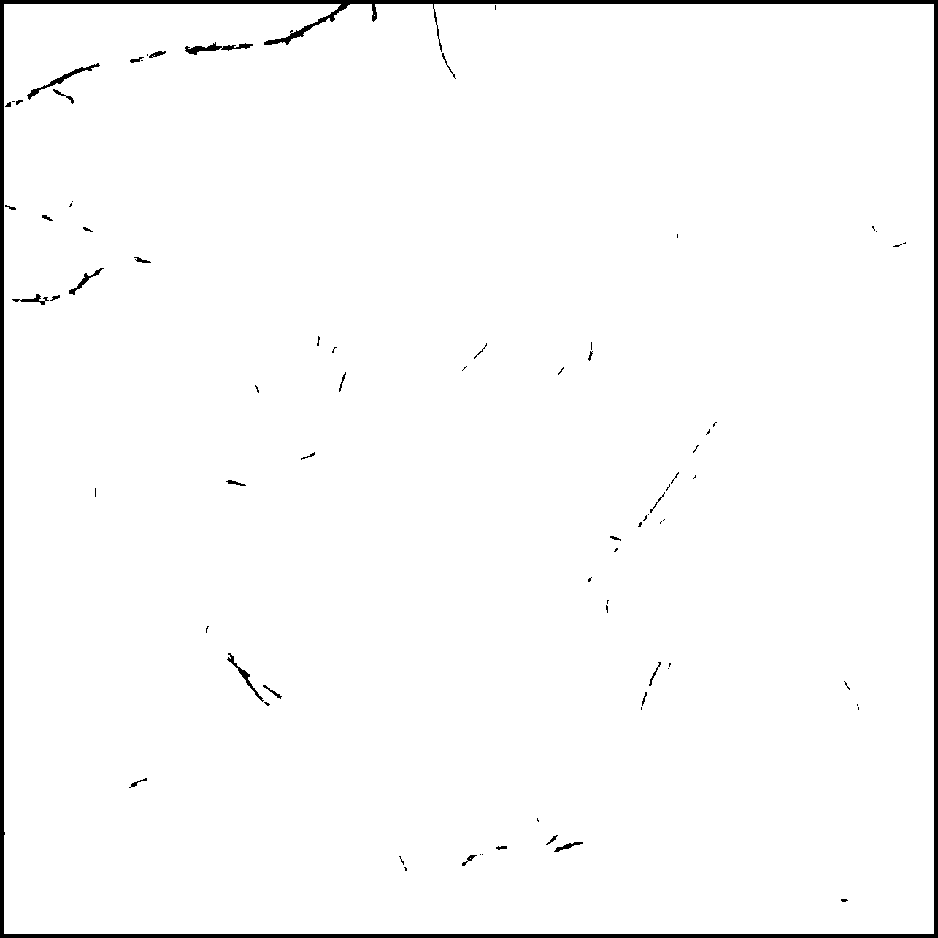}
\end{subfigure}
\begin{subfigure}{0.12\textwidth}
   \includegraphics[width=1\linewidth]{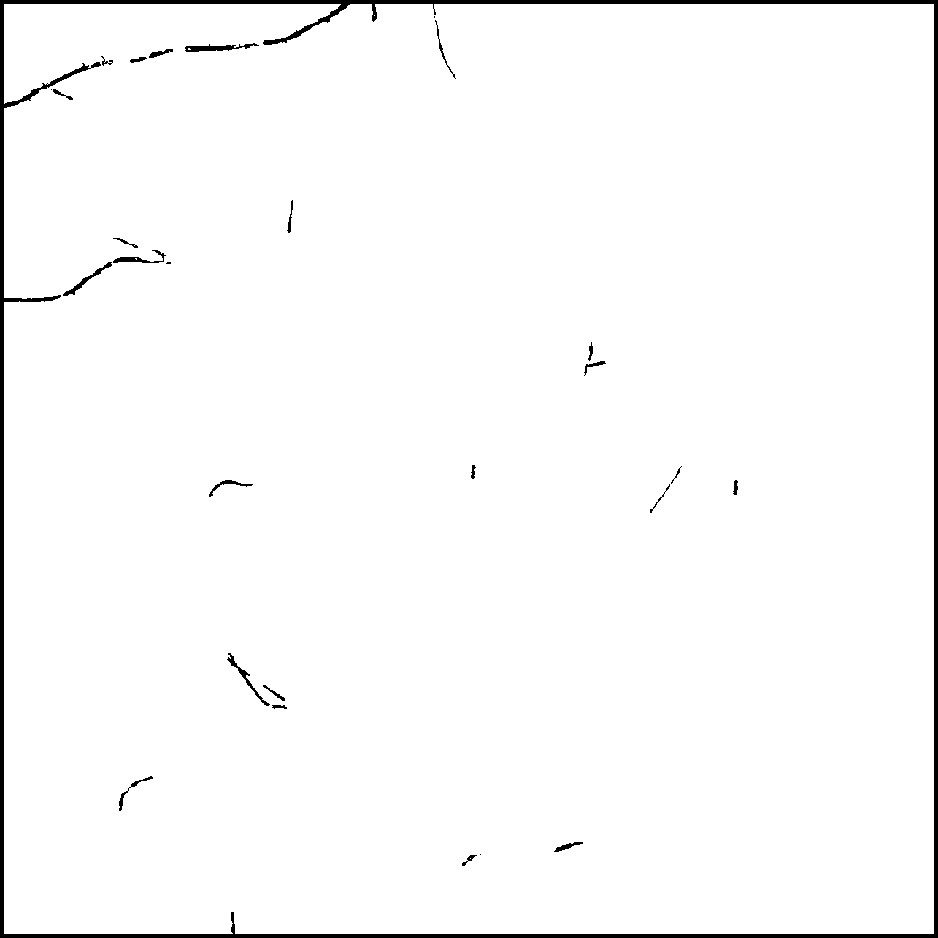}
\end{subfigure}
\\[\baselineskip]
\begin{subfigure}{0.12\textwidth}
   \includegraphics[width=1\linewidth]{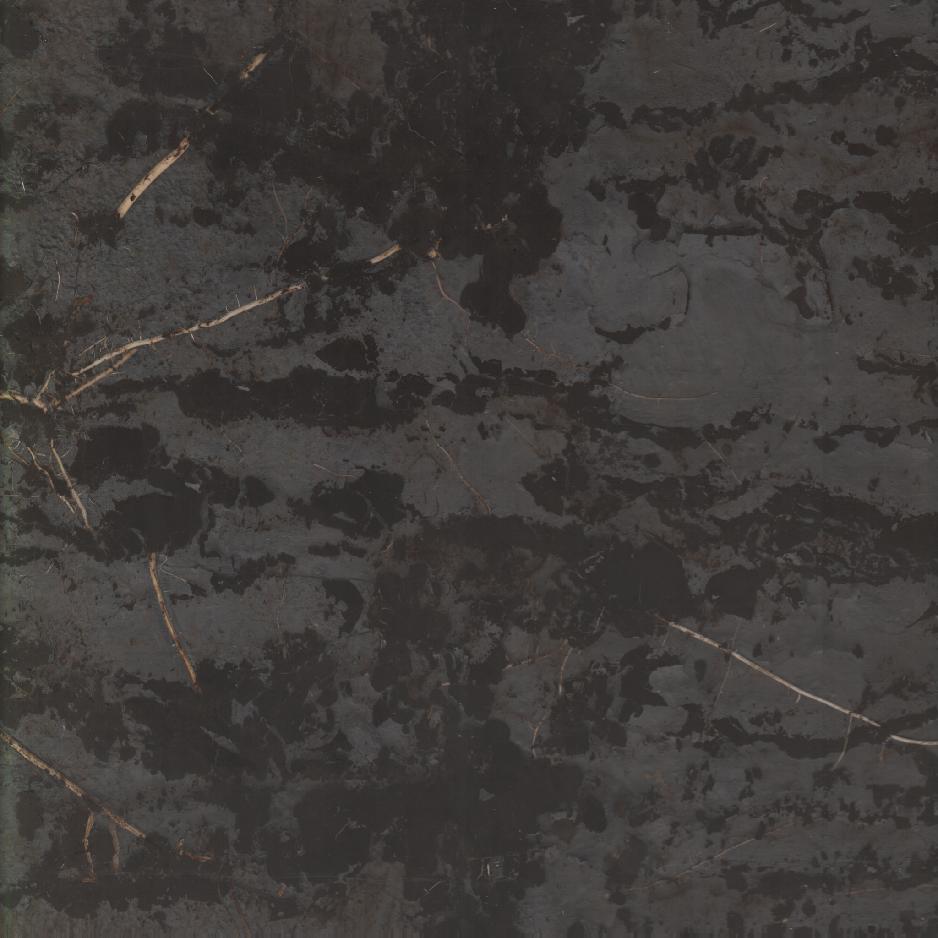}
\end{subfigure}
\begin{subfigure}{0.12\textwidth}
   \includegraphics[width=1\linewidth]{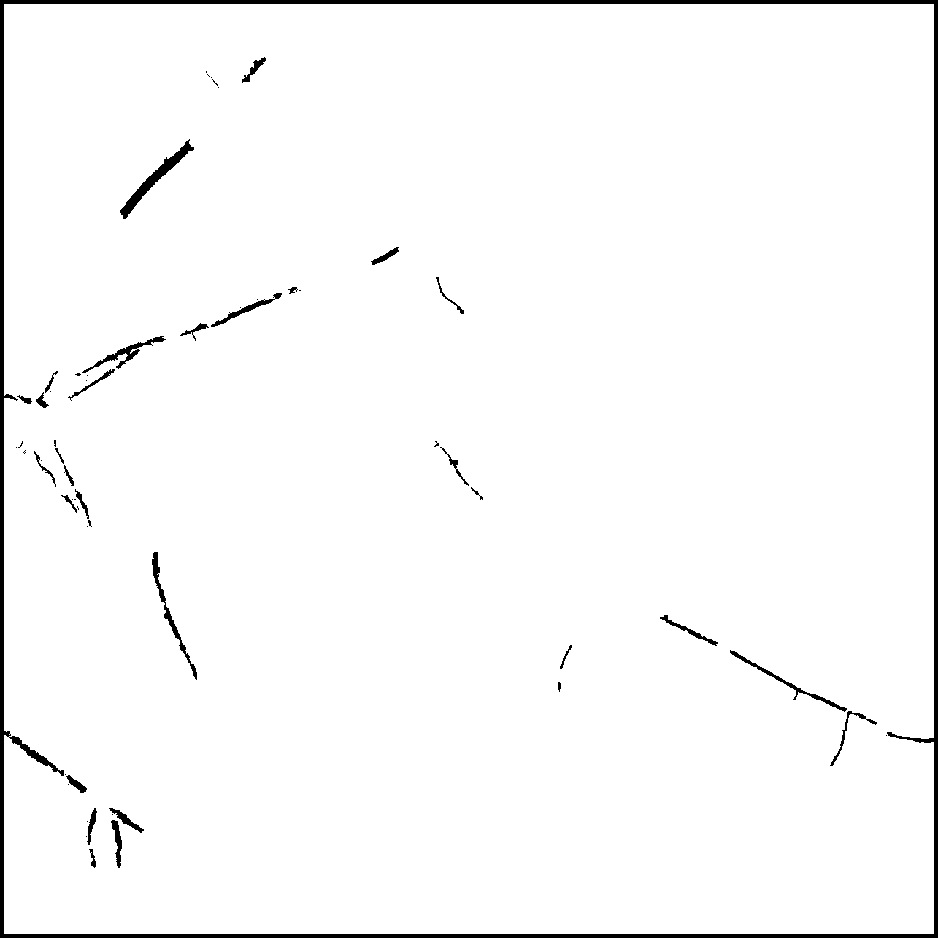}
\end{subfigure}
\begin{subfigure}{0.12\textwidth}
   \includegraphics[width=1\linewidth]{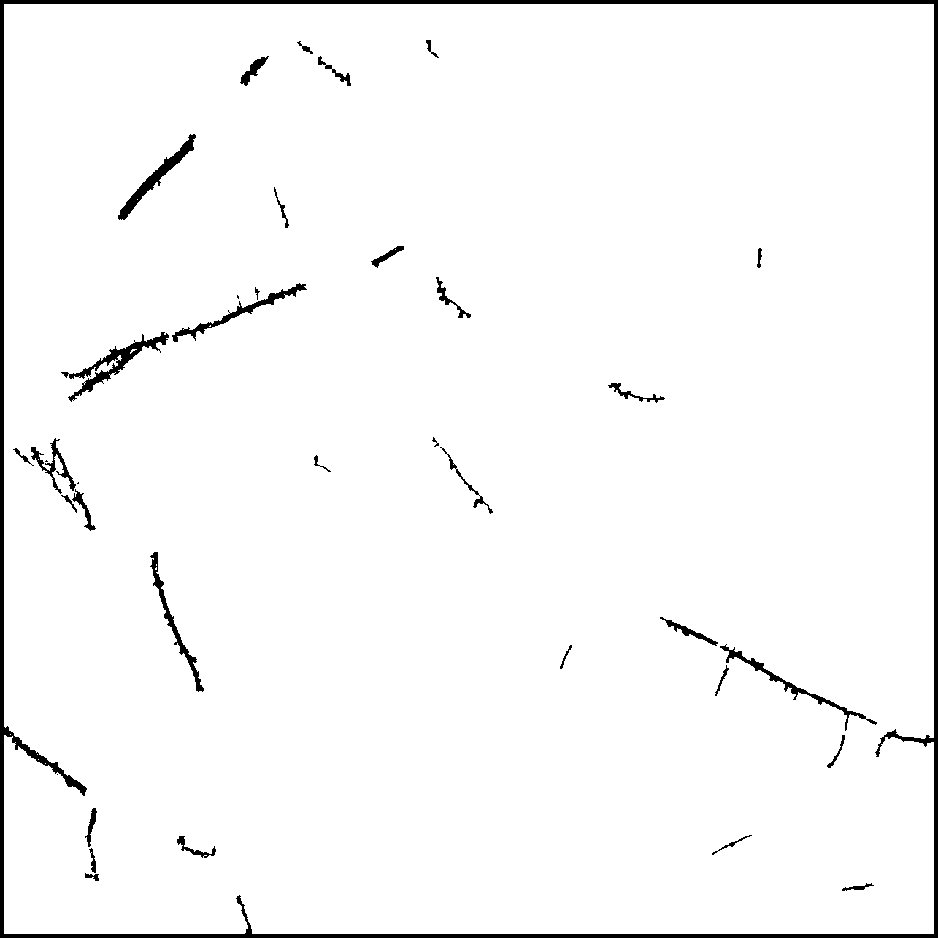}
\end{subfigure}
\begin{subfigure}{0.12\textwidth}
   \includegraphics[width=1\linewidth]{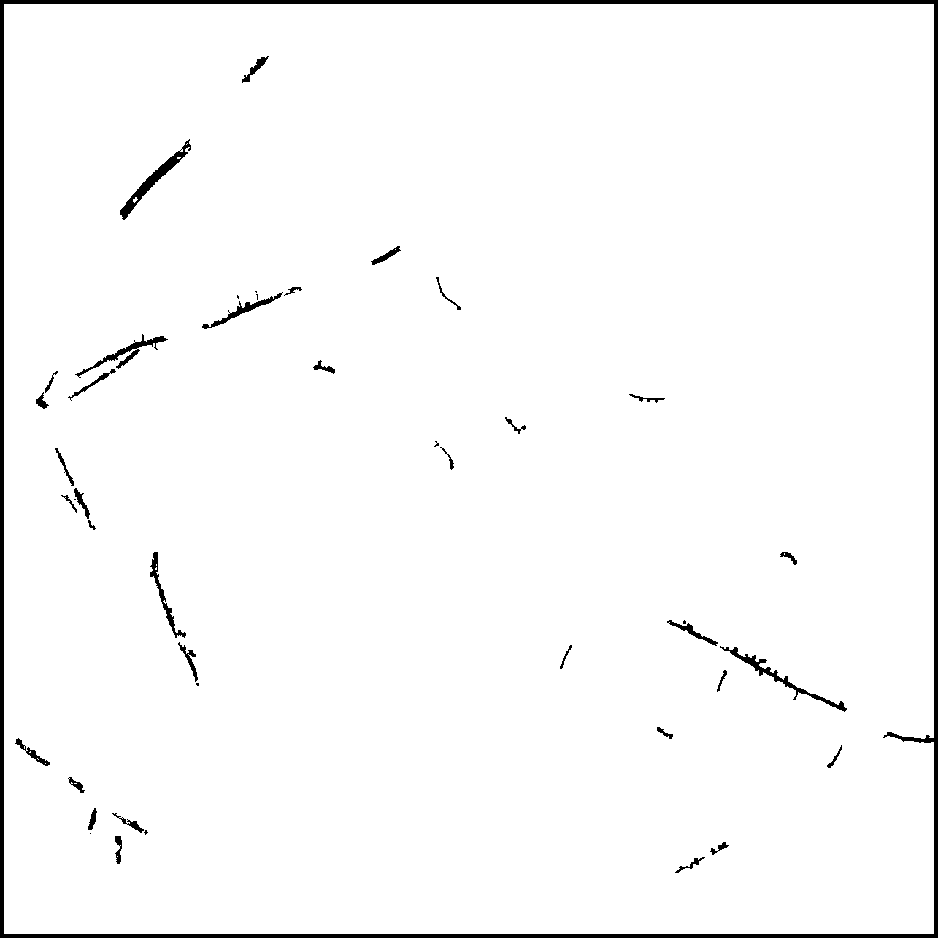}
\end{subfigure}
\begin{subfigure}{0.12\textwidth}
   \includegraphics[width=1\linewidth]{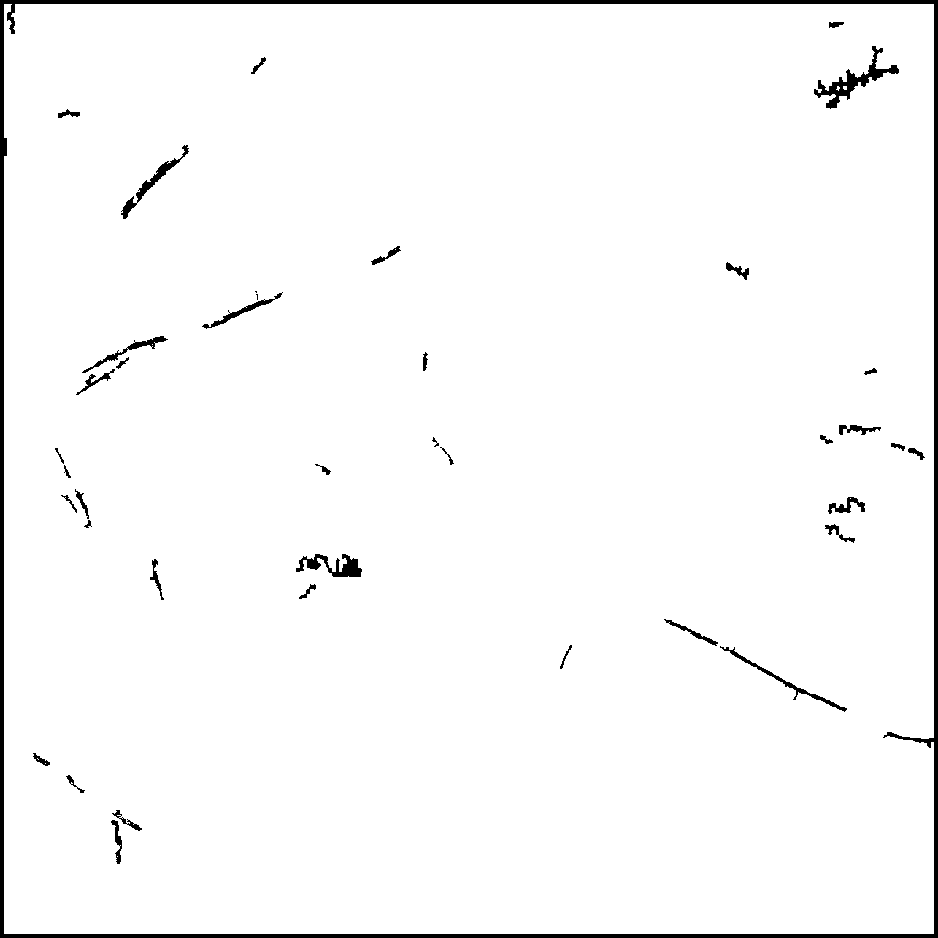}
\end{subfigure}
\begin{subfigure}{0.12\textwidth}
   \includegraphics[width=1\linewidth]{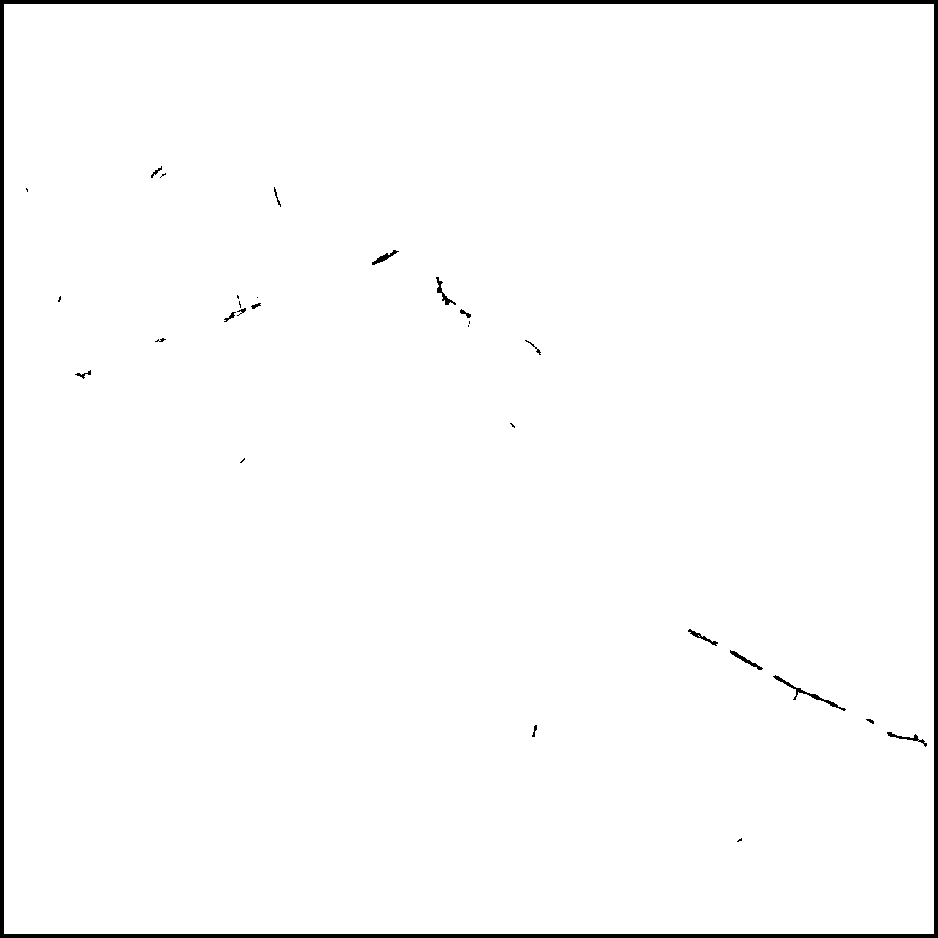}
\end{subfigure}
\begin{subfigure}{0.12\textwidth}
   \includegraphics[width=1\linewidth]{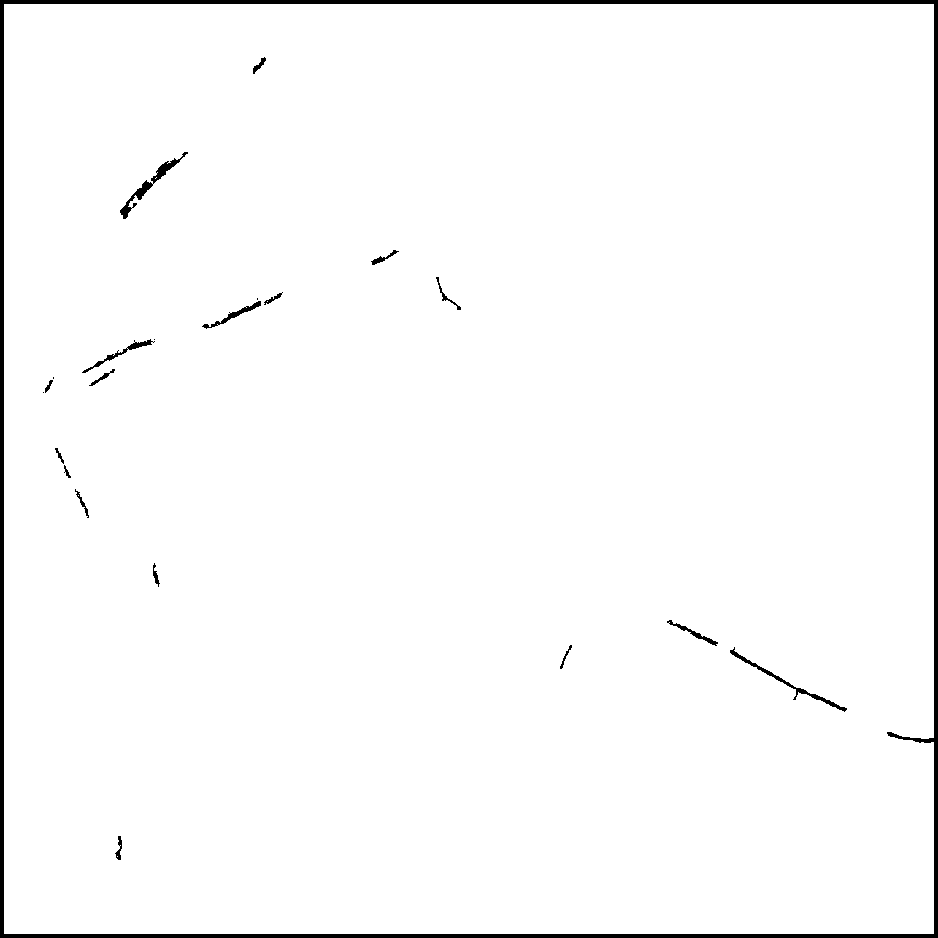}
\end{subfigure}
\\[\baselineskip]
\begin{subfigure}{0.12\textwidth}
   \includegraphics[width=1\linewidth]{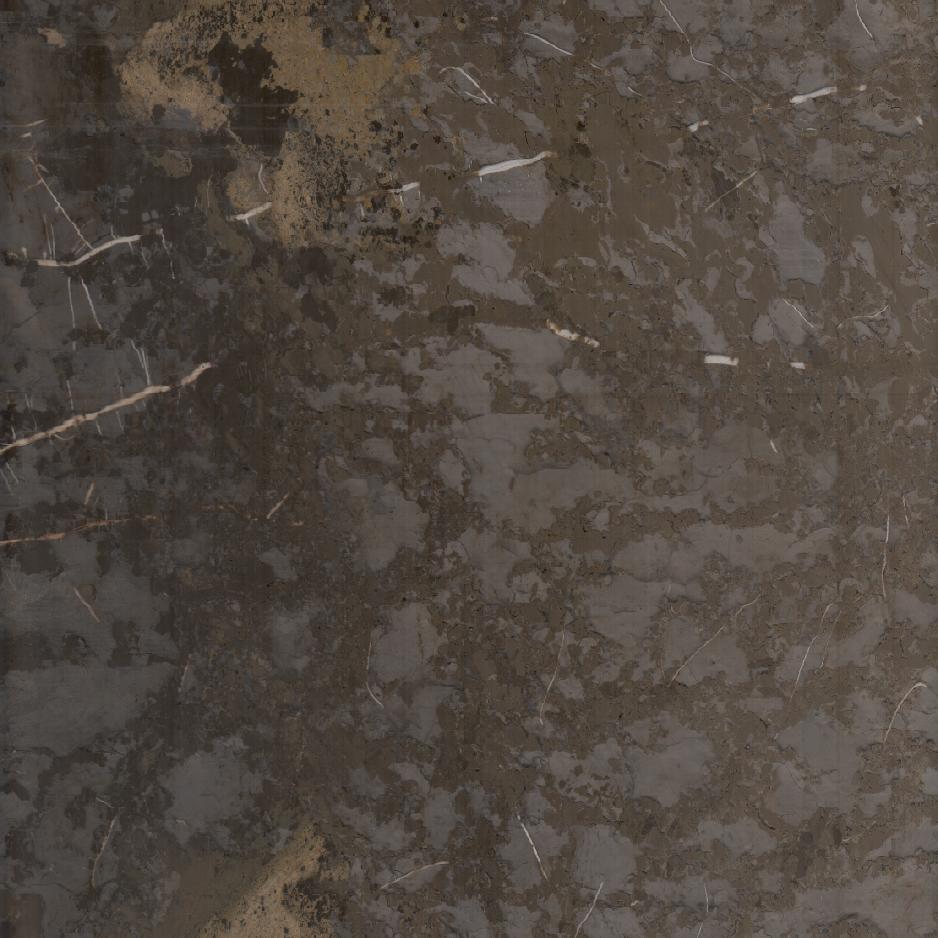}
\end{subfigure}
\begin{subfigure}{0.12\textwidth}
   \includegraphics[width=1\linewidth]{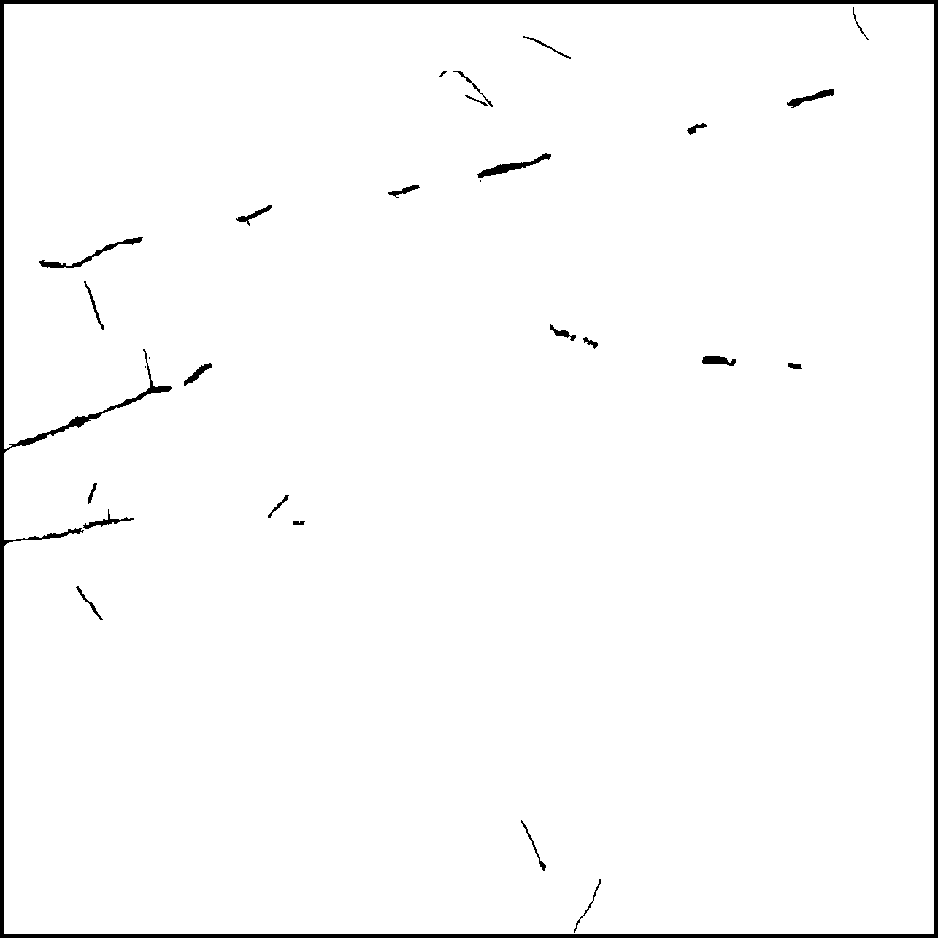}
\end{subfigure}
\begin{subfigure}{0.12\textwidth}
   \includegraphics[width=1\linewidth]{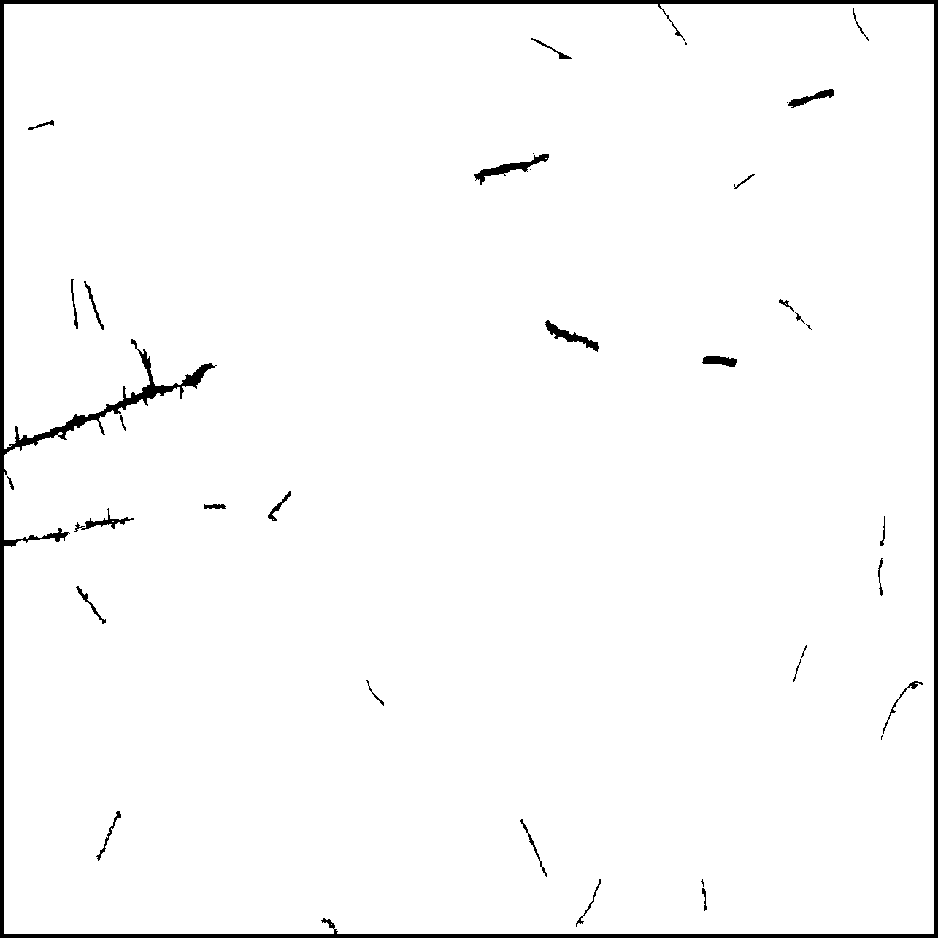}
\end{subfigure}
\begin{subfigure}{0.12\textwidth}
   \includegraphics[width=1\linewidth]{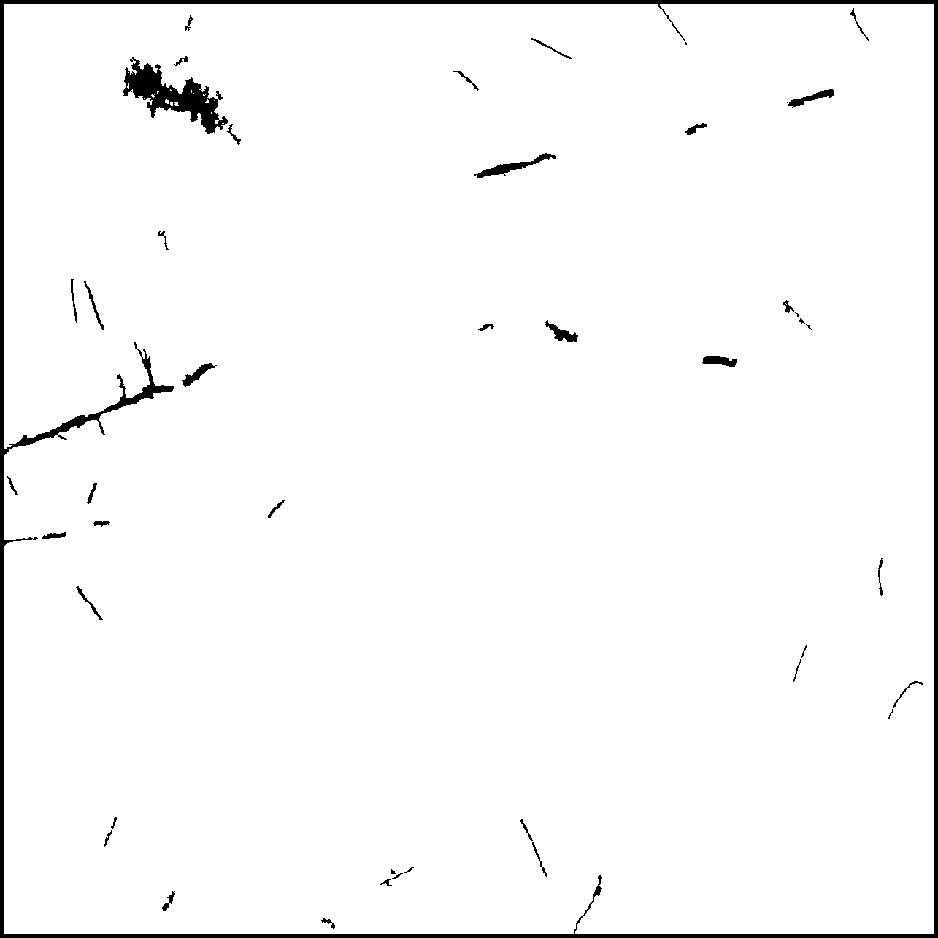}
\end{subfigure}
\begin{subfigure}{0.12\textwidth}
   \includegraphics[width=1\linewidth]{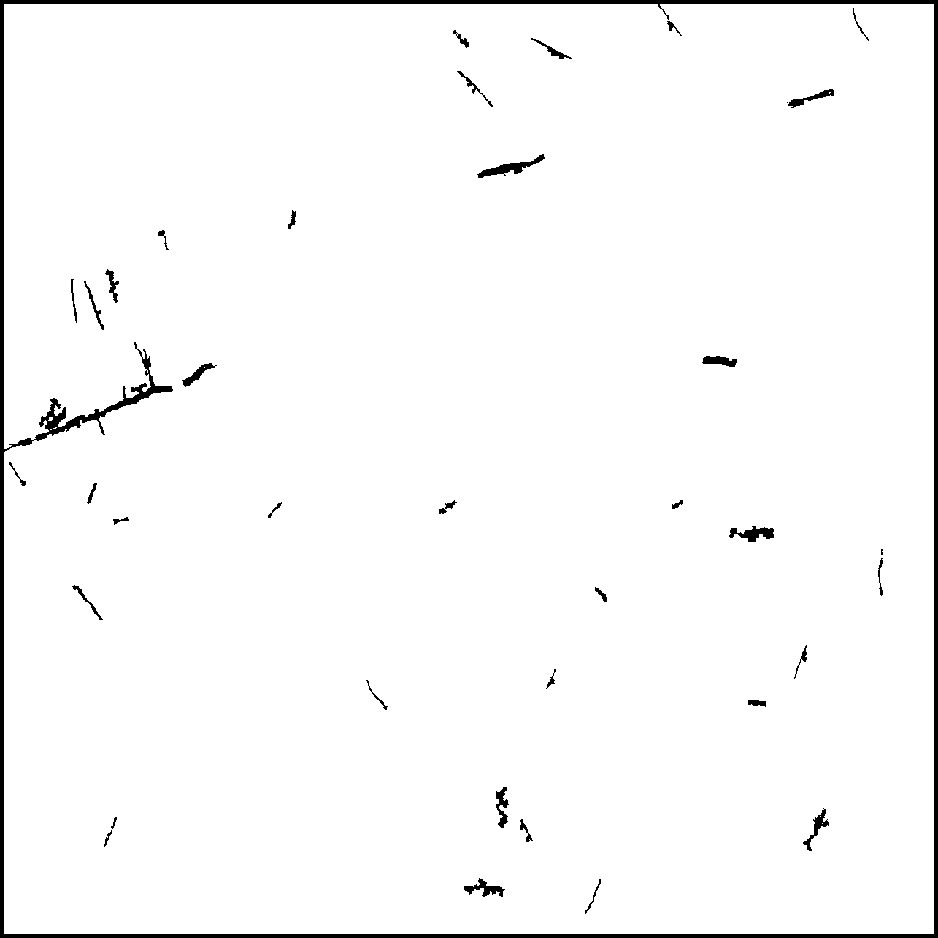}
\end{subfigure}
\begin{subfigure}{0.12\textwidth}
   \includegraphics[width=1\linewidth]{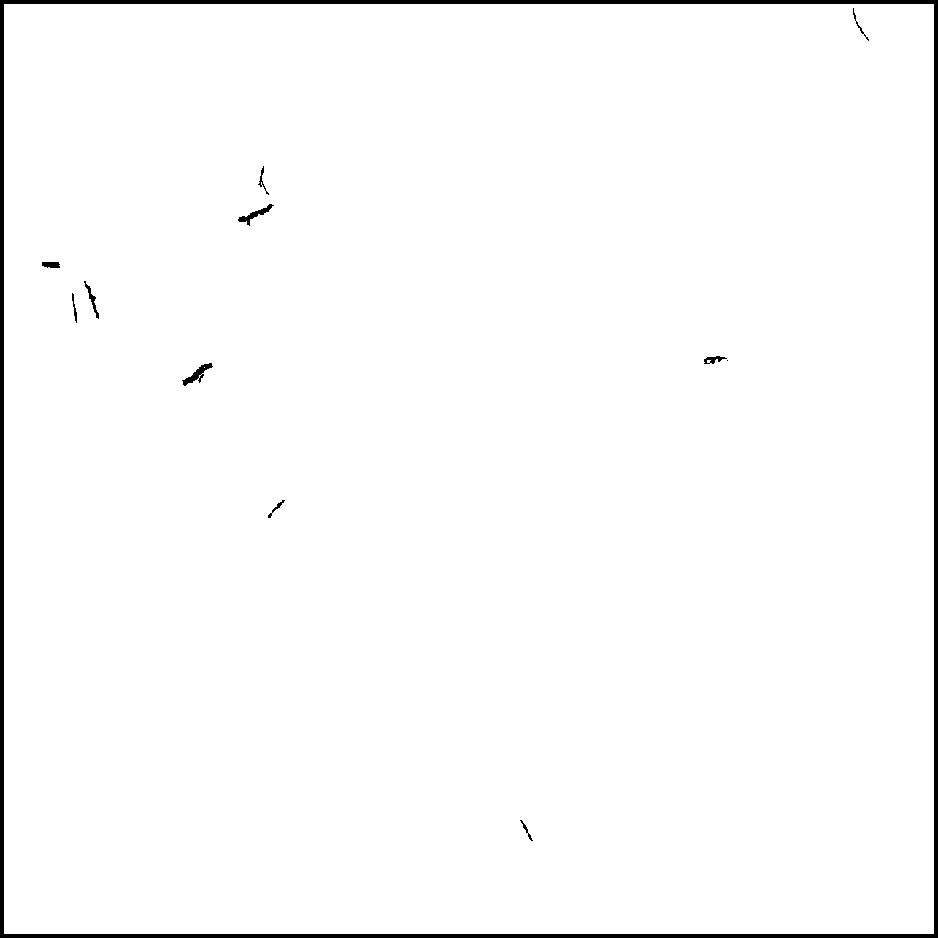}
\end{subfigure}
\begin{subfigure}{0.12\textwidth}
   \includegraphics[width=1\linewidth]{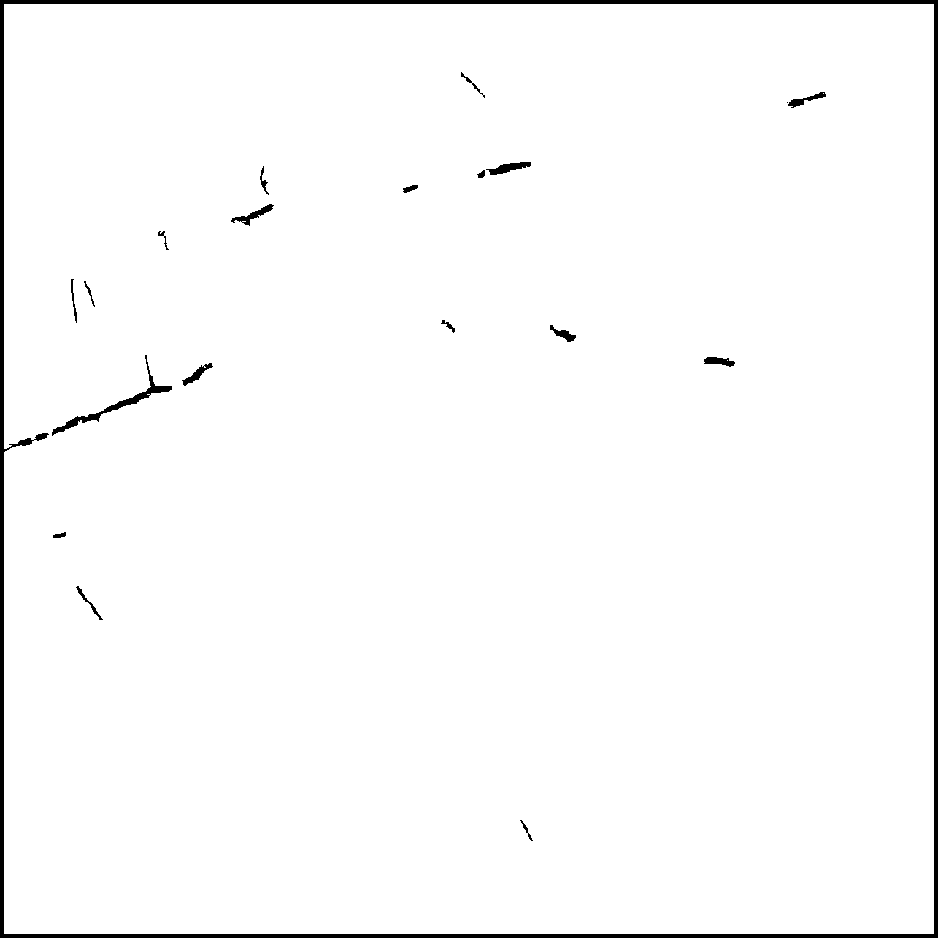}
\end{subfigure}
\\[\baselineskip]
\begin{subfigure}{0.12\textwidth}
   \includegraphics[width=1\linewidth]{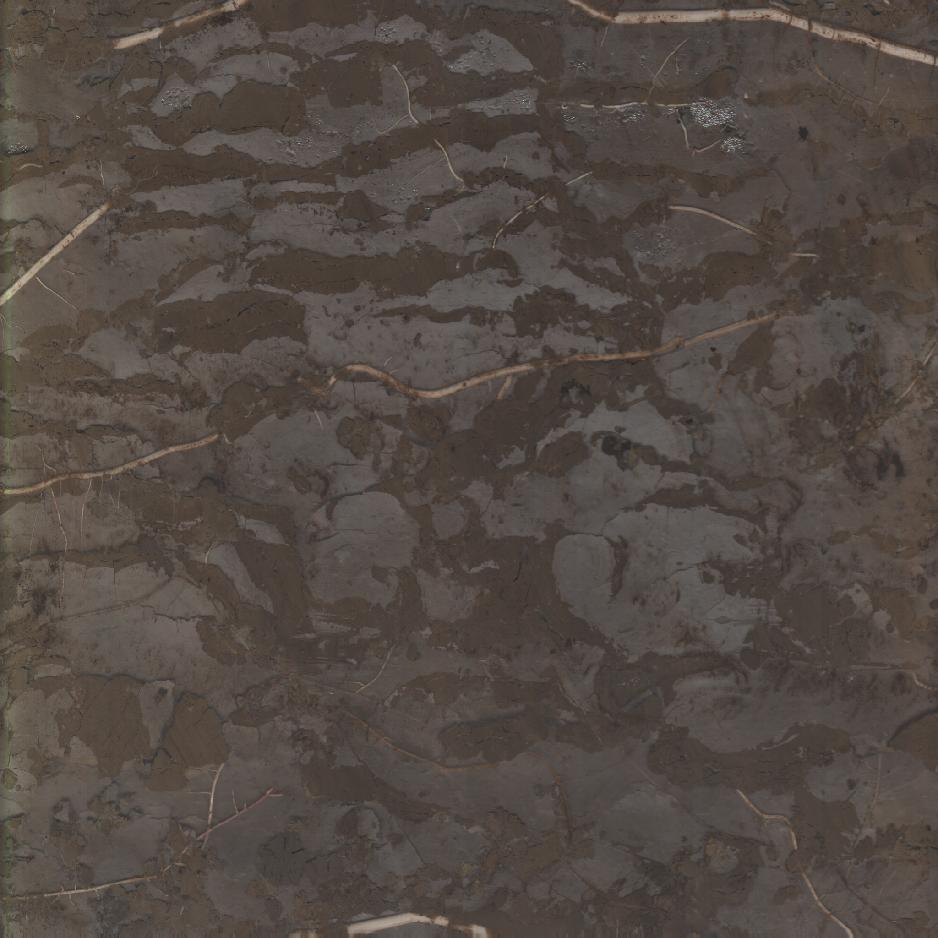}
\end{subfigure}
\begin{subfigure}{0.12\textwidth}
   \includegraphics[width=1\linewidth]{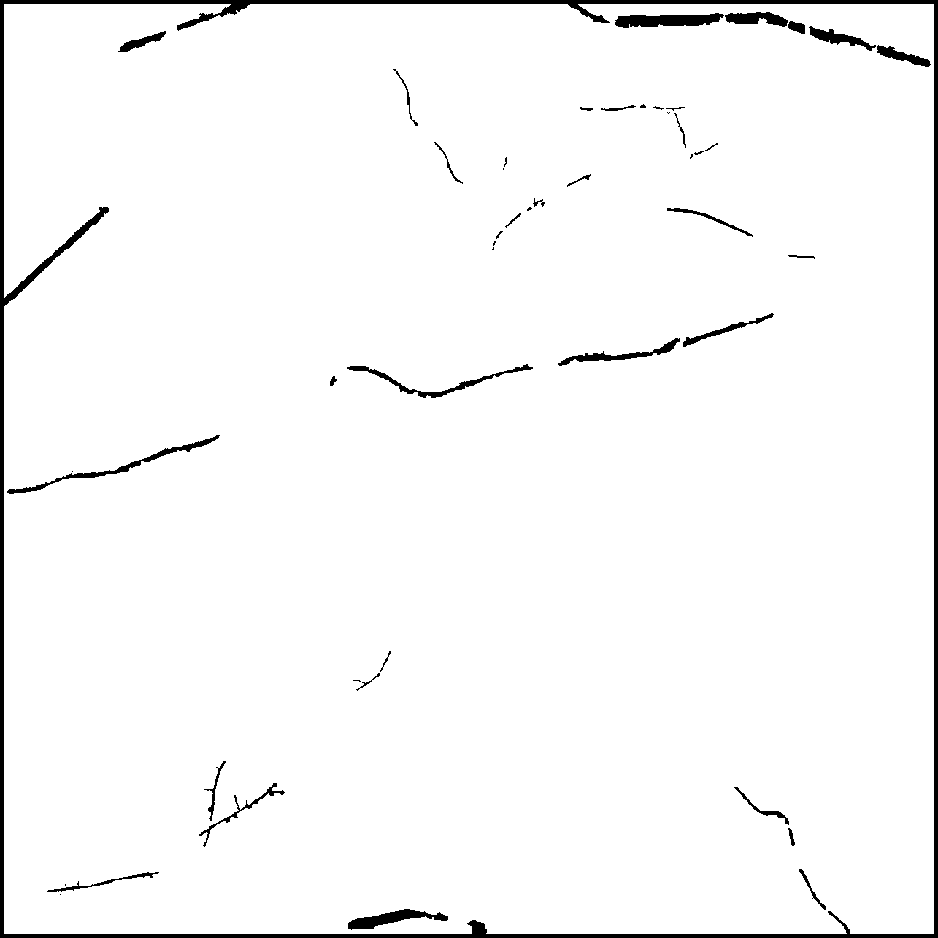}
\end{subfigure}
\begin{subfigure}{0.12\textwidth}
   \includegraphics[width=1\linewidth]{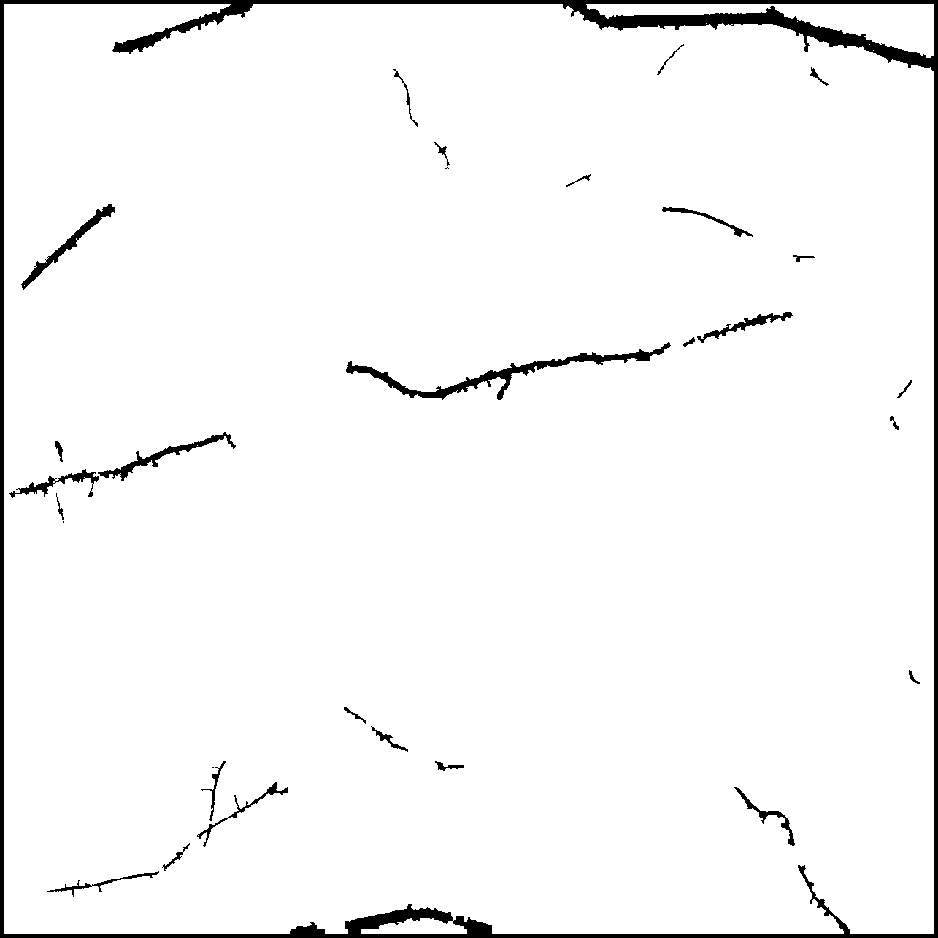}
\end{subfigure}
\begin{subfigure}{0.12\textwidth}
   \includegraphics[width=1\linewidth]{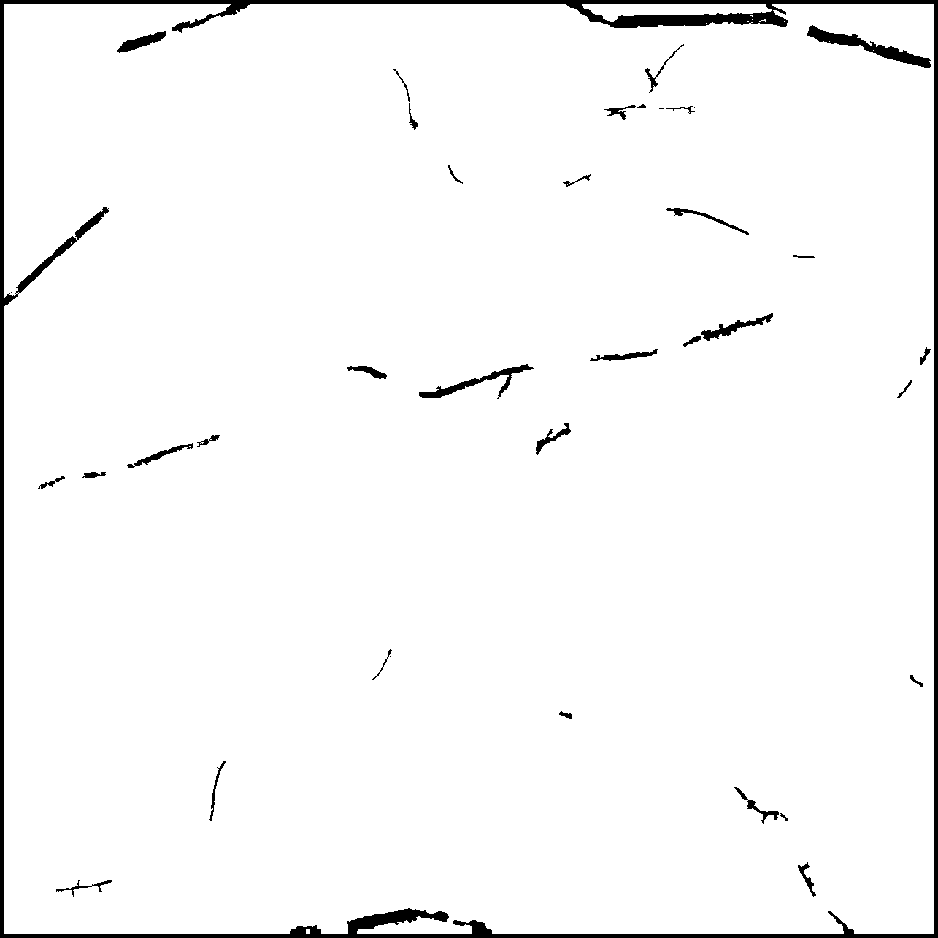}
\end{subfigure}
\begin{subfigure}{0.12\textwidth}
   \includegraphics[width=1\linewidth]{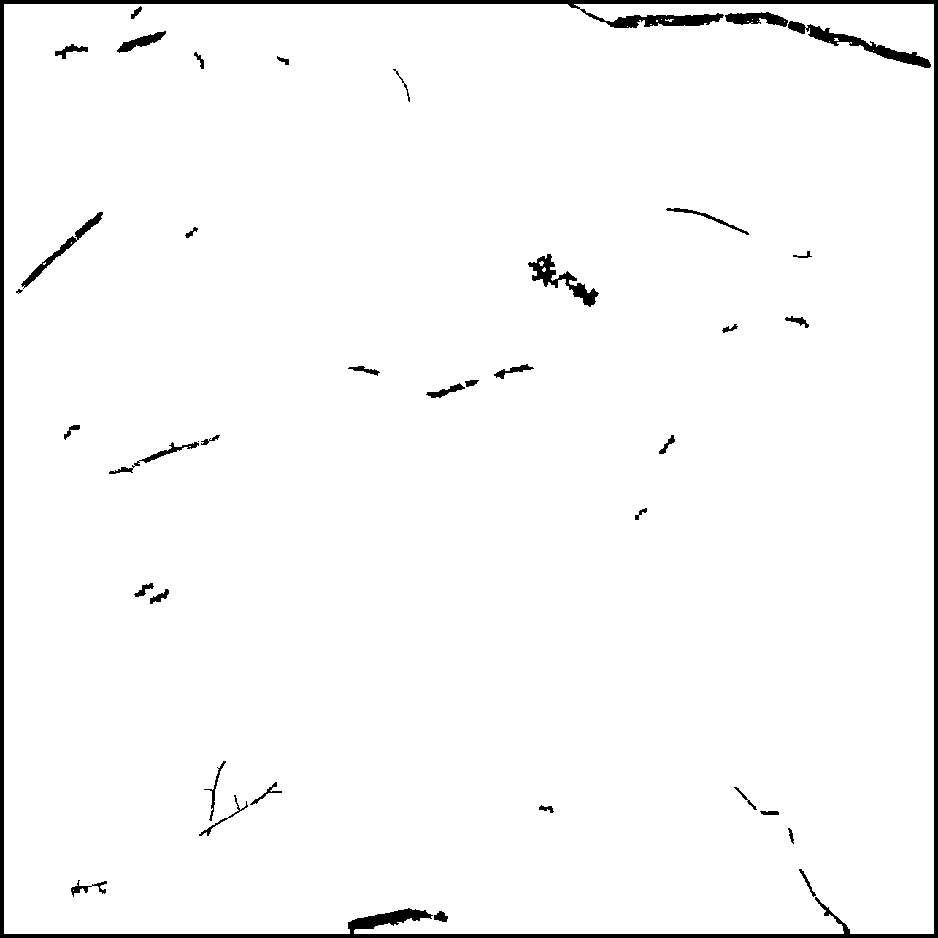}
\end{subfigure}
\begin{subfigure}{0.12\textwidth}
   \includegraphics[width=1\linewidth]{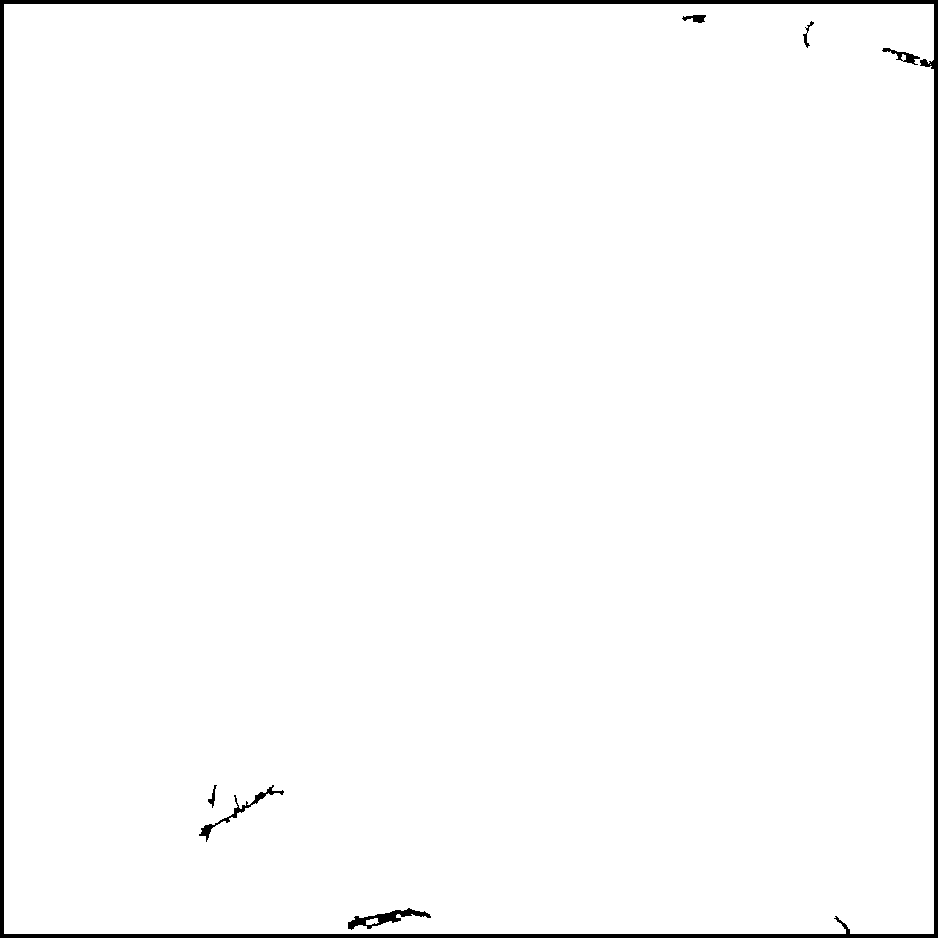}
\end{subfigure}
\begin{subfigure}{0.12\textwidth}
   \includegraphics[width=1\linewidth]{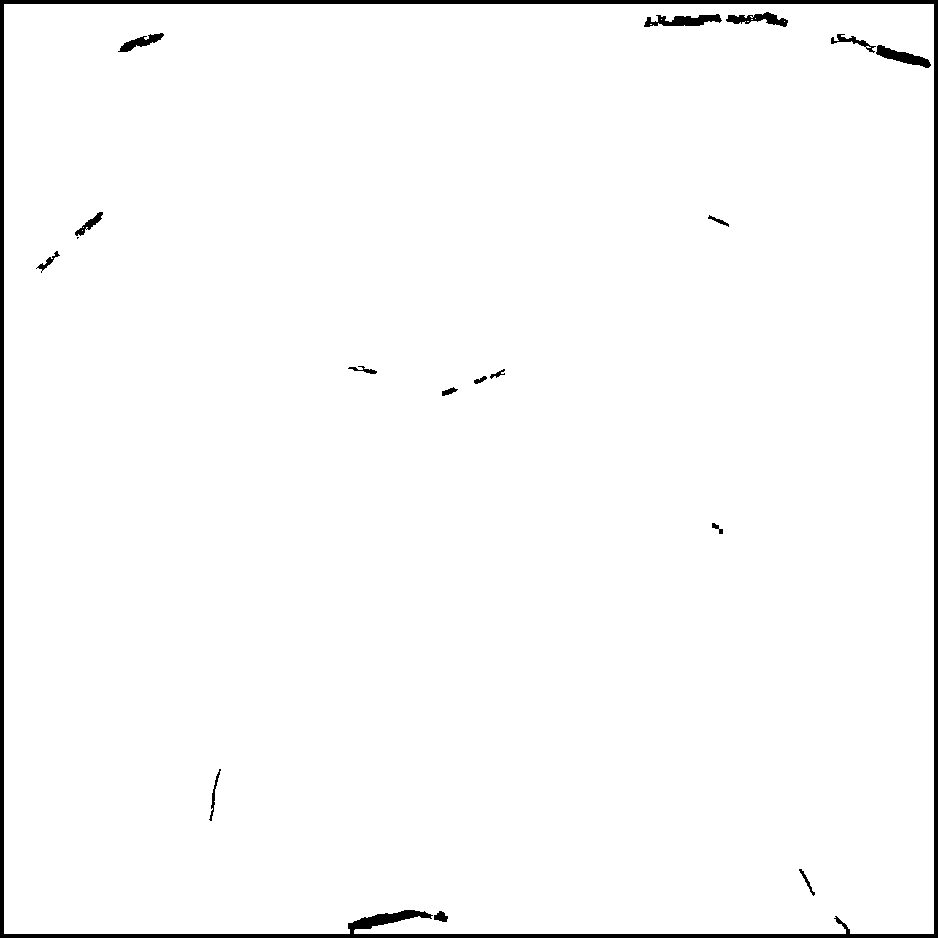}
\end{subfigure}
\\[\baselineskip]
\begin{subfigure}{0.12\textwidth}
   \includegraphics[width=1\linewidth]{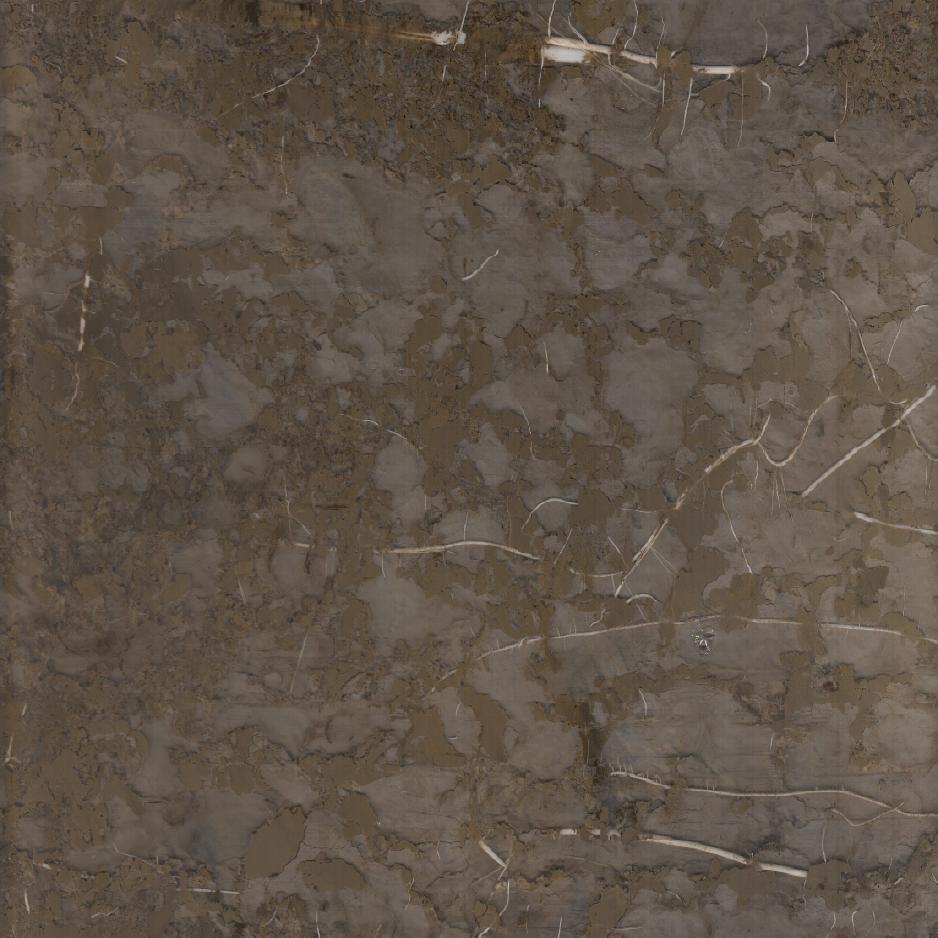}
\end{subfigure}
\begin{subfigure}{0.12\textwidth}
   \includegraphics[width=1\linewidth]{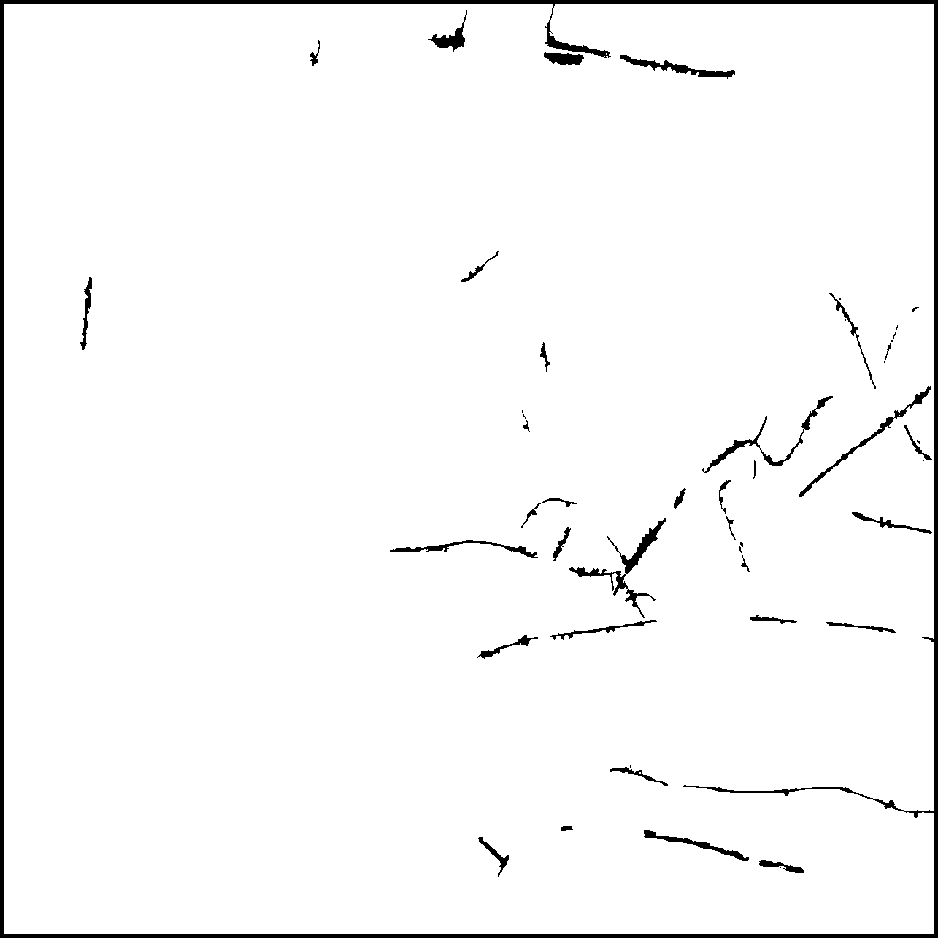}
\end{subfigure}
\begin{subfigure}{0.12\textwidth}
   \includegraphics[width=1\linewidth]{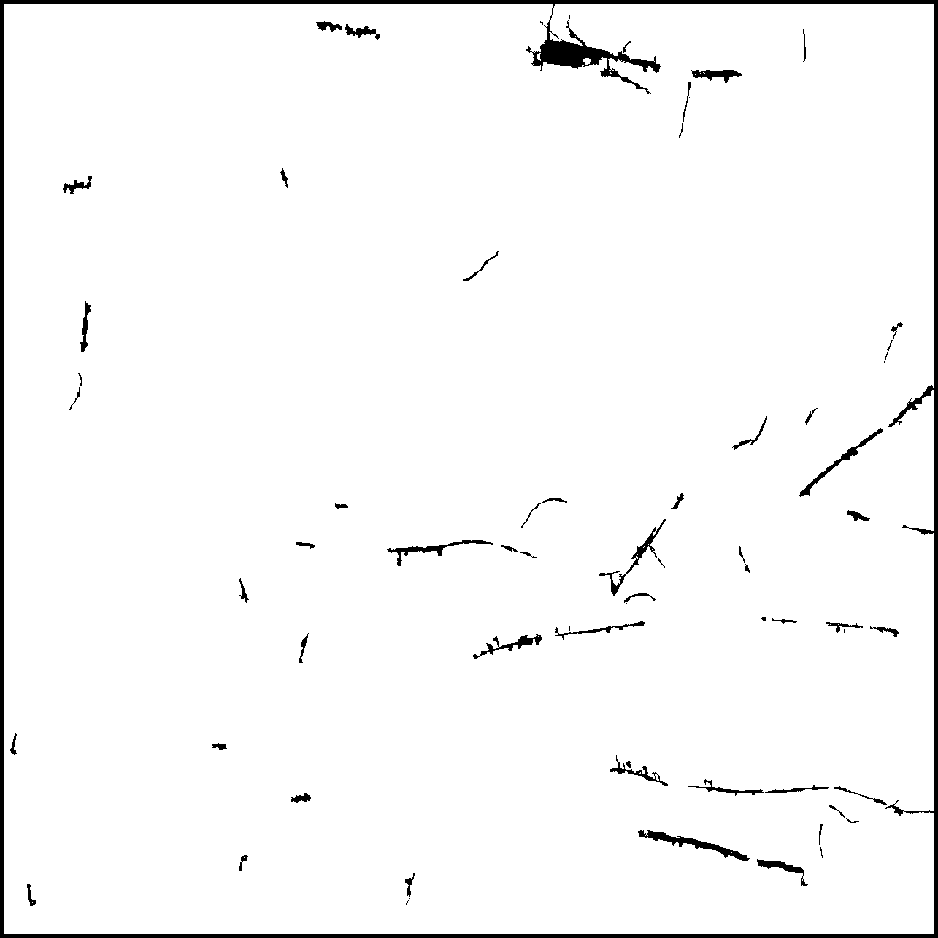}
\end{subfigure}
\begin{subfigure}{0.12\textwidth}
   \includegraphics[width=1\linewidth]{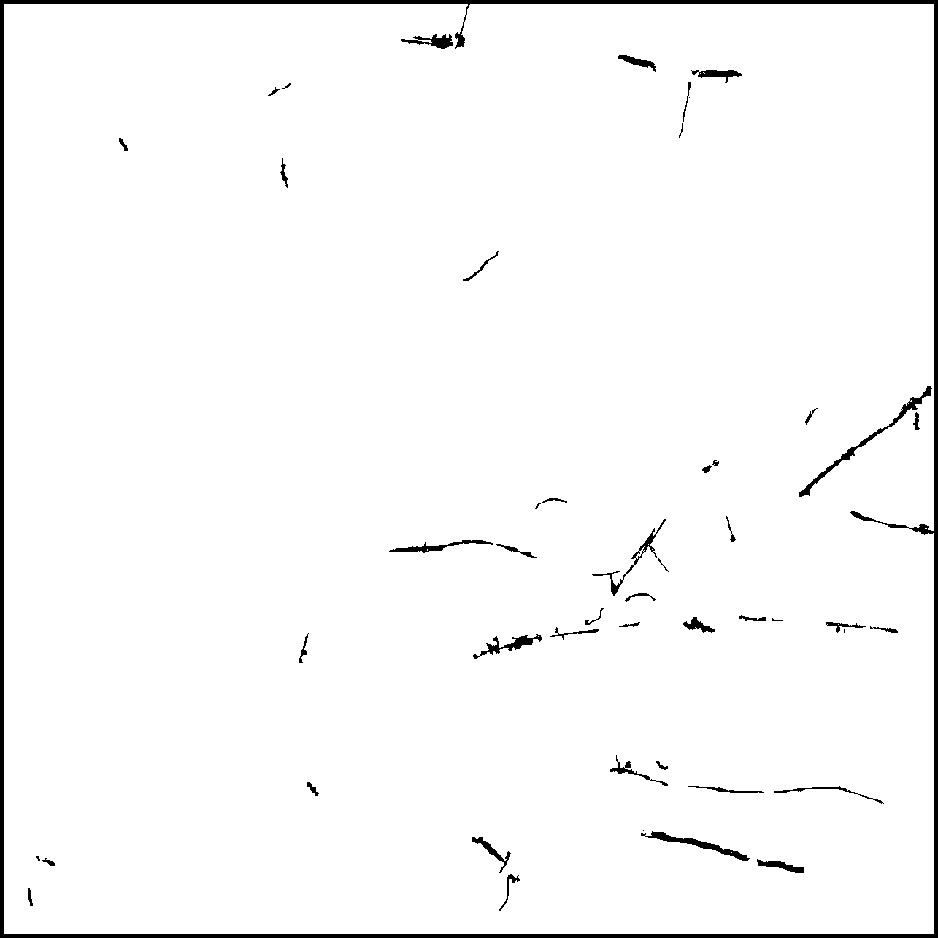}
\end{subfigure}
\begin{subfigure}{0.12\textwidth}
   \includegraphics[width=1\linewidth]{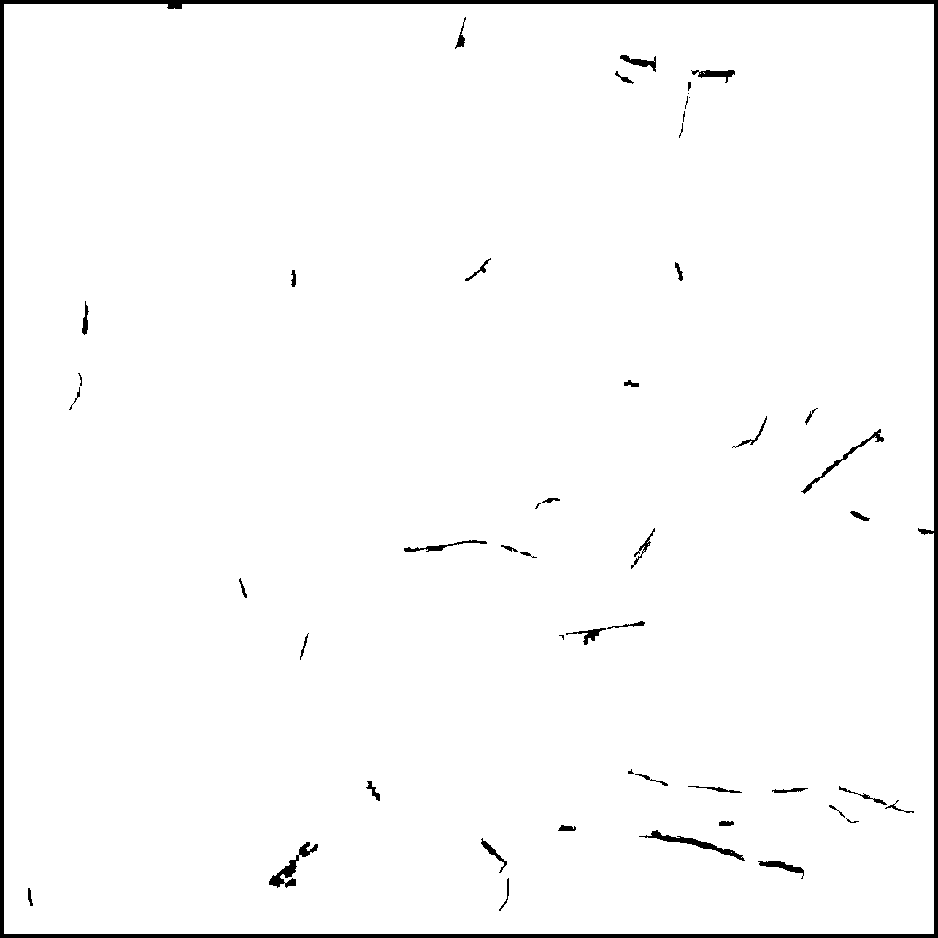}
\end{subfigure}
\begin{subfigure}{0.12\textwidth}
   \includegraphics[width=1\linewidth]{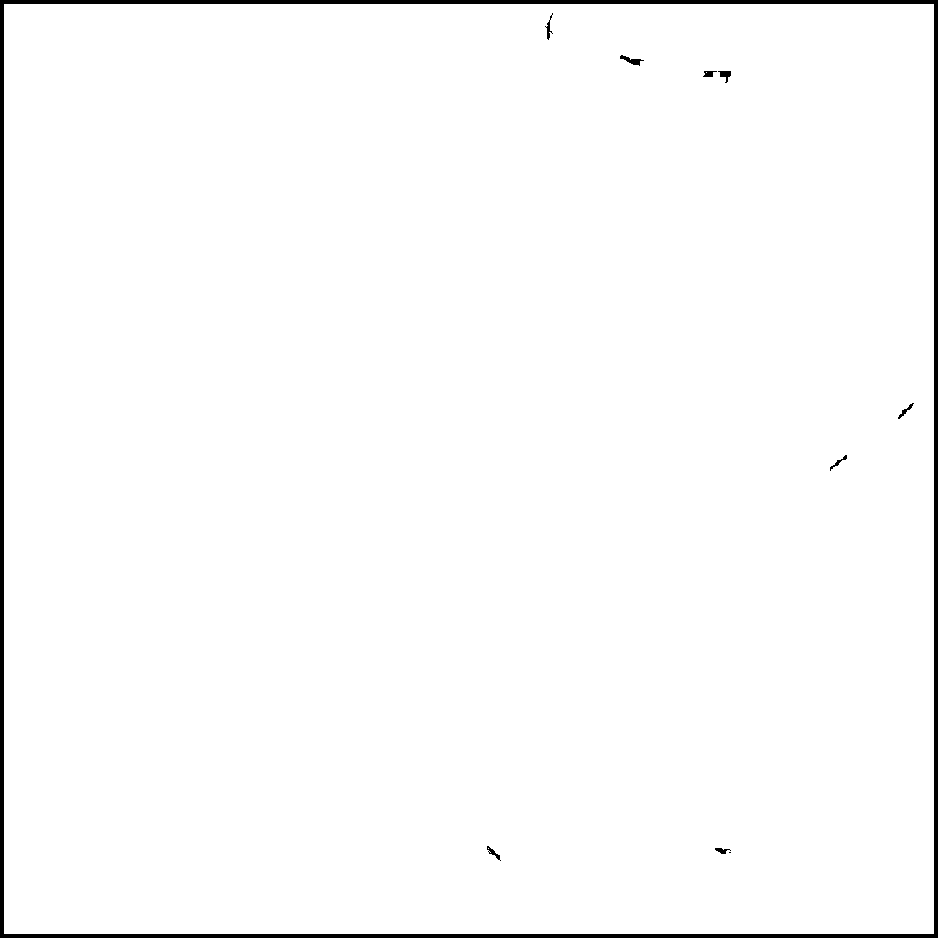}
\end{subfigure}
\begin{subfigure}{0.12\textwidth}
   \includegraphics[width=1\linewidth]{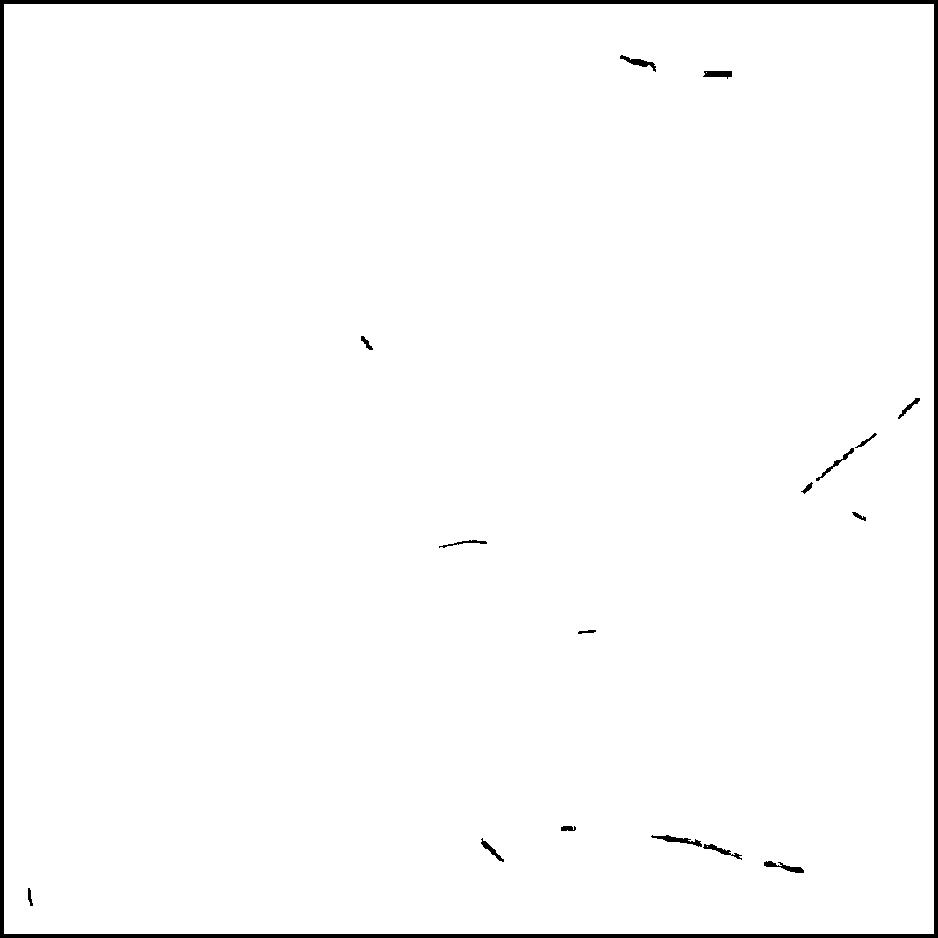}
\end{subfigure}
\\[\baselineskip]
\begin{subfigure}{0.12\textwidth}
   \includegraphics[width=1\linewidth]{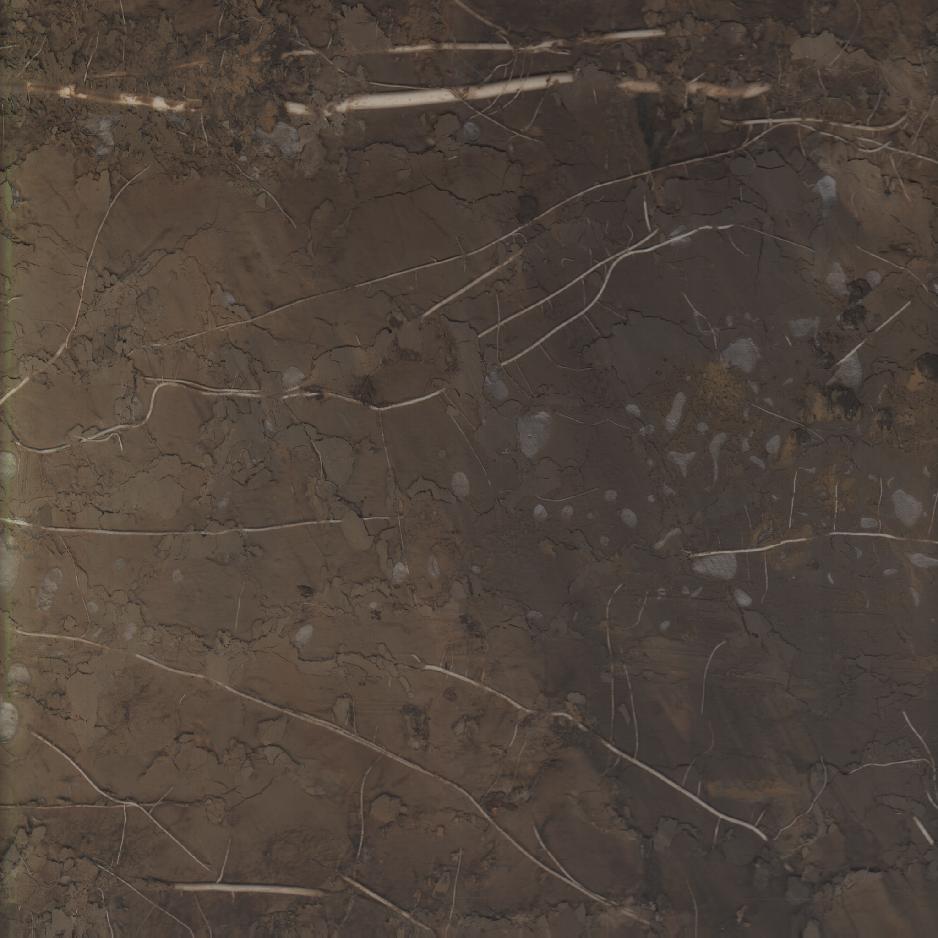}
\end{subfigure}
\begin{subfigure}{0.12\textwidth}
   \includegraphics[width=1\linewidth]{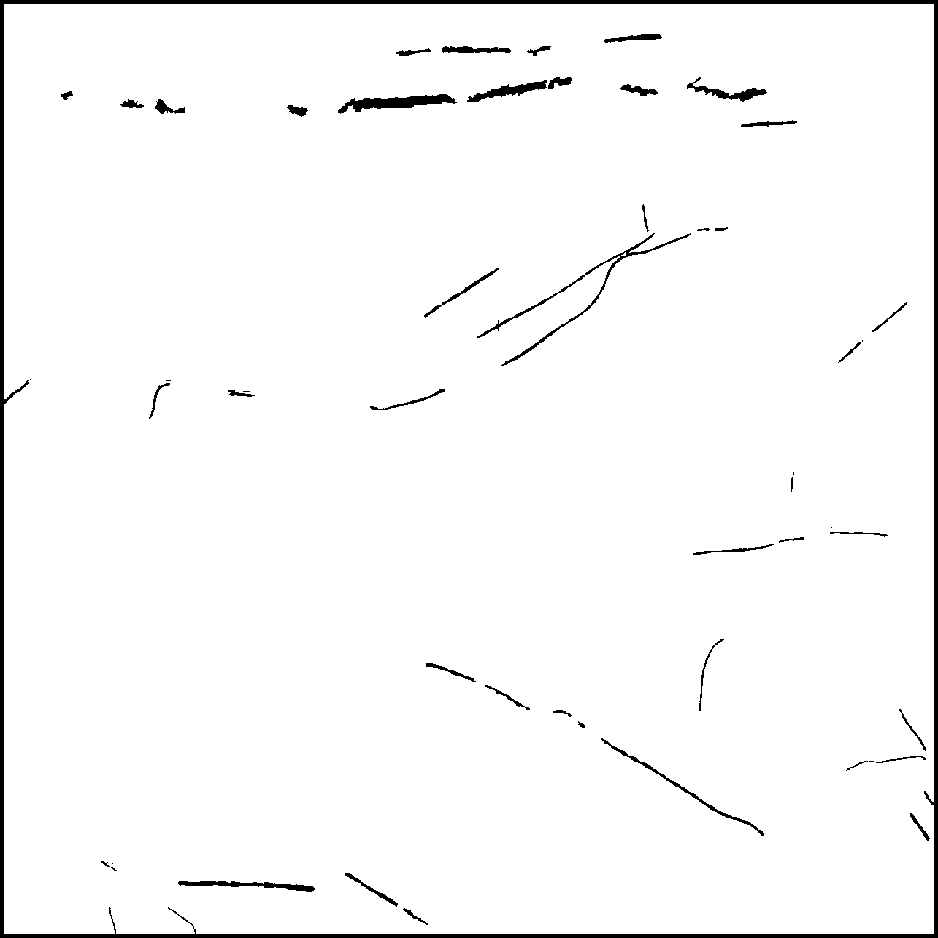}
\end{subfigure}
\begin{subfigure}{0.12\textwidth}
   \includegraphics[width=1\linewidth]{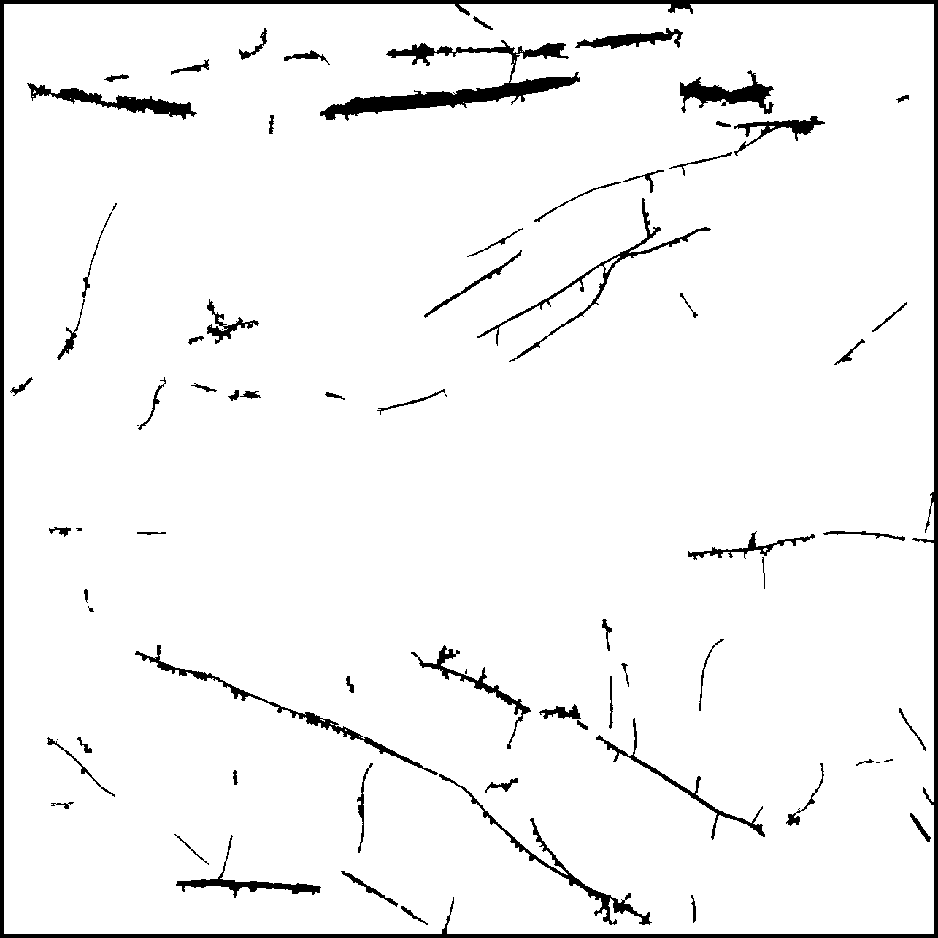}
\end{subfigure}
\begin{subfigure}{0.12\textwidth}
   \includegraphics[width=1\linewidth]{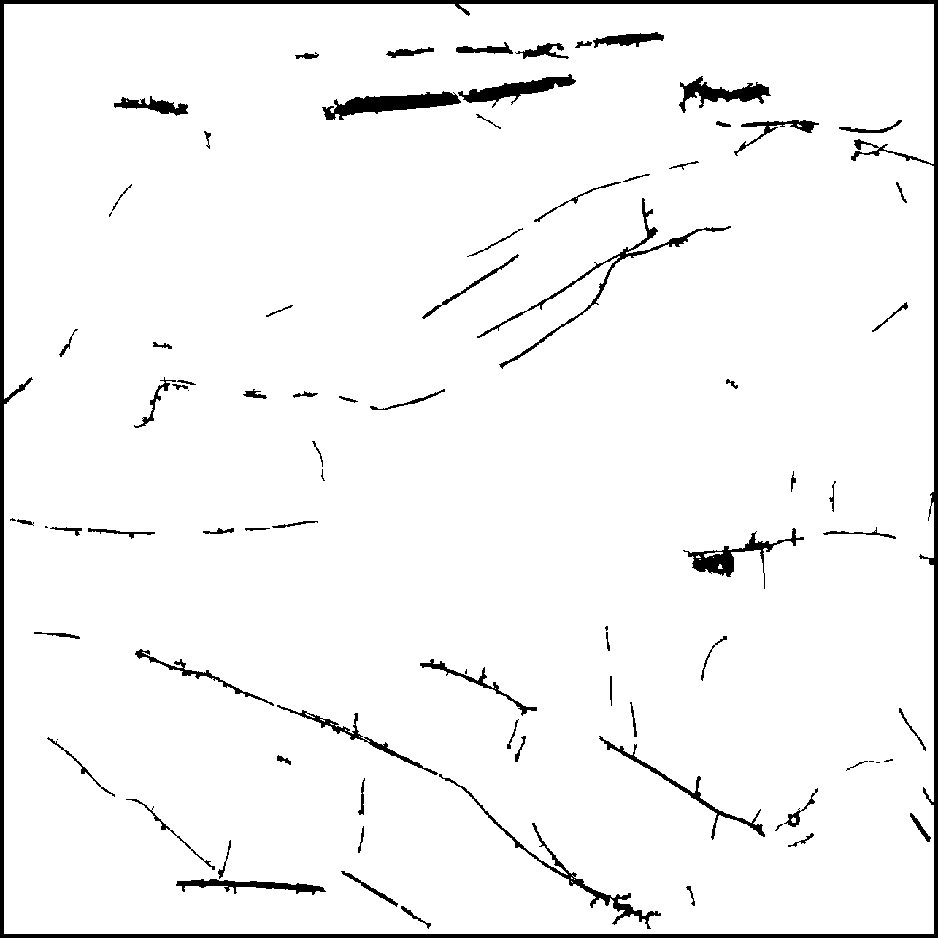}
\end{subfigure}
\begin{subfigure}{0.12\textwidth}
   \includegraphics[width=1\linewidth]{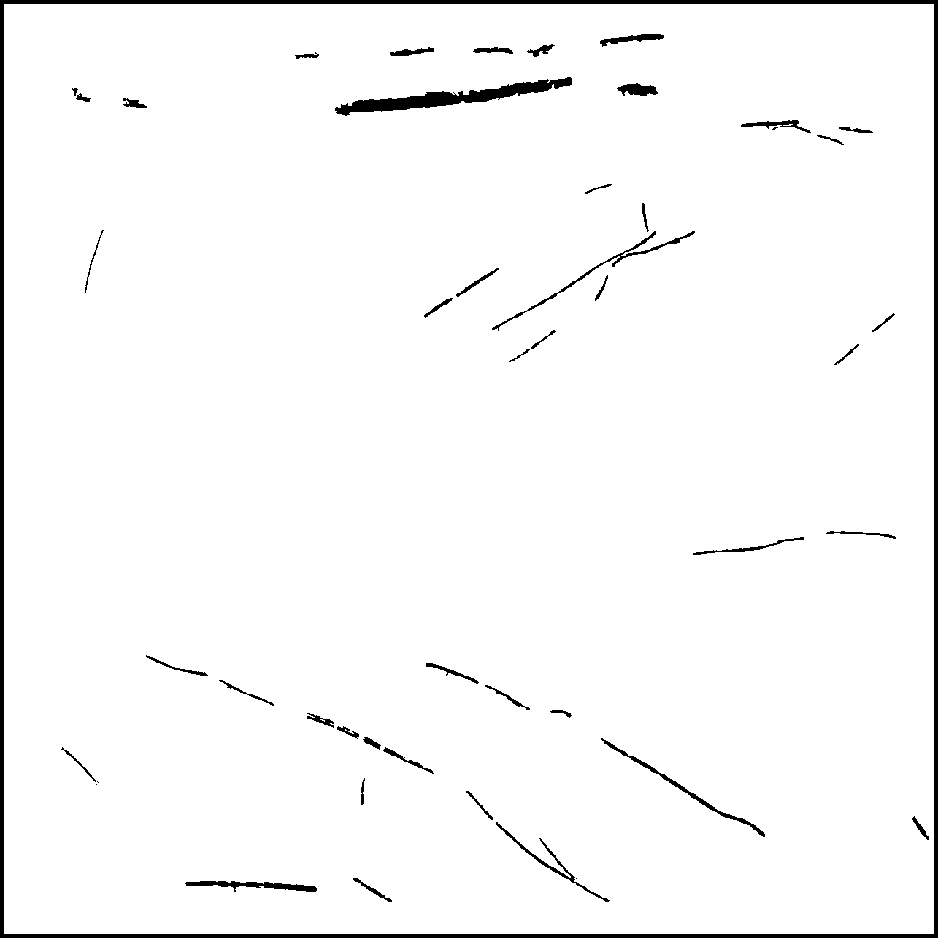}
\end{subfigure}
\begin{subfigure}{0.12\textwidth}
   \includegraphics[width=1\linewidth]{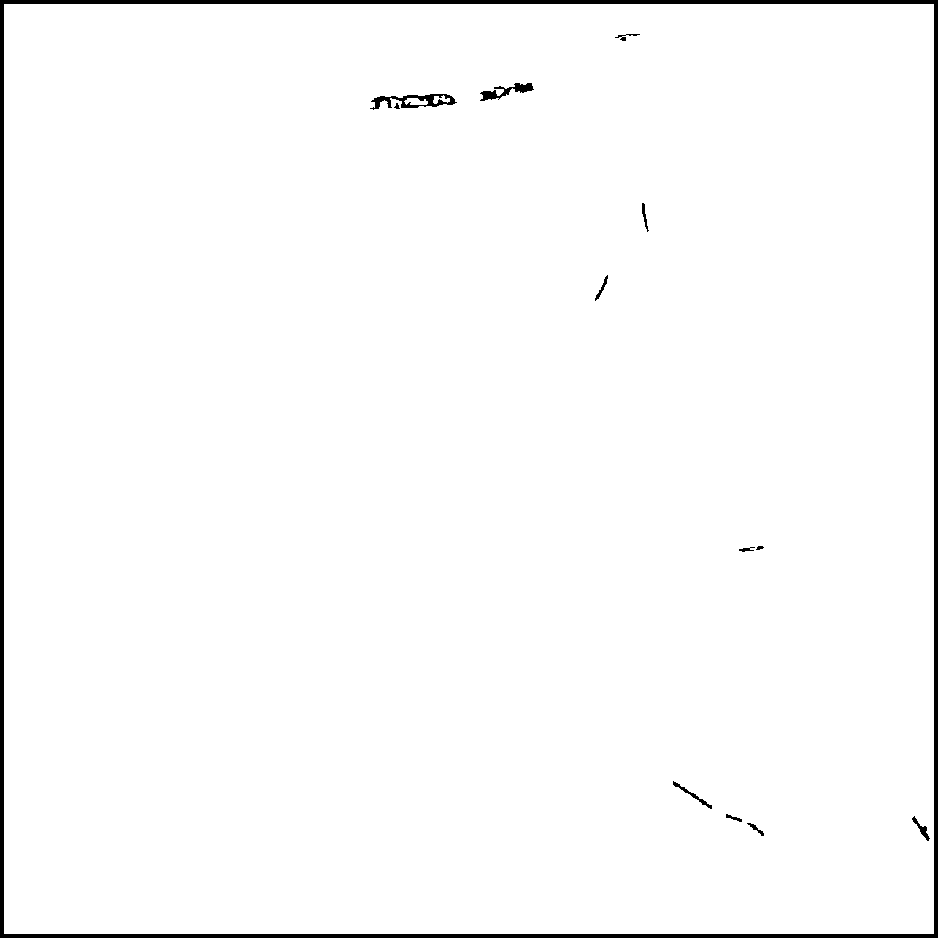}
\end{subfigure}
\begin{subfigure}{0.12\textwidth}
   \includegraphics[width=1\linewidth]{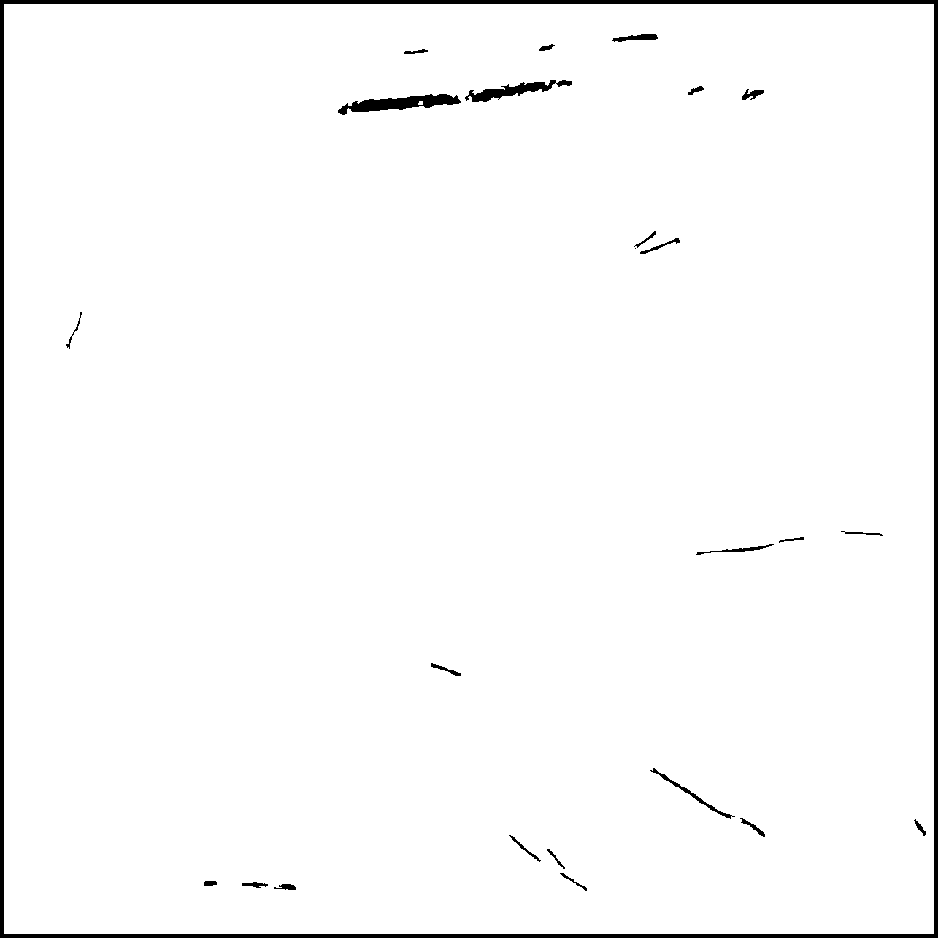}
\end{subfigure}
\\[\baselineskip]
\begin{subfigure}{0.12\textwidth}
   \includegraphics[width=1\linewidth]{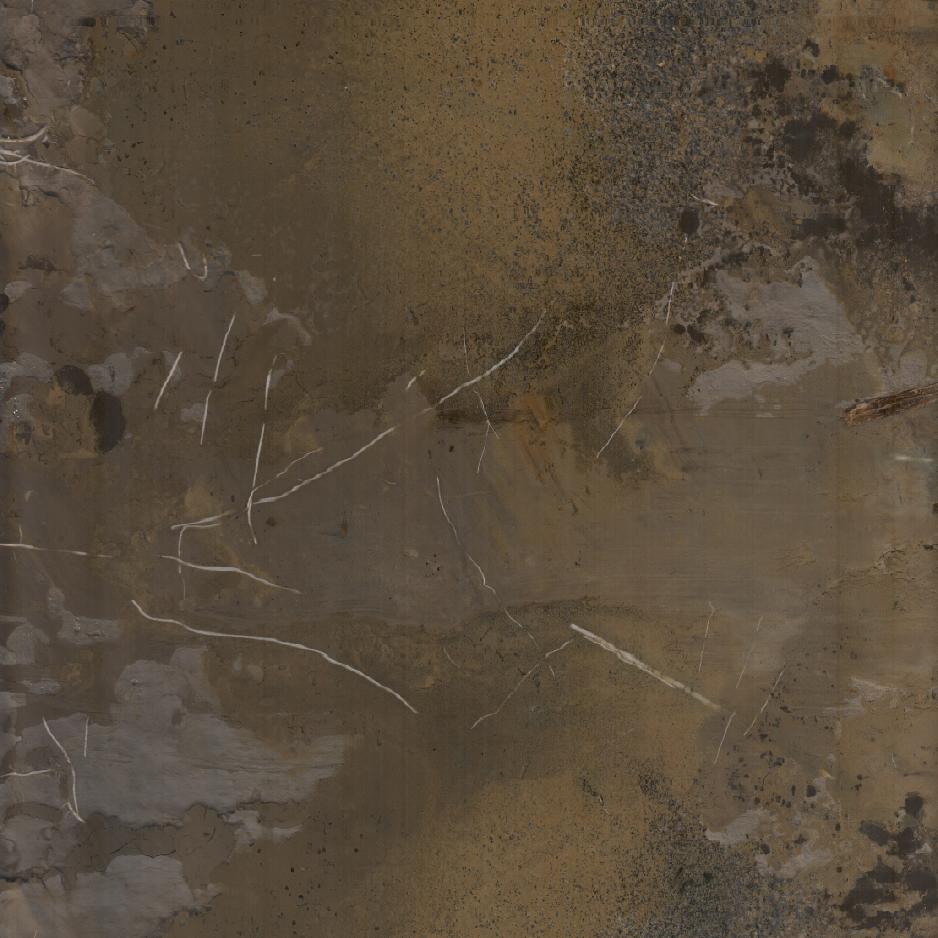}
   \caption{Images}
   \label{subfig:8_A}
\end{subfigure}
\begin{subfigure}{0.12\textwidth}
   \includegraphics[width=1\linewidth]{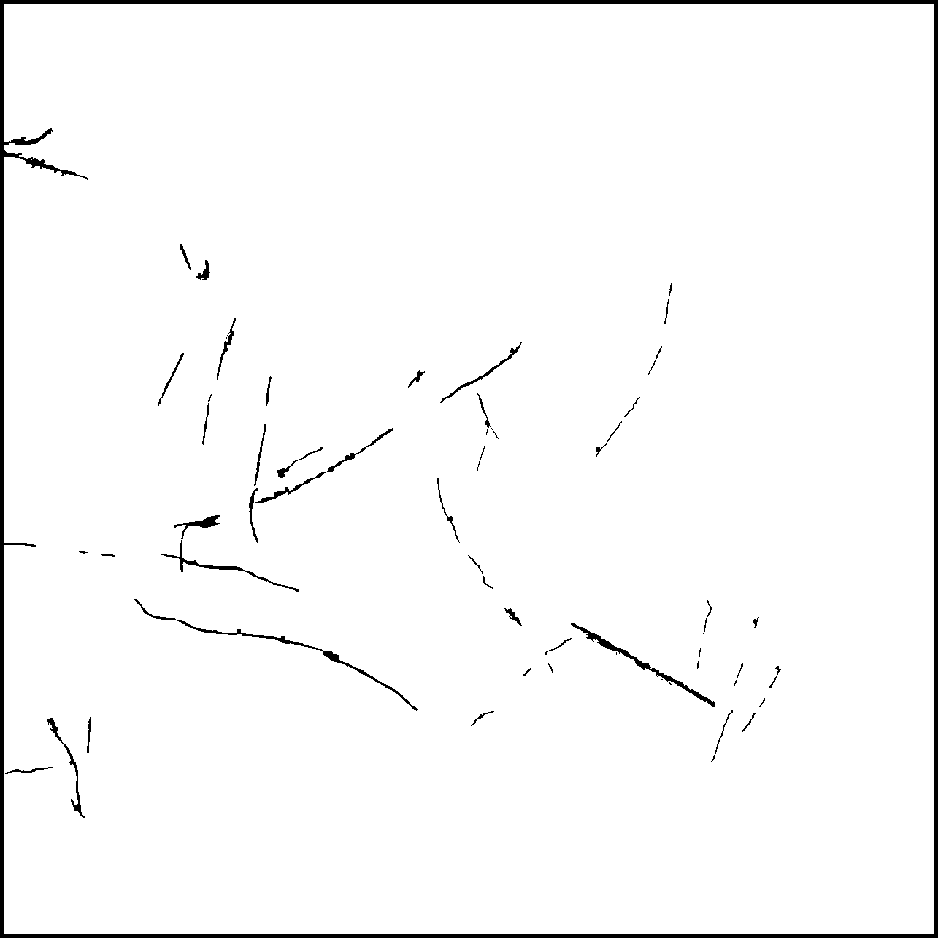}
   \caption{GT}
   \label{subfig:8_B}
\end{subfigure}
\begin{subfigure}{0.12\textwidth}
   \includegraphics[width=1\linewidth]{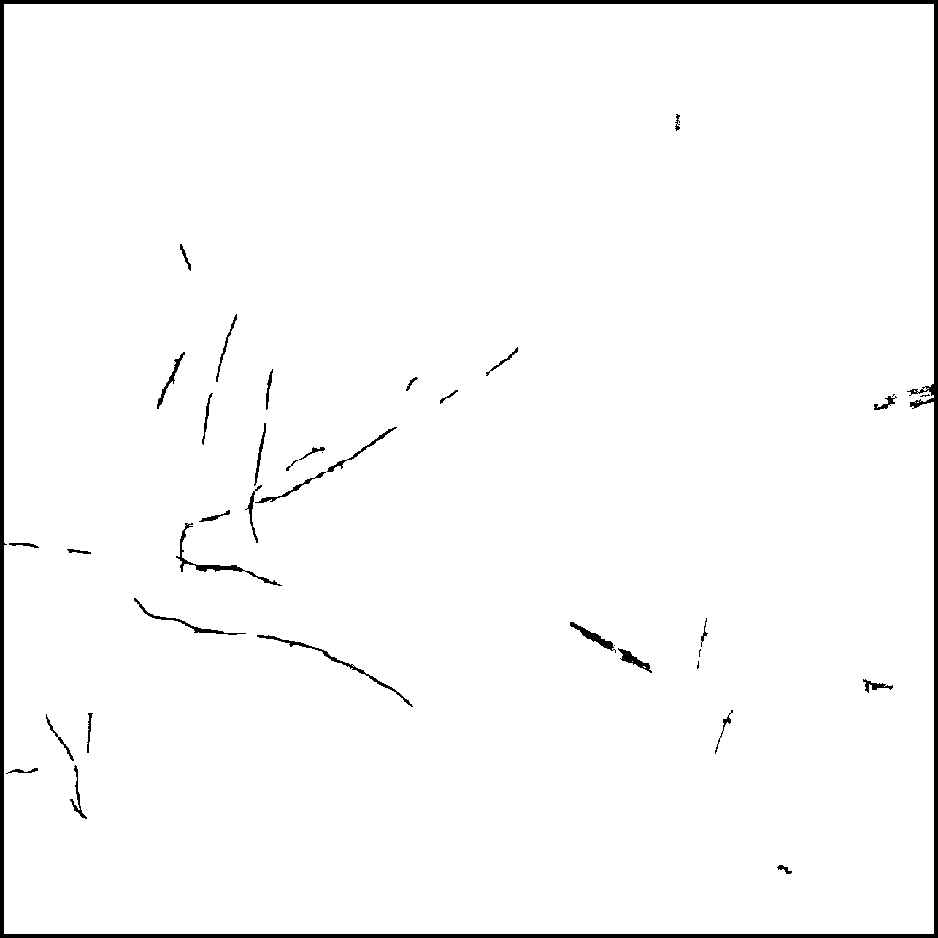}
   \caption{miSVM}
   \label{subfig:8_C}
\end{subfigure}
\begin{subfigure}{0.12\textwidth}
   \includegraphics[width=1\linewidth]{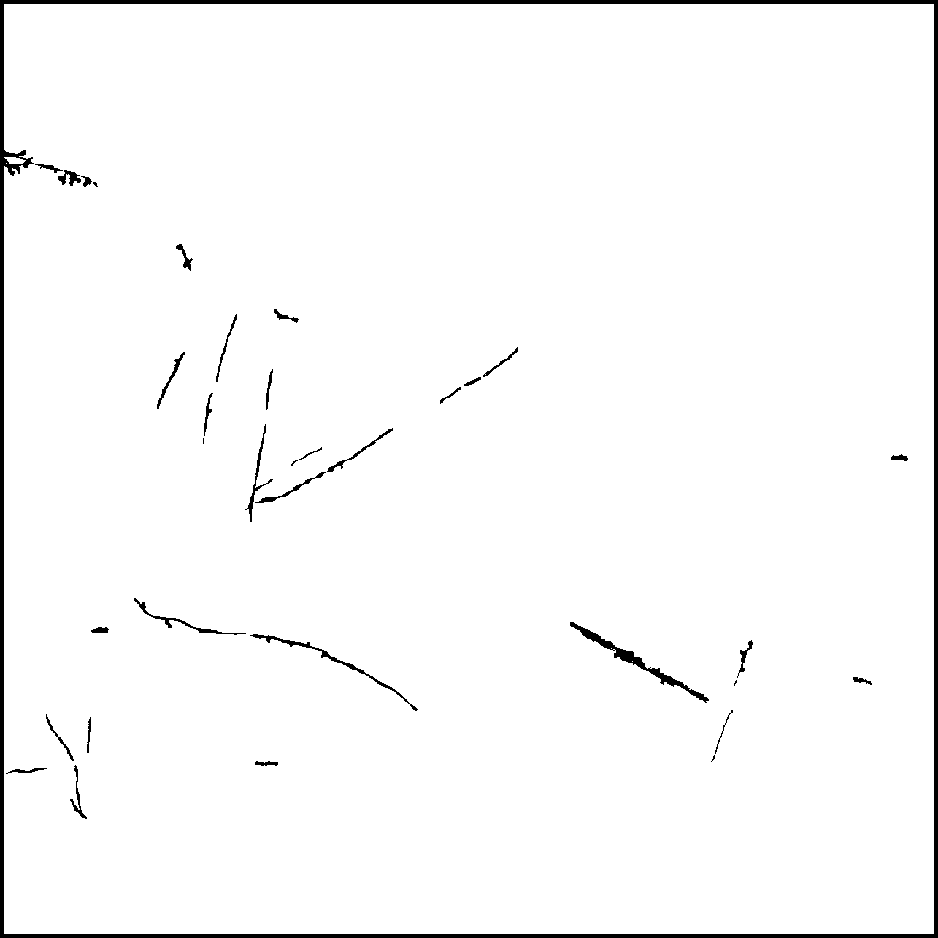}
   \caption{MI-ACE}
   \label{subfig:8_D}
\end{subfigure}
\begin{subfigure}{0.12\textwidth}
   \includegraphics[width=1\linewidth]{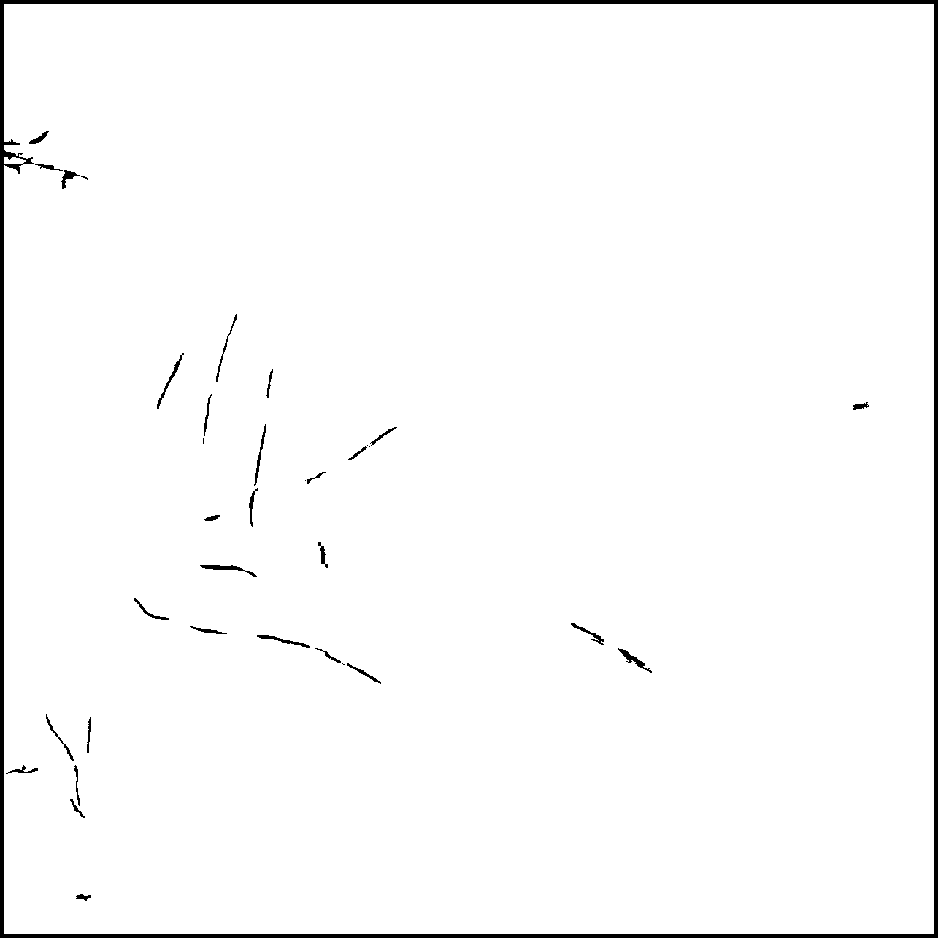}
   \caption{MIForests}
   \label{subfig:8_E}
\end{subfigure}
\begin{subfigure}{0.12\textwidth}
   \includegraphics[width=1\linewidth]{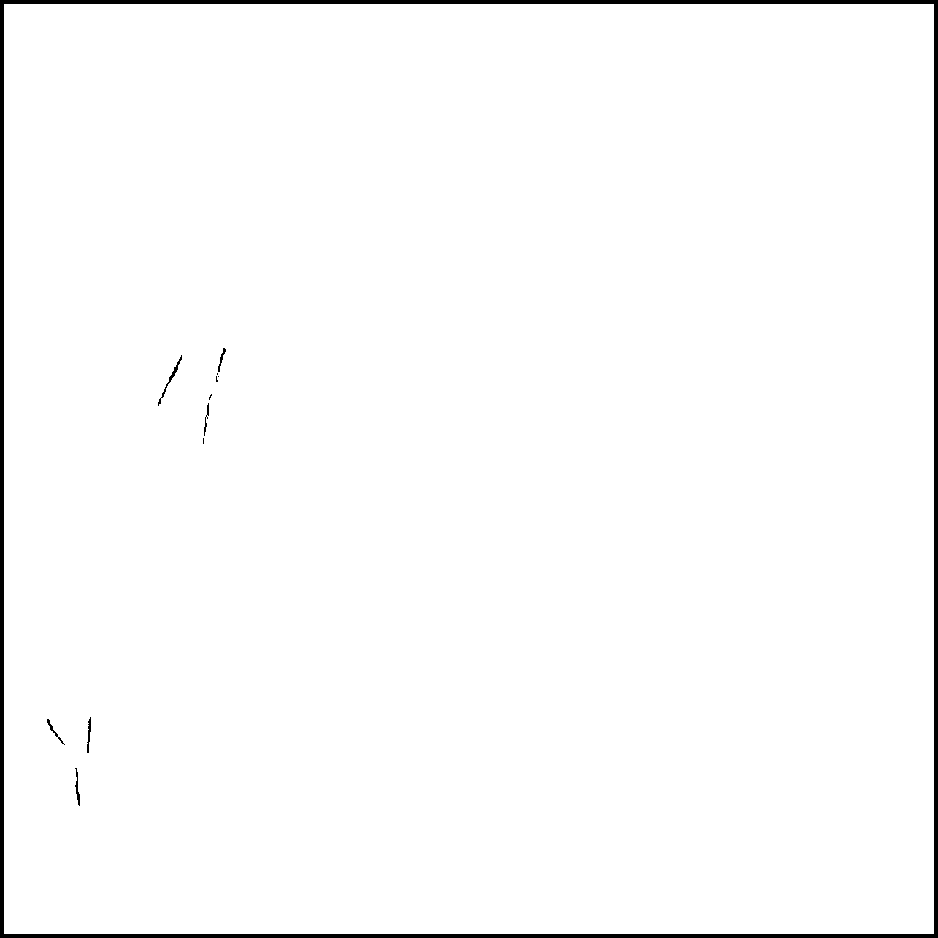}
   \caption{SVM}
   \label{subfig:8_F}
\end{subfigure}
\begin{subfigure}{0.12\textwidth}
   \includegraphics[width=1\linewidth]{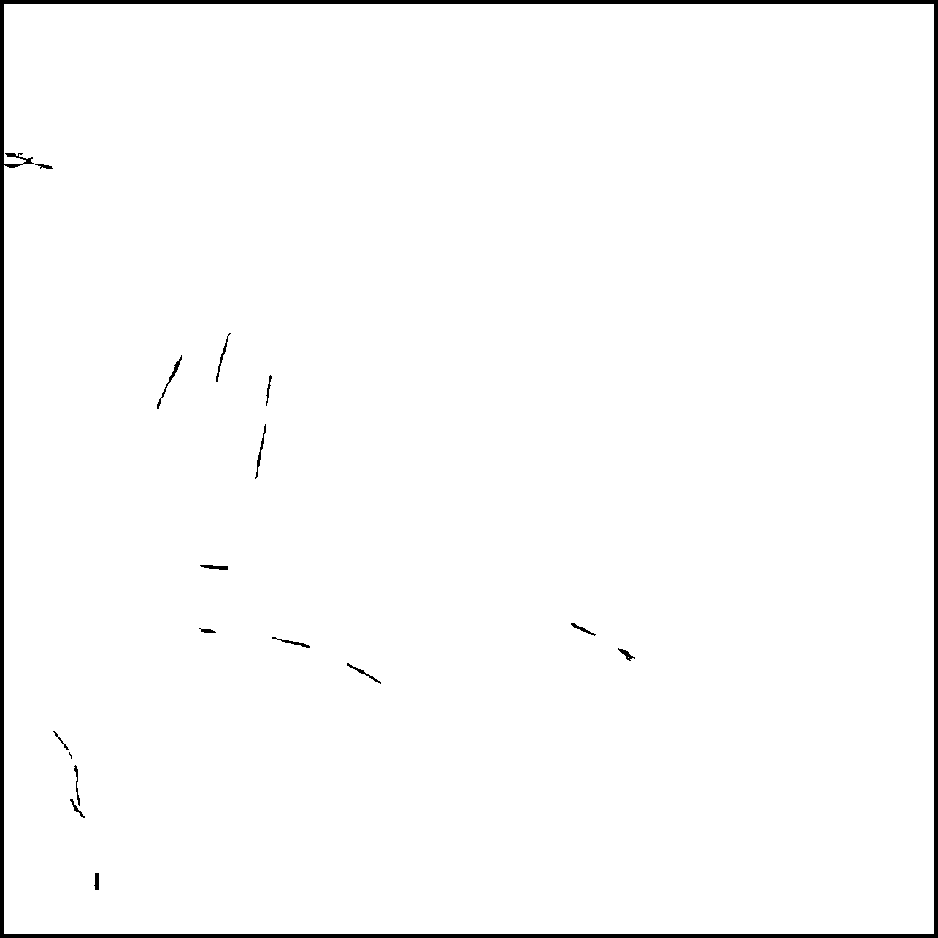}
   \caption{RF}
   \label{subfig:8_G}
\end{subfigure}

   \caption{Qualitative examples of root segmentation results with different method. The 1st column is the original images. The 2nd column is the groundtruth (GT)}
\label{fig:8}
\end{center}
\end{figure*}

\noindent\textbf {Qualitative results}:
Root detection results were qualitatively compared in Fig. \ref{fig:8} at threshold that FPR = 0.03 for different methods. The results were also post processed by manually setting the thresholds of area and eccentricity to be 300 and 0.95 respectively. The thresholds were determined by identifying the threshold value that provided a boost in performance over results without post-processing. In general, the MIL methods outperformed the non-MIL methods. The miSVM method detected the most roots in images and performed the best among all these methods.

\subsection{Experiment: Effect of Label Accuracy}
\subsubsection{Sample versus Image Level Label Training}
In this experiment, we compared the detection accuracy of the different methods when using training data paired with image level label versus training data paired with instance level label. The experimental runs using sample level label were trained using 10 images that were randomly selected from a total of 30 images that had been manually labeled. About 1000 samples labeled as root and 1000 samples labeled as soil were randomly selected in each image. In order to run MIL methods with instance level labeled data, we put only one instance per bag and the bag label was assigned to the label of the only instance in the bag.

\begin{figure}[h] 
\begin{center}
\begin{subfigure}{0.23\textwidth}
   \includegraphics[width=1\linewidth]{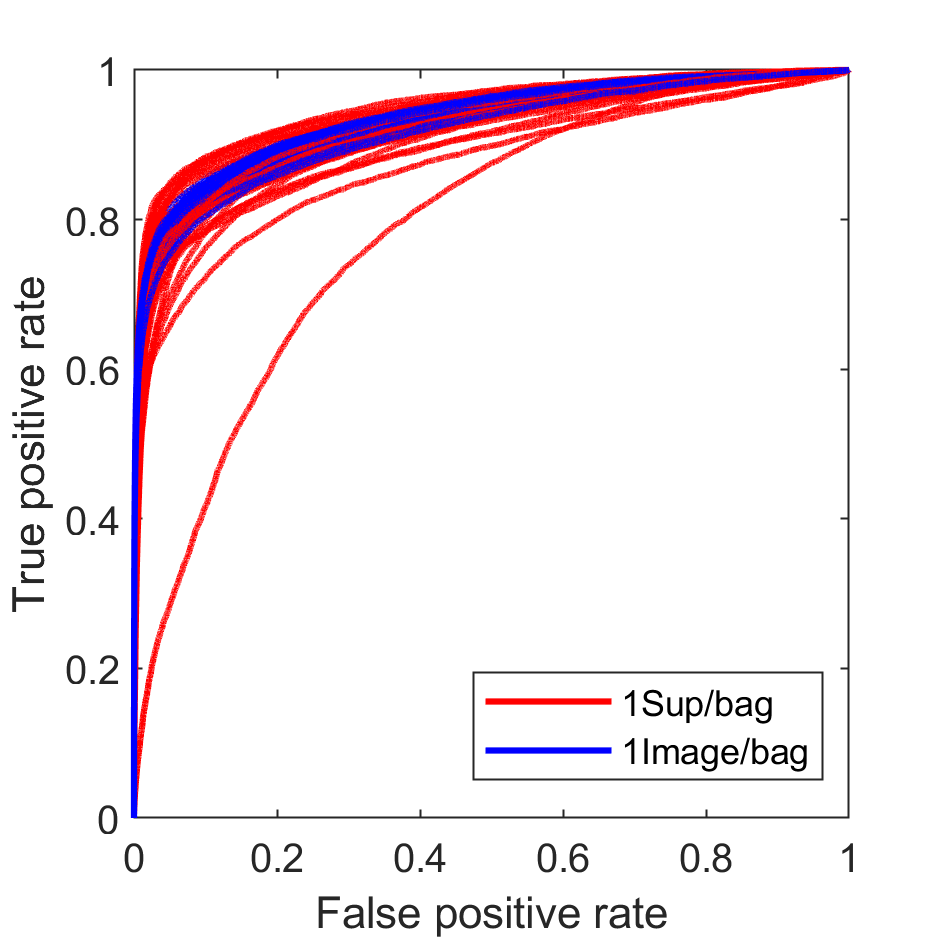}
   \caption{MI-ACE}
   \label{subfig:7_A}
\end{subfigure}
\begin{subfigure}{0.23\textwidth}
   \includegraphics[width=1\linewidth]{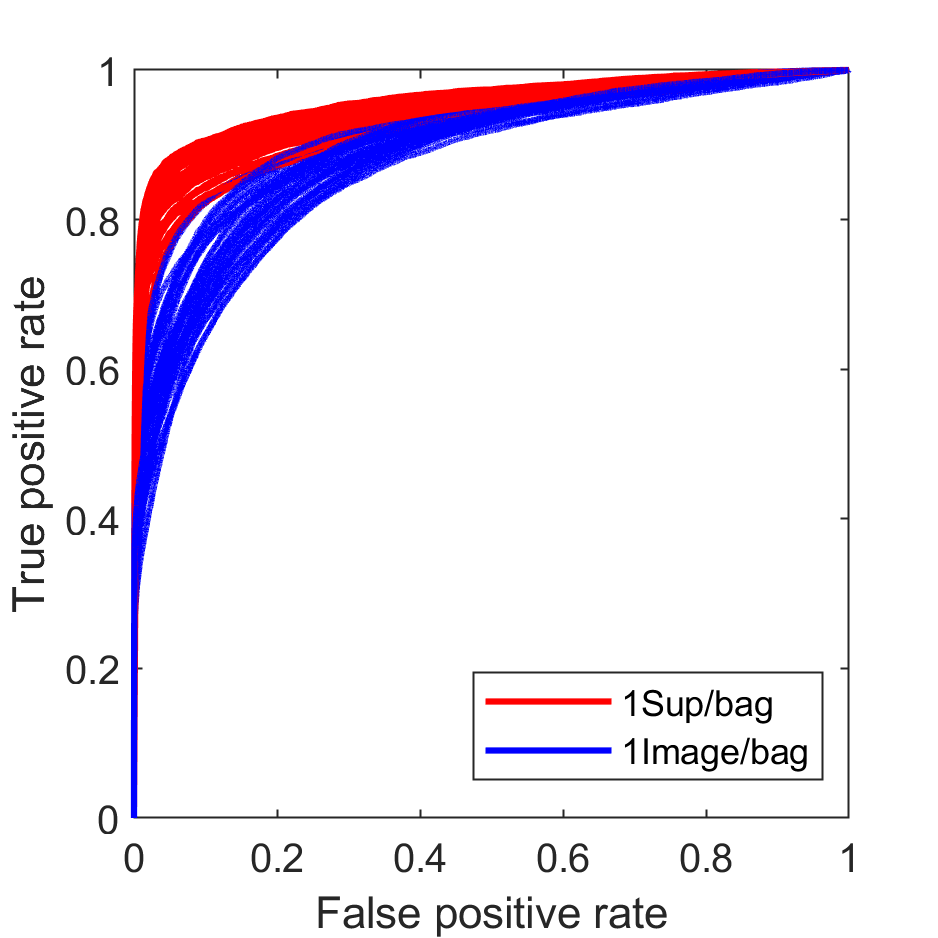}
   \caption{MIForests}
   \label{subfig:7_B}
\end{subfigure}

~~
\begin{subfigure}{0.23\textwidth}
   \includegraphics[width=1\linewidth]{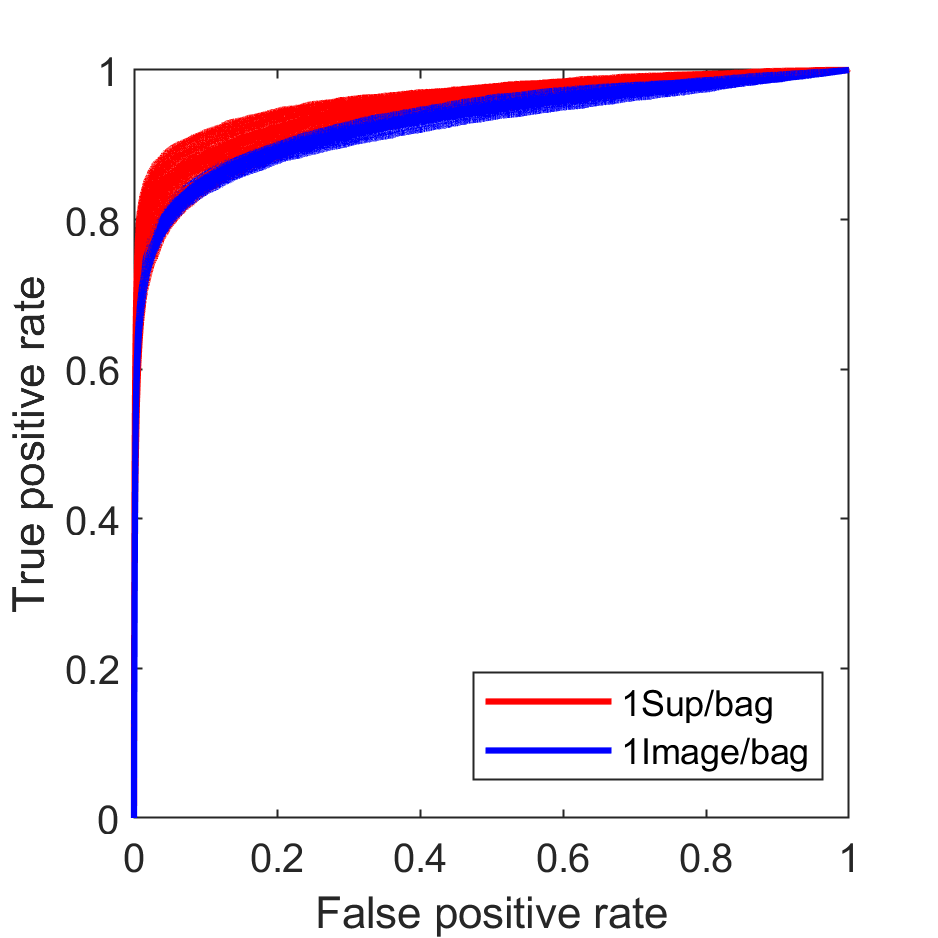}
   \caption{miSVM}
   \label{subfig:7_C}
\end{subfigure}
\begin{subfigure}{0.23\textwidth}
   \includegraphics[width=1\linewidth]{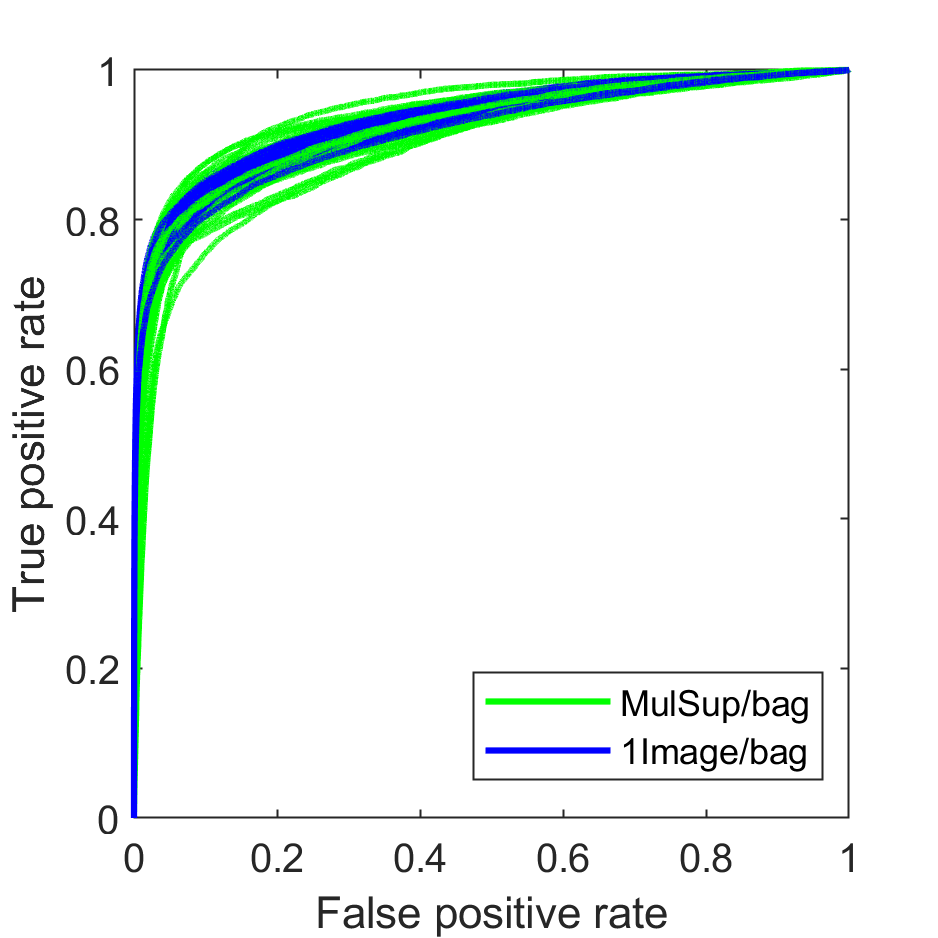}
   \caption{MI-ACE}
   \label{subfig:7_D}
\end{subfigure}

~~
\begin{subfigure}{0.23\textwidth}
   \includegraphics[width=1\linewidth]{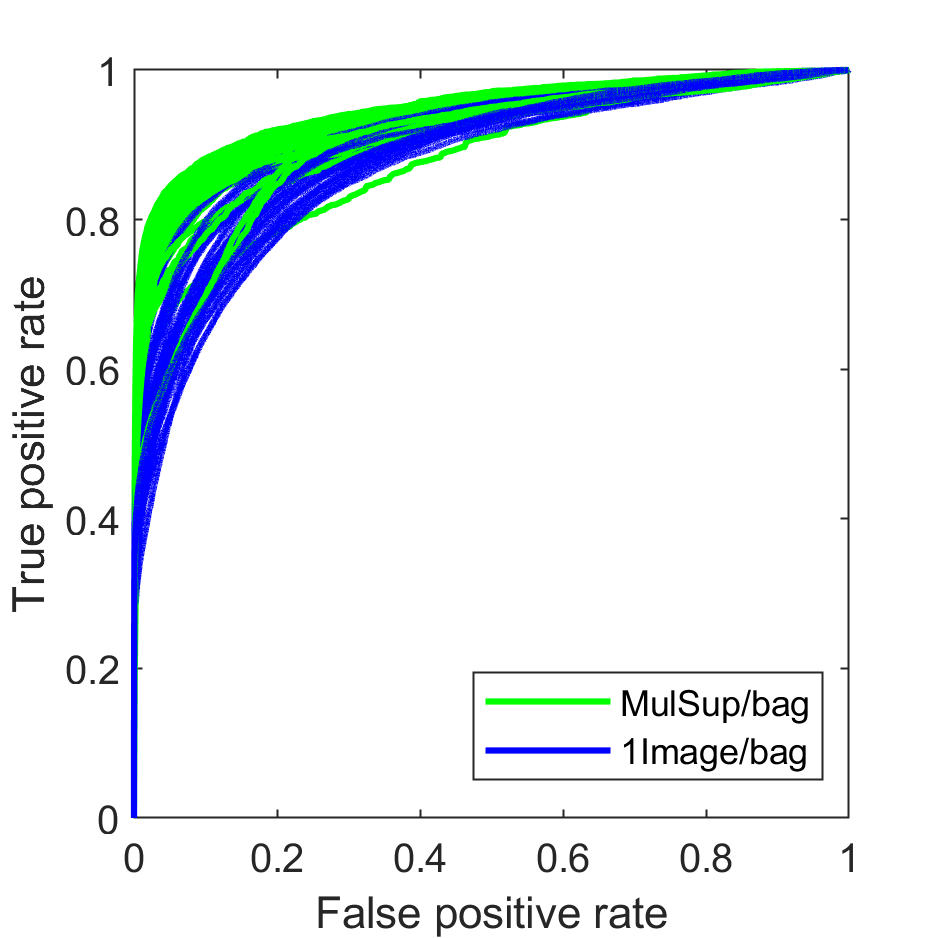}
   \caption{MIForests}
   \label{subfig:7_E}
\end{subfigure}
\begin{subfigure}{0.23\textwidth}
   \includegraphics[width=1\linewidth]{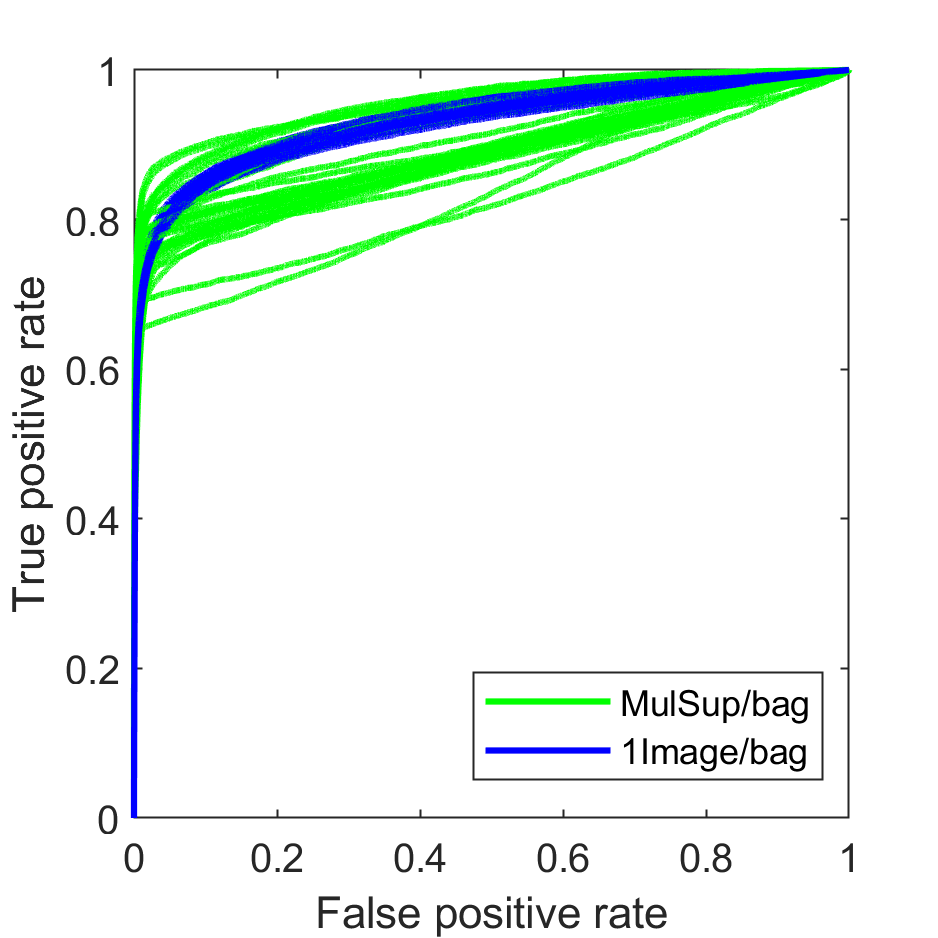}
   \caption{miSVM}
   \label{subfig:7_F}
\end{subfigure}

   \caption{ROC curves of MI-ACE, MIForests, and miSVM with different level of label accuracy of training data. Each algorithm is run 30 times}
\label{fig:7}
\end{center}
\end{figure}

ROC curves corresponding to instance and image level label accuracy can be seen in Fig. \ref{fig:7} (\protect\subref{subfig:7_A})-(\protect\subref{subfig:7_C}). Among all three MIL methods, MI-ACE was the most robust algorithm to noisy labeled training data. This was indicated by the overlapping of the ROC curves regardless of label accuracy (Fig. \ref{subfig:7_A}). MIForests and miSVM were more sensitive to label accuracy than MI-ACE (Fig. \ref{subfig:7_B} and \ref{subfig:7_C}). The ROC curves of MIForests and miSVM with image level label were lower than the ROC curves of MIForests and miSVM with instance level label. The results of MIForests and miSVM improved as the label became more accurate. The results also showed that, the variance of ROC curves training with instance level data was larger than that training with image level data, especially for MI-ACE and miSVM. The results of training with image level data was more consistent because it was easy to get a large number of images with image level label, but it was expensive to label every instance and hard to get a large amount of training data with instance level label. A large amount of training data helped the generalization of the learning model so that the test results were more consistent.

\subsubsection{Small Bag versus Image Level Label}
We also compared results using training data paired with image level label to results using training data paired with small bag level label. Small bags were generated by SLIC algorithm which were large enough to group about 10 adjacent instances into a bag. It was a positive bag if there were root samples in the bag. Otherwise, it was a negative bag. Similarly to training data with sample level label, the experiment runs using small bag level label were trained using 10 images that were randomly selected from a total of 30 images. About 100 positive bags and 100 negative bags were randomly selected in each image.

ROC curves corresponding to small bag level label and image level label can be seen in Fig. \ref{fig:7} (\protect\subref{subfig:7_D})-(\protect\subref{subfig:7_F}). The results of MI-ACE and miSVM using image level label training data were comparable with results using small bag level label because the ROC curves of MI-ACE and miSVM with different label accuracy overlapped with each other. But the results using image level labeled data were more consistent than the results using small bag level labeled data because there were more labeled data with image level label that improved the generalization of the learning model, indicating the importance of a large amount of data to the learning results. For MIForests, the detection results with small bag level training data were much better than that with image level training data, proved again the importance of precise label of training data to the learning results of MIForets.
\section{Discussion}
We compared three different MIL classifiers with same training data. These methods varied in segmentation performance. The miSVM algorithm outperformed the other approaches with a higher average TPR at the same FPR and MI-ACE was a close second. These methods varied in initialization strategies, optimization and model refinement strategies and the feature space used. For example, during initialization both miSVM and MIForests initializes instance labels by assigning each instance the same label as the bag whereas, for MI-ACE, a target signature is selected from a positive bag and used to initialize individual instance labels. During training the methods also use a variety of strategies to refine and update labels. For example, miSVM uses the model estimated from the last iteration to predict labels for each instance and, then, retrain and refine the model. MIForests randomly draws labels for instances in bag with respect to the probability distribution of labels predicted using the model trained from last iteration and retrains with these sampled labels. MI-ACE uses yet another refinement strategy in that it always selects the instance most likely to be target from each positive bag and uses these to update the target model. Also, these three different MIL algorithms work in different feature spaces. The miSVM algorithm transfer the input features into a high dimensional feature space through a kernel mapping, MI-ACE works in a whitened feature space, and MIForests works in the original input feature space. We believe that mi-SVM outperforms the others due to the high-dimensional kernel mapping and, thus, very non-linear decision boundary. MI-ACE, however, provides the advantage
 \clearpage
 
 \clearpage

\noindent that the learned target signature is interpretable and provides explainability to the user. Features played an important role in the performance of different classifiers. The mean, variance, and entropy of each band in the RGB and LAB color spaces were used as 18 dimensional features to represent each superpixel. This feature vector was effective at representing the color information of a set of pixels in a superpixel. We mainly considered the color information of each superpixel because the superpixels generated by SLIC were similar in shape. However, in future work, other high level spatial features such as object shape may be evaluated to determine differences between roots and soil pixels. We used SLIC to generate superpixels not only because it can generate superpixels similar in size and shape. It was also the most feasible method to make superpixels contain pixels from same class via controlling the average size (or the totally number) of superpixels and the spatial regularization. Using SLIC method to generate superpixels avoided large variance in supepixel size that the performance deteriorated sharply by a misclassified large superpixel. Our ongoing work will extend the experiments to other recently available minirhizotron root datasets to do evaluation in a larger scale.

\section{SUMMARY}
We propose the MIL approach to detect roots in minirhizotron images. Our MIL approach uses image-level labeled data to segment roots which significantly reduces the effort needed to label training data for the application of supervised learning algorithm. We tested the capability of three MIL methods to achieve competitive root detection over a range of soil textures and colors. Our superpixel feature histogram based downsampling strategy can effectively increase the target ratio in positive bag that the image-level labeled training data achieve comparable results to training data labeled in finer scale. The miSVM method performed the best in detecting roots with image-level labels among all three MIL algorithms. Overall, our results suggest application of MIL methods can substantially improve the imaging analysis bottleneck currently occurring in the study of roots by the minirhizotron technique.

\noindent\textbf{Guohao Yu} received the B.S. degree in electrical engineering from the Harbin Engineering University, Harbin, China, and the M.S. degree in optical engineering from the Chinese Academy of Sciences,Shanghai, China in 2008 and 2011, respectively. He is currently a graduate research assistant working toward the Ph.D. degree in the Department of Electrical and Computer Engineering, University of Florida. His research interests include multiple instance learning, semantic segmentation and minirhizotron image analysis.

\noindent\textbf{Alina Zare} received the Ph.D. degree from the University of Florida, Gainesville, in 2008. She is currently an associate professor in the Department of Electrical and Computer Engineering, University of Florida. Her research interests include machine learning and pattern recognition, sparsity promotion, target detection, hyperspectral image analysis, and remote sensing. She is a recipient of the 2014 National Science Foundation CAREER award and the 2014 National Geospatial-Intelligence Agency New Investigator Program Award. She is a senior member of the IEEE.

\noindent\textbf{Hudanyun Sheng} received the B.Sc. degree in physics from Tongji University, Shanghai, China, in 2015, the M.Sc. degree in Industrial and Systems Engineering from University of Florida, Gainesville, Florida. She is now working towards the M.Sc. degree in Electrical and Computer Engineering in University of Florida, and at the same time working as a graduate research assistant in the Machine Learning and Sensing Lab.

\noindent\textbf{Roser Matamala} received her M.Sc. and Ph.D. degree in biological sciences from the University of Barcelona, Spain, in 1993 and 1997 respectively. She was a Research Associate at Duke University in 1998 and then at Argonne National Laboratory in 2000. Currently, she is a Scientist at Argonne and a Fellow at the Northwestern-Argonne Institute of Science and Engineering and at the UChicago Consortium for Advanced Science and Engineering. She is an ecologist and biogeochemist who studies the carbon, water, and energy cycles at the ecosystem level and the changes occurring due to human factors, such as land-use and global climate changes. She is an investigator for many Department of Energy, Office of Science, Biological and Environmental Research projects. In these projects, she has been mapping the stocks and characterizing the composition of soil organic carbon in the Arctic region. She is also studying climate adaptation and sustainability of bioenergy crops across continental scale environmental gradients. In addition, she is developing new sensors for detecting changes in roots and soil moisture funded by ARPA.E and NSF. She has published over 50 articles in scientific journals, book chapters and scientific reports.

\noindent\textbf{Joel Reyes-Cabrera} was born in Leon, Nicaragua, where he lived 17 years and then moved to Costa Rica. He received his B.Sc. degree in agricultural sciences from EARTH University at Costa Rica in 2009. He entered graduate school at the University of Florida and received his M.Sc. and Ph.D. degrees in 2013 and 2017, respectively. His main areas of research interests are agricultural water management, plant physiology, bioenergy, and carbon fluxes in the agroecosystem. Dr. Reyes-Cabrera is currently a postdoctoral fellow in the Department of Plant Sciences at the University of Missouri where he conducts research on the root biology of bioenergy crops.

\noindent\textbf{Felix B. Fritschi} is a native of Peru and received an Ing. HTL degree from the Swiss College of Agriculture, an M.S. in Agronomy from the University of Florida, and a Ph.D. in Plant Biology from the University of California, Davis.  He joined the University of Missouri in 2007 and currently is the C. Alice Donaldson Professor of Bioenergy Crop Physiology and Genetics in the Division of Plant Sciences.  Dr. Fritschi is a member of the Interdisciplinary Plant Group at the University of Missouri and a Fellow of the Crop Science Society of America.
Dr. Fritschi has a long-standing interest in sustainable crop production and an active research program aimed at enhancing crop resource use efficiency, particularly as related to production under drought and heat stress conditions. Dr. Fritschi~\textquotesingle s program leverages physiological mechanism-based strategies and genetics to better understand and ultimately reduce the impacts of abiotic stress on crop plants. He is also engaged in the development of innovative tools and techniques to better characterize plant growth under field conditions, exploit natural genetic variation, and accelerate crop improvement.

\noindent\textbf{Thomas E. Juenger} is a Professor in the Department of Integrative Biology at the University of Texas at Austin. He received his Ph.D. in Ecology and Evolutionary Biology at the University of Chicago (1999) and completed postdoctoral research at the University of California Berkeley as a Miller Fellow (1999-2002). Tom~\textquotesingle s research focuses on interface of ecological and evolutionary quantitative genetics in natural plant populations. Much of his research has focused on studies of the genetics of physiological traits, abiotic stress tolerance, and local adaptation and gene-by-environment interaction. More recently, he has led several projects exploring local adaptation in C4 perennial grasses including studies of plant root systems.

\end{document}